\pdfoutput=1
\documentclass[10pt]{article}
\usepackage[preprint]{tmlr}



\usepackage{amsmath,amsfonts,bm}

\def\eqref#1{equation~\ref{#1}}
\def\1{\bm{1}}

\DeclareMathAlphabet{\mathsfit}{\encodingdefault}{\sfdefault}{m}{sl}
\SetMathAlphabet{\mathsfit}{bold}{\encodingdefault}{\sfdefault}{bx}{n}

\usepackage[T1]{fontenc}
\usepackage[utf8]{inputenc}
\usepackage{lmodern}
\usepackage{microtype}
\usepackage[bookmarks=true,bookmarksnumbered=true,bookmarksopen=true]{hyperref}
\usepackage{bookmark}
\usepackage{placeins}
\usepackage{url}
\usepackage{graphicx}
\usepackage{booktabs}
\usepackage{array}
\usepackage{makecell}
\usepackage{multirow}
\usepackage{tabularx}
\usepackage{float}
\usepackage{amsmath}
\usepackage{xcolor}
\usepackage[font=small,labelfont=bf]{caption}
\usepackage{enumitem}
\setlist{itemsep=1pt,topsep=3pt,parsep=0pt,before={\setlength{\parfillskip}{0pt plus 0.85\linewidth}}}
\AtBeginDocument{\setlength{\parfillskip}{0pt plus 0.85\linewidth}}
\hypersetup{pdftitle={What the Averages Hide: Monocular Metric Depth Models Still Fail in Everyday Scenes, Mostly on Scale},pdfauthor={Hualie Jiang, Yujie Luo, Ziyang Song, Ximeng Li, Zhiqiang Lou, Rui Xu},colorlinks=true,linkcolor=[rgb]{0.1,0.2,0.55},citecolor=[rgb]{0.1,0.2,0.55},urlcolor=[rgb]{0.1,0.2,0.55}}

\title{What the Averages Hide: Monocular Metric Depth Models Still Fail in Everyday Scenes, Mostly on Scale}

\author{\name Hualie Jiang$^{1}$, Yujie Luo$^{1,2,}$\thanks{Work done during an internship at Antigravity Tech, Insta360.}, Ziyang Song$^{1}$, Ximeng Li$^{1}$, Zhiqiang Lou$^{1}$, Rui Xu$^{1,}$\thanks{Corresponding author: \texttt{jerry1@insta360.com}.} \\ \addr $^{1}$Antigravity Tech, Insta360 \quad $^{2}$Nanyang Technological University \\ \addr Project page: \href{https://antigravity-tech.github.io/DIMED/}{\upshape\texttt{antigravity-tech.github.io/DIMED}}}

\def\month{MM}
\def\year{YYYY}
\def\openreview{\url{https://openreview.net/forum?id=XXXX}}

\begin{document}

\maketitle

\begin{abstract}
Zero-shot monocular metric depth models are evaluated by computing metric and scale-aligned metrics on each dataset and averaging them across datasets. Averages alone cannot show how many images still have severely wrong metric depth. The gap between the two scores shows how much error comes from scale, but not whether depth is underestimated or overestimated, or how consistently; and without a per-image judgement of failure, they cannot compare whether different models fail on the same images. We propose an evaluation built on per-image results: the failure rate (metric $\delta_1$ $<$ 0.5) as the criterion for whether the task is solved; scale-aligned metrics for structure, and the bias, dispersion and absolute error of the alignment scale factor for scale; and a cross-model comparison on the same images that separates shared from model-specific failures. We apply it to 11 recent architectures with and without ground-truth intrinsics (18 model evaluations) on the 7 everyday-scene benchmarks (4,406 images) used by MoGe-2. Unless stated otherwise, rates are averaged over the 7 datasets with equal weight. Metric depth estimation is not solved in everyday scenes, and the remaining difficulty is mainly scale: the best of the 18 evaluations still has a failure rate of 7.7\%; 78.8\% of failures have intact structure and are wrong only in scale, and after an ideal scale correction from the ground truth 98.3\% no longer fail (pooled over all failures of the 18 evaluations). More than a third of the error of the top 6 models falls on images that most models fail: with intrinsics, images failed by at least 6 architectures make up 7.3\% of the data, concentrated on HAMMER (4 of its 7 scenes) and DIODE; even when selected only by architectures of other families, such images hold 34\% to 40\% of these models' $1-\delta_1$ error, against 14\% to 18\% for as many random images from the same datasets. The scale error varies with image content (in HAMMER pairs, adding objects reduces the scale error of most models); ground-truth intrinsics mainly affect the scale, and in most families larger models reduce both structure and scale errors, but the failures remain far from removed. We recommend reporting failure rates, structure and scale separately, and results stratified by how many models fail. We will release the evaluation code, the per-image results with their shared-failure labels, and two diagnostic sets built from TUM RGB-D for studying how image cues relate to the scale error.
\end{abstract}

\section{Introduction}\label{sec:1}

Recent zero-shot monocular metric depth models have advanced quickly (e.g. \citealp{hu2024metric3dv2,bochkovskii2025depthpro,wang2025moge2,piccinelli2025unidepthv2,ma2026metricanything}): they are trained on large mixtures of data, are not tuned to the test sets, and report high mean accuracy on several benchmarks. Their papers also show qualitative results, many of them marking object sizes or point-to-point distances on the predicted point cloud next to the ground truth \citep{yin2023metric3d,hu2024metric3dv2,wang2025moge2,ma2026metricanything,liu2026foundationgeo}; their limitation sections discuss fine structures, translucent surfaces, camera models and similar issues, and when the ambiguity of scale is mentioned, it is usually attributed to out-of-distribution scenes (e.g. \citealp{wang2025moge2}). These models aim to generalise zero-shot to any image \citep{bochkovskii2025depthpro} or to open-domain data \citep{wang2025moge2,liu2026foundationgeo}, a range that includes everyday indoor and outdoor scenes; but neither the demonstrations nor the means show how many images of such scenes still have severely wrong metric depth. We test this on such scenes (not specialised domains such as aerial or endoscopic imagery, Appendix~\hyperref[app:A]{A}). For the four models that the MoGe-2 paper evaluated under the same protocol, our results essentially match its values (Appendix~\hyperref[app:E]{E}), so we do not dispute the reported numbers but reveal what the averages hide: which images fail, whether the error is in structure or scale, in which direction the scale is off, and whether a failure is shared.

Current evaluations compute metric and scale-aligned metrics on each dataset and then average the scores or ranks across datasets. This practice leaves three gaps. \textbf{First, it looks at means, not failures.} A mean does not distinguish ``every image is slightly off'' from ``most images are good and a few are severely wrong'', yet depth is used one image at a time. With intrinsics, for example, MoGe-3 \citep{kong2026moge3} reaches nearly the same mean metric $\delta_1$ on KITTI \citep{geiger2013vision} and ETH3D \citep{schops2017eth3d}, 0.926 and 0.927, yet fails (metric $\delta_1$ $<$ 0.5) on 0.3\% and 5.5\% of their images. \textbf{Second, alignment discards the scale.} A metric score requires both structure and scale to be right; a scale-aligned score first aligns the prediction to the ground truth with one global scale and then measures structure only. Papers often report the two side by side, and their gap reflects the size of the scale error; per image, they also tell scale failures from structure failures (Section~\hyperref[sec:4.3]{4.3}). But the scale factor $s^{*}$ found during alignment is discarded after use, although it carries the direction of the scale error and whether it is consistent across images. The metric and scale-aligned scores reported by current evaluations are unsigned, so neither tells underestimation from overestimation of depth, nor a systematic scale bias (all images of a dataset off in the same direction by similar amounts, fixed by one global factor) from an unstable scale (sometimes too small and sometimes too large, not fixed by any global correction); the former is measured by the bias and the latter by the dispersion (Section~\hyperref[sec:3.3]{3.3}). Existing surveys and benchmarks do not measure scale separately either: they align predictions affinely before computing metrics, which removes the scale error \citep{liu2026monocular,ge2024geobench}, or evaluate through downstream tasks, which cannot tell scale errors from structural ones \citep{li2025benchdepth}. \textbf{Third,} current zero-shot metric depth evaluations report means and do not judge whether a single image fails, so they cannot compare whether different models fail on the same images, nor tell whether a model's failures are its own or shared. A model paper need only report its own results; this can be seen only by comparing several models on the same images, which is what an evaluation study can add. The two differ in meaning: a model-specific failure shows that other models solve the same image, so the problem lies mostly in that model; a shared failure is a difficulty common to most models and the one future models most need to solve. These images can hold most of a model's error on a dataset: 89\% of the $1-\delta_1$ error of MoGe-2 on HAMMER (\citealp{jung2023hammer}; Section~\hyperref[sec:5.4]{5.4}).

Our evaluation is built on per-image results and answers the three gaps in turn (Section~\hyperref[sec:3.3]{3.3}). Like MoGe-2, we still average per dataset and then across datasets, but we first keep the metric $\delta_1$, the scale-aligned $\delta_1$ and $s^{*}$ of every image and derive all statistics from them. \textbf{First,} we judge whether the problem is solved by the failure rate, the share of images whose metric depth is severely wrong. \textbf{Second,} we split the metric error into structure, measured with scale-aligned metrics, and scale, measured with the bias, dispersion and absolute error of $s^{*}$. The split is needed because the two errors call for different remedies. The absolute scale is harder to judge from pixels alone than the structure: a photograph of a blank wall may show a wall 1 m away or 5 m away; the relative geometry of the wall (a plane and its orientation), which we call structure, also relies on image cues, but not on objects of known size, and does not depend on the distance. When structure is right and scale is wrong, in principle a single known length corrects all depths, whereas a structural error cannot be corrected this way. Structure and scale are analysis dimensions applied at evaluation time, not properties of the models, so models with and without an explicit decomposition (Section~\hyperref[sec:3.1]{3.1}) are compared on the same terms. \textbf{Third,} we compare 11 architectures on the same images, grade failures into shared, intermediate and model-specific tiers by the number of architectures that fail, and split the mean error of each model by these tiers. On top of this, we run two comparisons: how the scale error changes when only the image content changes, and what ground-truth intrinsics and model size do to the scale and the failures (Sections~\hyperref[sec:6]{6}, 7).

On the 7 everyday-scene benchmarks of the MoGe-2 evaluation (4,406 images), we evaluate 11 recent architectures with and without ground-truth intrinsics, for 18 model evaluations, and the other public sizes within 5 families (Section~\hyperref[sec:3]{3}, Appendix~\hyperref[app:I]{I}). Our main findings are as follows. (i) The problem is not yet solved, and the remaining difficulty lies mainly in scale: the best of the 18 evaluations still has a failure rate of 7.7\%; pooled over all failures of the 18 evaluations, 78.8\% have intact structure (scale-aligned $\delta_1$ $\geq$ 0.9) and are wrong only in scale, after scale correction 98.3\% no longer fail, flawed structure occurs mainly on the DDAD dataset \citep{guizilini2020packing}, and structure failures mainly in the DepthVLM model \citep{yu2026depthvlm} (Section~\hyperref[sec:4]{4}). (ii) More than a third of the error of the top 6 models falls on the few images that most models fail: with intrinsics, the images failed by at least 6 architectures make up only 7.3\% of the data, are concentrated on HAMMER (4 of its 7 scenes) and DIODE \citep{vasiljevic2019diode}, and on HAMMER are mostly empty-table shots; even when selected only by architectures of other families, such images account for 34\% to 40\% of these models' $1-\delta_1$ error, against 14\% to 18\% for as many random images from the same datasets; a ranking by metric $\delta_1$ is decided by the other images, whereas a ranking by Rel depends on these (Section~\hyperref[sec:5]{5}). (iii) The scale error varies with image content: in HAMMER image pairs that differ only in whether objects are present, adding objects (with the texture and geometry they bring) reduces the mean scale error of 10 of the 11 models (Section~\hyperref[sec:6.3]{6.3}); on the 4 sequences of TUM RGB-D \citep{sturm2012benchmark} with planar printed patterns, 43 of the 44 ``model $\times$ sequence'' cells stably underestimate depth, and these sequences are separate recordings (only a correlation; Section~\hyperref[sec:6]{6}). (iv) Ground-truth intrinsics mainly affect the scale, and in most families larger models reduce both structure and scale errors, but the failures remain far from removed: intrinsics raise metric $\delta_1$ yet lower the scale-aligned $\delta_1$ of the four models that share the MoGe solver (Section~\hyperref[sec:7]{7}).

\paragraph{Contributions}

\begin{enumerate}
\item An evaluation protocol that uses the failure rate as the main criterion and measures structure and scale separately, with dedicated statistics for scale (the bias, dispersion and absolute error of $s^{*}$); it suits any metric depth model, decomposed or not (Section~\hyperref[sec:3.3]{3.3}), and addresses the first two gaps.
\item A per-image, cross-model analysis that grades failures into shared, intermediate and model-specific tiers by the number of architectures failing on the same image, and splits the mean error of each model by these tiers, separating the failures that lower the means of all models from those that decide the ranking (Section~\hyperref[sec:5]{5}). It addresses the third gap.
\item Public resources: the evaluation code; per-image shared-failure labels on the 7 benchmarks and the per-image results of all model evaluations (the 18 of the main table and another 18 size evaluations in Appendix~\hyperref[app:I]{I}), so that a new model can split its error by shared failures directly against these labels; and two diagnostic sets built from TUM RGB-D that vary structure, texture and distance under the same camera and intrinsics (each sequence a separate recording) (Sections~\hyperref[sec:5]{5}, 6, 8).
\end{enumerate}

\section{Related Work}\label{sec:2}

\textbf{In-domain depth estimation.} Monocular depth estimation was first trained and tested on a single dataset \citep{eigen2014depth,laina2016deeper,fu2018deep,lee2019from,bhat2021adabins}, and the scale-invariant error of \citet{eigen2014depth}, together with the Rel and $\delta_1$ metrics it adopted, is still in use; self-supervised methods \citep{zhou2017unsupervised,godard2019digging,spencer2022deconstructing} recover depth only up to scale and are evaluated after per-image median alignment. The metric scale of these models, where they have one, comes from the camera and scene distribution of a single training set and usually does not transfer to other datasets.

\textbf{Zero-shot relative depth estimation.} To generalise across datasets, one line of work gives up metric scale and predicts only relative depth. \citet{chen2016singleimage} trained a depth network on relative-depth annotations of random point pairs in web images. MiDaS \citep{ranftl2020towards} proposed scale- and shift-invariant losses, trained on a mixture of datasets with different depth ranges, and evaluated zero-shot on unseen datasets; DPT \citep{ranftl2021vision} replaced the backbone with a vision transformer. Depth Anything and Depth Anything V2 \citep{yang2024depthanything,yang2024depthv2} scaled up training with large amounts of unlabelled and synthetic images; Marigold \citep{ke2024repurposing,pavlovic2026marigoldv2} fine-tuned a pretrained image diffusion model into an affine-invariant depth estimator; MoGe \citep{wang2025moge} predicts an affine-invariant point map and solves for the focal length and shift from it. Their evaluation aligns predictions affinely to the ground truth and so assesses only structure; they show that relative depth can be estimated accurately zero-shot across a wide range of scenes.

\textbf{Zero-shot metric depth estimation.} Recent work tries to recover both structure and scale zero-shot. Most methods start from the camera: an object imaged near with a short focal length can look the same as one imaged far with a long focal length, which a network that sees only the image cannot distinguish. Earlier work fed camera parameters as extra input channels (CAM-Convs, \citealp{facil2019camconvs}). Metric3D \citep{yin2023metric3d} proposed the canonical camera space transformation, which maps all training data to one canonical focal length, and Metric3D v2 \citep{hu2024metric3dv2} keeps this transformation, switches to a ViT backbone and adds joint depth–normal optimisation. ZeroDepth \citep{guizilini2023towards} uses geometric embeddings derived from the intrinsics as input. UniDepth and UniDepthV2 \citep{piccinelli2024unidepth,piccinelli2025unidepthv2} predict camera rays and condition the depth prediction on them, and UniK3D \citep{piccinelli2025unik3d} extends this to other camera models. Depth Pro \citep{bochkovskii2025depthpro} has a focal length prediction head. MoGe-2 \citep{wang2025moge2} predicts an affine-invariant point map and a metric scale, and solves for the focal length and shift afterwards. MoGe-3 \citep{kong2026moge3} adds sparse volumetric refinement. MetricAnything \citep{ma2026metricanything} scales up pretraining on noisy, heterogeneous data sources. FoundationGeo \citep{liu2026foundationgeo} predicts spatially varying per-pixel fields. DA3METRIC \citep{lin2025da3} is the metric branch of Depth Anything 3, and DepthVLM \citep{yu2026depthvlm} builds on a vision-language model.

\textbf{Surveys and evaluations.} The metric depth survey of \citet{zhang2025survey} collects the results reported in the original papers and notes that inconsistent datasets, training protocols and model settings make fair comparison difficult. The survey of \citet{liu2026monocular} evaluates many depth foundation models on 10 benchmarks under one protocol, but likewise aligns metric models affinely first, comparing only the depth distribution and not scale. GeoBench \citep{ge2024geobench} compares training recipes and fine-tuning strategies of geometry estimation models in a unified codebase, restricts the evaluation to affine-invariant depth, and leaves metric depth to future work. BenchDepth \citep{li2025benchdepth} argues that alignment-based metrics introduce bias and instead evaluates 8 models through 5 downstream tasks, including depth completion, stereo matching and SLAM. RoboDepth \citep{kong2023robodepth} tests the robustness of 42 models under 18 image corruptions such as weather and noise. The iBims-1 evaluation \citep{koch2018evaluation} proposed directed depth errors, the share of pixels predicted too far or too near relative to a fixed reference plane (e.g. 3 m); they are per pixel and do not separate structure from scale. The papers of the 11 evaluated models have not adopted such metrics. Van Dijk and de Croon~\citeyearpar{dijk2019how} modify input images to analyse which depth cues a self-supervised model relies on, and find that it relies mainly on the vertical position of objects rather than their apparent size. None of the papers of the 11 evaluated models contains a metric dedicated to depth scale or reports per-image depth results or failure rates (Depth Pro reports per-image success rates of its focal length estimate). In image classification, \citet{geirhos2020beyond} measure error consistency, i.e. whether two models err on the same inputs, and \citet{meding2022trivial} find that most ImageNet validation images are either classified correctly by all or all but one model (46\%) or by none or only one (11.5\%, partly due to label problems), so model differences show only on the remaining images. We bring this analysis to metric depth and add the direction and consistency of scale.

\section{Evaluation Protocol}\label{sec:3}

\subsection{Models}\label{sec:3.1}

\textbf{Selection criteria.} We evaluate recent state-of-the-art models that perform monocular metric depth estimation across indoor and outdoor scenes with a single set of weights, and exclude models with separate indoor and outdoor weights, models that output only relative depth, models that cannot output a dense depth map (such as DepthLM, \citealp{cai2025depthlm}), and variants of MoGe-2 patched for aerial and very distant scenes (the excluded models and the reasons are given in Appendix~\hyperref[app:A]{A}).

\textbf{Selected models.} The 11 selected architectures are listed in Table~\hyperref[tab:1]{1}. Below, ``K'' denotes the ground-truth camera intrinsics, and ``with K / without K'' indicates whether the ground-truth intrinsics are given to the model at evaluation; one complete evaluation of one architecture under one intrinsics condition is called a ``model evaluation''. Table~\hyperref[tab:1]{1} groups them into four classes by how the intrinsics enter the model: canonical camera space with output conversion, canonical camera space with input rescaling, post-hoc geometric solve, and rays as network input; this is needed to understand whether supplying the intrinsics helps (mechanisms and formulas in Appendix~\hyperref[app:D]{D}). Metric3D-v2, DA3METRIC, MetricAnything-DepthMap and DepthVLM are designed around a known focal length and are evaluated only with ground-truth intrinsics; the other 7 are evaluated both without and with intrinsics, for 18 model evaluations in total. 9 architectures use a ViT-L-level DINOv2 \citep{oquab2024dinov2} or DINOv3 \citep{simeoni2025dinov3} encoder; MetricAnything-DepthMap has only one public size, DINOv3 ViT-H+/16; and DepthVLM is built on the vision-language model Qwen3-VL \citep{bai2025qwen3vl} (Table~\hyperref[tab:1]{1}). Other sizes are evaluated separately (Section~\hyperref[sec:7]{7}, Appendix~\hyperref[app:I]{I}). All weights are the authors' public releases. The training data of DepthVLM \citep{yu2026depthvlm}, Metric3D-v2 \citep{hu2024metric3dv2}, DA3METRIC \citep{lin2025da3} and the two MetricAnything models (MA-DepthMap and MA-PointMap; \citealp{ma2026metricanything}) include DDAD, so DDAD is not a zero-shot test for these five models, which matters when reading their DDAD results. The other 6 architectures list none of our 7 benchmarks in their training data.

\begin{table}[t]
\centering
\renewcommand{\thetable}{1}
\caption{The 11 architectures evaluated in this paper, grouped by how the intrinsics enter the model (mechanisms in Appendix~\hyperref[app:D]{D}). In the ``K'' column, ``optional'' means the model still produces metric depth without intrinsics, and ``required'' means it has no metric scale without them and is evaluated only with them. The last column is the source of the focal length without intrinsics; backbones are read from code and weight files. $^\dagger$ Fine-tuned from MoGe-2.}
\label{tab:1}
\footnotesize
\setlength{\tabcolsep}{3.5pt}
\begin{tabularx}{\linewidth}{@{}>{\raggedright\arraybackslash}p{2.25cm}l>{\raggedright\arraybackslash}X ll@{}}
\toprule
Use of intrinsics & Model & Backbone & K & Focal length w/o K \\
\midrule
\multirow{4}{=}{Canonical camera space, output conversion} & Metric3D-v2 \citep{hu2024metric3dv2} & DINOv2 ViT-L & required & none \\
 & DA3METRIC \citep{lin2025da3} & DINOv2 ViT-L & required & none \\
 & Depth Pro \citep{bochkovskii2025depthpro} & DINOv2 ViT-L/16 (384) & optional & FoV head \\
 & MetricAnything-DepthMap \citep{ma2026metricanything} & DINOv3 ViT-H+/16 & required & none \\
\midrule
Canonical camera space, input rescaling & DepthVLM \citep{yu2026depthvlm} & Qwen3-VL (about 4B) & required & none \\
\midrule
\multirow{4}{=}{Post-hoc geometric solve} & MoGe-2 \citep{wang2025moge2} & DINOv2 ViT-L & optional & point-map solve \\
 & MoGe-3 \citep{kong2026moge3} & DINOv2 ViT-L & optional & point-map solve \\
 & MetricAnything-PointMap \citep{ma2026metricanything} & DINOv2 ViT-L$^\dagger$ & optional & point-map solve \\
 & FoundationGeo \citep{liu2026foundationgeo} & DINOv3 ViT-L/16 & optional & point-map solve \\
\midrule
\multirow{2}{=}{Rays as network input} & UniDepthV2 \citep{piccinelli2025unidepthv2} & DINOv2 ViT-L/14 & optional & camera head \\
 & UniK3D \citep{piccinelli2025unik3d} & DINOv2 ViT-L/14 & optional & camera head \\
\bottomrule
\end{tabularx}
\end{table}

\subsection{Benchmarks}\label{sec:3.2}

Test data were chosen by three rules: the data are the original public release; the ground truth does not depend on any evaluated model; and the data were not designed around the weaknesses of a particular model. We therefore use, unchanged, the 7 real-world benchmarks with metric ground truth in the MoGe-2 evaluation (its other 3 benchmarks, GSO, Sintel and Spring, are rendered and outside our scope): NYUv2 (\citealp{silberman2012indoor}; 654 images), KITTI (\citealp{geiger2013vision}; 652), ETH3D (\citealp{schops2017eth3d}; 454), iBims-1 (\citealp{koch2018evaluation}; 100), DDAD (\citealp{guizilini2020packing}; 1,000), DIODE (\citealp{vasiljevic2019diode}; 771) and HAMMER (\citealp{jung2023hammer}; 775), 4,406 images in total. We did not select images, and we keep the preprocessing of the MoGe evaluation loader (which reprojects every image to a view with the principal point at the centre). Section~\hyperref[sec:5]{5} counts failures by recorded scene, taken from the dataset directory structure: 13 for ETH3D, 6 for DIODE (3 indoor and 3 outdoor) and 7 for HAMMER; the other 4 datasets are not counted by scene. In 6 of the 7 HAMMER scenes, a robot arm recorded the same trajectory twice, once with objects on the table and once with an empty table (called ``empty table'' below); of the 775 images, 404 show objects and 371 an empty table. The DIODE ground truth was captured with a FARO Focus S350 laser scanner (nominal range 0.6 to 350 m, error about $\pm$1 mm); regions beyond the range (including the sky) and highly specular surfaces return no depth, so the ground truth of outdoor images covers only part of the pixels \citep{vasiljevic2019diode}. We take the released ground truth of each benchmark as given and do not check it independently. The intrinsics follow the MoGe benchmark. Appendix~\hyperref[app:A]{A} lists datasets we considered but did not use (e.g. MetricScenes).

\subsection{Metrics}\label{sec:3.3}

\textbf{Aggregation.} We still average with datasets weighted equally, as in the MoGe-2 paper: unless stated otherwise, the ``average over the 7 benchmarks'' first averages over all images within each dataset and then takes the mean over the 7 datasets with equal weight, and the three scale statistics are computed the same way. The difference is that we first keep the metric $\delta_1$, the scale-aligned $\delta_1$ and $s^{*}$ of every image, and derive the failure rate, the scale statistics and the cross-model grading from them.

\begin{figure}[t]
\centering
\renewcommand{\thefigure}{1}

\includegraphics[width=1.00\linewidth]{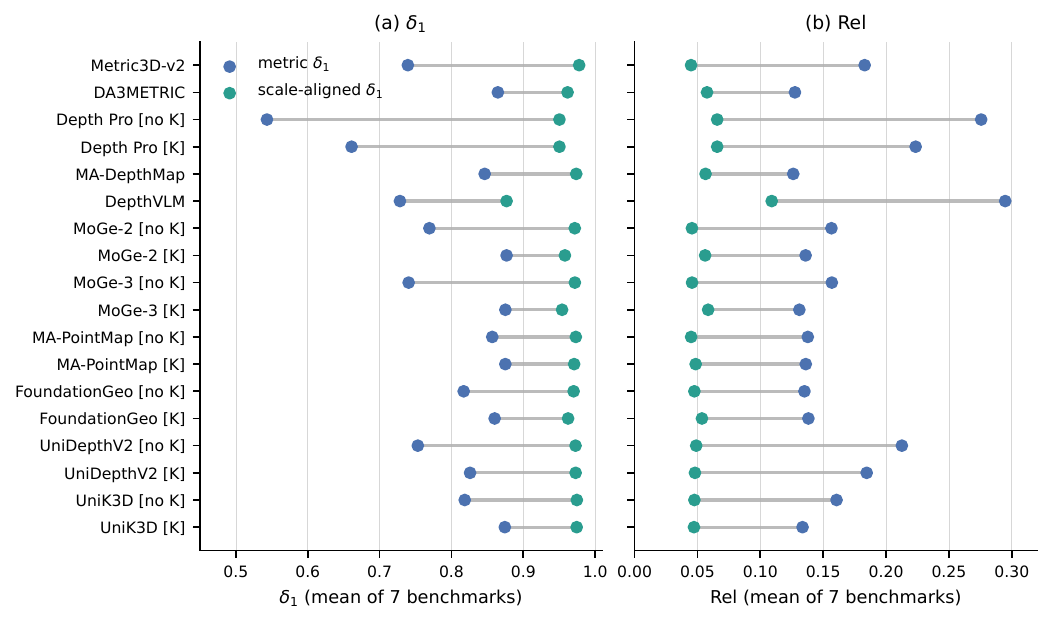}
\caption{Metric and scale-aligned accuracy of each model and condition, averaged over the 7 benchmarks. (a) $\delta_1$; (b) Rel. The grey bars show the part of the error that a single global scale removes.}
\label{fig:1}
\end{figure}

\textbf{Decomposing the metric error.} The error of monocular metric depth has two aspects: structure (whether the relative geometry of the scene is correct) and scale (how much the scene is enlarged or shrunk as a whole). Metric metrics contain both, so we measure them separately: structure with scale-aligned metrics, and scale with statistics of the alignment scale $s^{*}$; Figure~\hyperref[fig:1]{1} contrasts the two for every model. The metric error can also be split into a structure term and a scale term by decomposing the log error \citep{eigen2014depth}; the conclusions agree with ours, but its exact additive split holds only for the log error, its scale term $\mu^2$ carries no sign, and it does not separate systematic bias from instability (Appendix~\hyperref[app:B]{B}).

For each image, let the predicted depth be $p$ and the ground truth on valid pixels be $g$:

\begin{itemize}
\item \textbf{Metric} Rel and $\delta_1$ compare $p$ and $g$ directly. For one image, Rel is the mean of $\lvert p - g\rvert / g$ over its valid pixels, and $\delta_1$ is the fraction of valid pixels with $\max(p/g, g/p) < 1.25$.
\item \textbf{Structure}, i.e. scale-aligned Rel and $\delta_1$, is computed after multiplying $p$ by $s^{*}$, as in the scale-invariant metrics of MoGe and related work, but with $s^{*}$ estimated differently: $s^{*}$ is the median of $g/p$ over the valid pixels, the L1-optimal scale for the per-pixel log error; it marks fewer intact images as structure failures than the log mean or the ratio of medians of early self-supervised evaluation \citep{zhou2017unsupervised,godard2019digging} (Appendix~\hyperref[app:C]{C}). We do not use the MoGe protocol's minimiser of $\textstyle\sum \lvert s\,p - g\rvert / g$ on a 64$\times$64 grid: it favours small $s^{*}$, and a few outliers can pull it close to 0 and turn an image with intact structure into a structure failure (Appendix~\hyperref[app:C]{C}). $s^{*}$ does not maximise $\delta_1$, so the scale-aligned $\delta_1$ is occasionally slightly below the metric $\delta_1$, by small amounts.
\item \textbf{Scale} is the alignment scale $s^{*}$ itself (also called the ``scale factor'' below; earlier evaluations compute it for the scale-aligned metrics and then discard it), with $s^{*} \cdot p \approx g$: $s^{*} > 1$ means the model underestimates depth, and $s^{*} < 1$ that it overestimates depth. Over all $s^{*}$ of a dataset or a model evaluation, we report three statistics on $\ln s^{*}$, so that $s^{*} = 0.5$ and $s^{*} = 2$ count as equal errors:
\begin{itemize}
\item \textbf{Bias}: $\operatorname{mean}(\ln (1/s^{*}))$. A positive value means depth is overestimated overall (predicted depth larger than the ground truth) and a negative value underestimated, matching the sign of ``estimate minus truth'' (opposite to the direction of $s^{*}$).
\item \textbf{Dispersion}: $\operatorname{std}(\ln s^{*})$, computed within each dataset; larger means a less consistent scale across the images of a dataset. The bias and dispersion of Table~\hyperref[tab:3]{3} are equally weighted averages of within-dataset statistics and do not reflect differences in bias between datasets (Figure~\hyperref[fig:F8]{F.8}).
\item \textbf{Absolute error}: $\operatorname{mean}(\lvert\ln s^{*}\rvert)$. Bias cannot replace it, as under- and overestimates cancel in the mean.
\end{itemize}
\item \textbf{Failure}: an image fails when its metric $\delta_1$ $<$ 0.5, i.e. fewer than half of its valid pixels are within a factor of 1.25 of the ground truth; the ``failure rate'' is the fraction of failed images. This criterion targets uses that need metric scale; its sensitivity is given in Appendix~\hyperref[app:G]{G}.
\end{itemize}

\begin{table}[t]
\centering
\renewcommand{\thetable}{2}
\caption{Failure categories and thresholds for one ``image $\times$ model evaluation''; the thresholds were not tuned on the results. Pure scale failure is defined only among metric failures; the other categories are defined for every pair.}
\label{tab:2}
\small
\setlength{\tabcolsep}{4pt}
\begin{tabular}{lll}
\toprule
Name & Criterion & Meaning \\
\midrule
Metric failure & metric $\delta_1$ $<$ 0.5 & under half of the pixels within 1.25$\times$ of the truth \\
Pure scale failure & metric failure, scale-aligned $\delta_1$ $\geq$ 0.9 & structure intact; scale alone fails \\
Flawed structure & scale-aligned $\delta_1$ $<$ 0.9 & many pixels still off by more than 1.25$\times$ even after $s^{*}$ \\
Structure failure & scale-aligned $\delta_1$ $<$ 0.5 & subset of flawed; fails after scale correction \\
\bottomrule
\end{tabular}
\end{table}

For a model whose structure is exactly right and whose only error is scale, the Rel at scale factor $s^{*}$ is $\lvert 1/s^{*} - 1\rvert$; we use this as the reference curve in Figure~\hyperref[fig:2]{2}.

\textbf{Intervals.} All 95\% intervals are bootstrap intervals, with the resampling unit and number of repetitions stated in each case; paired comparisons reuse the same resamples. The failure-rate intervals of Section~\hyperref[sec:4.1]{4.1} and the HAMMER-pair intervals of Appendix~\hyperref[app:H]{H} resample recorded scenes (6 on DIODE, 7 on HAMMER, 13 on ETH3D, images on the other datasets); the size comparisons of Appendix~\hyperref[app:I]{I} resample images, treating each image as independent, while images from the same recorded scene are highly similar, so those intervals are too narrow (Section~\hyperref[sec:8]{8}).

\section{Where the Metric Error Comes From: Scale versus Structure}\label{sec:4}

\subsection{Main Results}\label{sec:4.1}

Table~\hyperref[tab:3]{3} has one row per model evaluation and groups the columns into metric, structure and scale, so each row shows how much of the metric error comes from scale.

\begin{table}[t]
\centering
\renewcommand{\thetable}{3}
\caption{Results of the 18 model evaluations on the 7 benchmarks (4,406 images). Every column is computed over all images within each dataset and then averaged over the 7 datasets with equal weight (Section~\hyperref[sec:3.3]{3.3}). The three scale columns are defined in Section~\hyperref[sec:3.3]{3.3}; a positive bias means overestimated depth. ``Fail.'' is the failure rate: under ``Metric'' the share of images with metric $\delta_1$ $<$ 0.5, under ``Structure'' the share with scale-aligned $\delta_1$ $<$ 0.5; ``Disp.'' the dispersion and ``Abs. err.'' the absolute error of $\ln s^{*}$. Per-dataset results are given in Tables~\hyperref[tab:F1]{F.1} to F.7 of Appendix~\hyperref[app:F]{F}. $\uparrow$/$\downarrow$: higher/lower is better; $\to 0$: closer to zero is better.}
\label{tab:3}
\footnotesize
\setlength{\tabcolsep}{2.6pt}
\begin{tabular*}{\linewidth}{@{\extracolsep{\fill}}llrrrrrrrrr@{}}
\toprule
& & \multicolumn{3}{c}{Metric} & \multicolumn{3}{c}{Structure} & \multicolumn{3}{c}{Scale ($s^{*}$)} \\ \cmidrule(lr){3-5}\cmidrule(lr){6-8}\cmidrule(lr){9-11} Model & K & $\delta_1\uparrow$ & Rel$\downarrow$ & Fail.$\downarrow$ & $\delta_1\uparrow$ & Rel$\downarrow$ & Fail.$\downarrow$ & Bias ($\to 0$) & Disp.$\downarrow$ & Abs.\ err.$\downarrow$ \\
\midrule
Metric3D-v2 & with K & 0.739 & 0.183 & 23.1\% & 0.978 & 0.045 & 0.0\% & ${+0.107}$ & 0.084 & 0.140 \\
\addlinespace[1.5pt]
DA3METRIC & with K & 0.865 & 0.128 & 9.0\% & 0.962 & 0.058 & 0.1\% & ${-0.004}$ & 0.106 & 0.096 \\
\addlinespace[1.5pt]
Depth Pro & w/o K & 0.543 & 0.276 & 44.0\% & 0.950 & 0.066 & 0.4\% & ${-0.162}$ & 0.305 & 0.298 \\
Depth Pro & with K & 0.661 & 0.224 & 31.4\% & 0.950 & 0.066 & 0.4\% & ${-0.137}$ & 0.280 & 0.250 \\
\addlinespace[1.5pt]
MA-DepthMap & with K & 0.846 & 0.126 & 10.8\% & 0.974 & 0.056 & 0.1\% & ${-0.042}$ & 0.109 & 0.108 \\
\addlinespace[1.5pt]
DepthVLM & with K & 0.728 & 0.295 & 20.4\% & 0.877 & 0.109 & 2.7\% & ${+0.092}$ & 0.149 & 0.174 \\
\addlinespace[1.5pt]
MoGe-2 & w/o K & 0.769 & 0.157 & 20.5\% & 0.972 & 0.046 & 0.1\% & ${-0.069}$ & 0.165 & 0.146 \\
MoGe-2 & with K & 0.877 & 0.136 & 8.1\% & 0.958 & 0.056 & 0.2\% & ${+0.032}$ & 0.133 & 0.094 \\
\addlinespace[1.5pt]
MoGe-3 & w/o K & 0.740 & 0.157 & 23.8\% & 0.972 & 0.046 & 0.1\% & ${-0.083}$ & 0.160 & 0.150 \\
MoGe-3 & with K & 0.875 & 0.131 & 7.7\% & 0.954 & 0.059 & 0.2\% & ${+0.033}$ & 0.125 & 0.090 \\
\addlinespace[1.5pt]
MA-PointMap & w/o K & 0.857 & 0.138 & 11.2\% & 0.973 & 0.045 & 0.1\% & ${-0.019}$ & 0.155 & 0.116 \\
MA-PointMap & with K & 0.875 & 0.136 & 9.0\% & 0.971 & 0.049 & 0.2\% & ${+0.001}$ & 0.129 & 0.106 \\
\addlinespace[1.5pt]
FoundationGeo & w/o K & 0.817 & 0.135 & 15.4\% & 0.970 & 0.048 & 0.2\% & ${-0.026}$ & 0.152 & 0.123 \\
FoundationGeo & with K & 0.860 & 0.138 & 9.7\% & 0.962 & 0.054 & 0.2\% & ${+0.040}$ & 0.136 & 0.107 \\
\addlinespace[1.5pt]
UniDepthV2 & w/o K & 0.753 & 0.213 & 21.6\% & 0.973 & 0.049 & 0.1\% & ${+0.064}$ & 0.180 & 0.163 \\
UniDepthV2 & with K & 0.826 & 0.185 & 14.4\% & 0.973 & 0.048 & 0.1\% & ${+0.074}$ & 0.145 & 0.131 \\
\addlinespace[1.5pt]
UniK3D & w/o K & 0.818 & 0.161 & 15.0\% & 0.974 & 0.048 & 0.1\% & ${+0.046}$ & 0.149 & 0.127 \\
UniK3D & with K & 0.874 & 0.134 & 9.8\% & 0.974 & 0.047 & 0.1\% & ${+0.047}$ & 0.111 & 0.097 \\
\bottomrule
\end{tabular*}
\end{table}

The ranking by metric $\delta_1$ depends on the condition. With intrinsics, MoGe-2 (\hyperref[tab:E1]{0.877}), MoGe-3 (\hyperref[tab:3]{0.875}), MA-PointMap (\hyperref[tab:3]{0.875}) and UniK3D (\hyperref[tab:E1]{0.874}) are within 0.005 of each other; without intrinsics MA-PointMap leads (\hyperref[tab:3]{0.857}) and the MoGe models drop to 0.74 to 0.77. The mean of Metric3D-v2 is pulled down by DIODE: there it reaches only 0.02, with a scale factor almost constant at 0.68, the same result as reported in the MoGe-2 paper under the same protocol (Table~\hyperref[tab:E1]{E.1}); without DIODE its mean over the other 6 benchmarks is 0.859, against \hyperref[tab:3]{0.865} for MA-DepthMap (per-dataset results in Appendix~\hyperref[app:F]{F}).

\textbf{Intervals of the failure rate and a per-image best reference.} Resampling recorded scenes (2,000 times), the failure rate of MoGe-3 with K is \hyperref[tab:3]{7.7\%} (95\% interval 4.4\% to 11.3\%) and that of MoGe-2 with K \hyperref[tab:3]{8.1\%} (5.0\% to 11.4\%). The differences between the top evaluations lie within these intervals, so ``the best evaluation'' below only indicates a lower bound on the failure rate and is not used for ranking. Almost every image is solved by some evaluation: the best of the 18 per image gives an equally weighted failure rate of only \hyperref[tab:3]{0.2\%} (DIODE 0.5\%, HAMMER 1.2\%). But the successful evaluations differ between images. This per-image best reference is optimistic, since some images may be solved only because the overall bias of a model happens to fit (Section~\hyperref[sec:5.2]{5.2}). Only 13 images fail in all 18 evaluations (4 on DIODE, 9 on HAMMER), all of them among the 16 images of Section~\hyperref[sec:5.2]{5.2} that all 11 architectures fail when given intrinsics.

\begin{figure}[t]
\centering
\renewcommand{\thefigure}{2}

\includegraphics[width=1.00\linewidth]{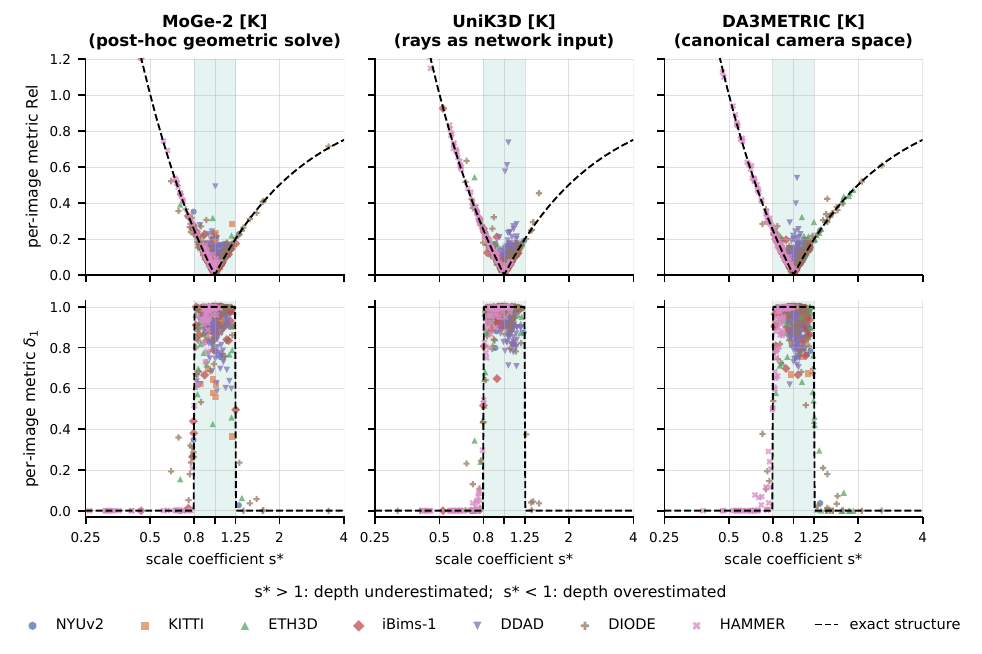}
\caption{Per-image metric error against the scale factor $s^{*}$. The three columns show, for three of the four intrinsics classes of Table~\hyperref[tab:1]{1}, the model with the highest metric $\delta_1$ (Table~\hyperref[tab:3]{3}), all with intrinsics: MoGe-2 (post-hoc geometric solve), UniK3D (rays as input) and DA3METRIC (canonical camera space, output conversion). The input-rescaling class contains only DepthVLM, whose structural error is large (Section~\hyperref[sec:4.3]{4.3}), and is not shown. Top row: metric Rel; bottom row: metric $\delta_1$. For each dataset, 100 images are drawn at random, the same for all three models; datasets are shown with different colours and markers. Dashed lines are the values for exactly correct structure: $\lvert 1/s^{*} - 1\rvert$ in the top row; 1 for $s^{*}$ in [0.8, 1.25] and 0 outside in the bottom row. The green band marks $s^{*}$ in [0.8, 1.25].}
\label{fig:2}
\end{figure}

\subsection{Most Metric Failures Are Scale Failures}\label{sec:4.2}

The four kinds of failure used in this section (metric failure, pure scale failure, flawed structure, structure failure) and their thresholds are summarised in Table~\hyperref[tab:2]{2}. Figure~\hyperref[fig:1]{1} shows the metric and scale-aligned $\delta_1$ and Rel of each model evaluation. Except for DepthVLM, the scale-aligned $\delta_1$ of every model evaluation lies between \hyperref[tab:3]{0.950} and \hyperref[tab:3]{0.978}, whatever its metric $\delta_1$, and the failure rates, \hyperref[tab:3]{7.7\%} to \hyperref[tab:3]{44.0\%} before alignment, fall to \hyperref[tab:4]{0.0\%} to \hyperref[tab:3]{0.4\%} after it (\hyperref[tab:3]{2.7\%} for DepthVLM). The scale-aligned Rel (\hyperref[tab:3]{0.045} to \hyperref[tab:3]{0.109}) is 2.2 to 4.3 times smaller than the metric Rel (median 3.0).

\textbf{Per image, most metric failures are scale failures.} The 18 model evaluations give 79,308 ``image $\times$ model evaluation'' pairs, of which 15,085 are metric failures. After scale correction with $s^{*}$, which is computed from all ground-truth pixels and is thus an upper bound for any practical correction, 98.3\% of metric failures no longer fail (98.8\% with datasets weighted equally) and only \hyperref[tab:4]{1.7\%} still fail. This share partly depends on the criterion: a scale error above 25\% fails the whole image, whereas a structural error counts as a failure only if half of the pixels stay outside a factor of 1.25 after per-image alignment; we call scale-aligned $\delta_1$ $<$ 0.5 a \textbf{structure failure}. For a finer split by structure (Table~\hyperref[tab:4]{4}), we call the structure \textbf{intact} when the scale-aligned $\delta_1$ is at least 0.9 and \textbf{flawed} otherwise. In \hyperref[tab:4]{78.8\%} of metric failures the structure is intact and the failure comes only from scale; we call these \textbf{pure scale failures}. Of the pure scale failures, \hyperref[tab:F6]{99.9\%} have $s^{*}$ outside [0.8, 1.25]; only 17 have a scale inside the range, all of them right at its edge. The remaining \hyperref[tab:4]{21.2\%} have flawed structure, and 94\% of them also have $s^{*}$ outside the range, i.e. a wrong scale, so 98.6\% of all metric failures have a wrong scale. The share of pure scale failures varies across models and datasets: only 17\% for DepthVLM against 67\% to 98\% for the other 17 model evaluations, and only 52\% on DDAD against 77\% to 95\% on the other 6 datasets. The log-error decomposition gives the same conclusion without any threshold: on metric failures, 93\% of the log error comes from the overall scale, and with failure judged by $\delta_2$ or $\delta_3$, 99.7\% and \hyperref[tab:F6]{99.9\%} of the failures no longer fail after scale correction (Appendix~\hyperref[app:B]{B}).

\begin{table}[t]
\centering
\renewcommand{\thetable}{4}
\caption{Metric failures (metric $\delta_1$ $<$ 0.5) by structure, 18 model evaluations pooled, per dataset. The categories are defined in Table~\hyperref[tab:2]{2}; columns 3 to 5 are shares of all metric failures of the dataset; the last column is the share of flawed-structure failures with $s^{*}$ outside [0.8, 1.25].}
\label{tab:4}
\small
\setlength{\tabcolsep}{4pt}
\begin{tabular}{lrrrrr}
\toprule
Dataset & \makecell{Metric\\failures} & \makecell{Pure scale\\($\geq$ 0.9)} & \makecell{Flawed\\structure ($<$ 0.9)} & \makecell{of which structure\\failure ($<$ 0.5)} & \makecell{Flawed structure\\with wrong scale} \\
\midrule
NYUv2 & 210 & 88.1\% & 11.9\% & 0.0\% & 100\% \\
KITTI & 1,015 & 95.0\% & 5.0\% & 0.3\% & 94\% \\
ETH3D & 1,362 & 86.7\% & 13.3\% & 1.3\% & 94\% \\
iBims-1 & 113 & 91.2\% & 8.8\% & 0.9\% & 90\% \\
DDAD & 2,313 & 51.9\% & 48.1\% & 1.7\% & 92\% \\
DIODE & 4,838 & 76.7\% & 23.3\% & 1.9\% & 91\% \\
HAMMER & 5,234 & 86.8\% & 13.2\% & 2.0\% & 100\% \\
\midrule
All & 15,085 & 78.8\% & 21.2\% & 1.7\% & 94\% \\
\bottomrule
\end{tabular}
\end{table}

\textbf{Metric $\delta_1$ is determined mainly by whether the scale falls in a narrow band.} Figure~\hyperref[fig:2]{2} plots the per-image metric error against $s^{*}$; the dashed line is the value for a prediction with exactly right structure and only a scale error. For such a prediction, $\delta_1$ equals 1 when $s^{*}$ is between 0.8 and 1.25 and 0 otherwise; the actual results are close to this step: for the three models of Figure~\hyperref[fig:2]{2}, 98\% to 99\% of failed images have $s^{*}$ outside [0.8, 1.25]. The metric Rel follows the reference curve $\lvert 1/s^{*} - 1\rvert$ monotonically, with a median per-image Spearman correlation of 0.89 over the 18 model evaluations (binned values in Appendix~\hyperref[app:G]{G}).

\textbf{Scale errors take different forms.} The last three columns of Table~\hyperref[tab:3]{3} give the three statistics of $s^{*}$. Scale errors of the same size can have different sources, which metric $\delta_1$ alone does not reveal. The direction of the bias depends on the model, the condition and the dataset (Figure~\hyperref[fig:F8]{F.8}): without intrinsics, MoGe-2 and MoGe-3 underestimate depth on 6 and 7 of the 7 datasets; with intrinsics they are nearly unbiased (|bias| $\leq$ 0.03) on every dataset except HAMMER, where both overestimate (+0.20 and +0.19); without intrinsics, UniDepthV2 and UniK3D overestimate mainly on HAMMER (+0.29 and +0.22). Dispersion and bias are different things: Metric3D-v2 has the smallest within-dataset dispersion averaged over the 7 datasets (\hyperref[tab:3]{0.084}), yet its bias varies across datasets: it is nearly unbiased on five of them ($-$0.03 to +0.06) but overestimates on DIODE (+0.40) and HAMMER (+0.28). The scale can also be widely dispersed when the bias is near zero: on DIODE, MoGe-2 with intrinsics is nearly unbiased (+0.005), yet its dispersion is \hyperref[tab:F6]{0.205} and its absolute error \hyperref[tab:F6]{0.143}, and it fails on \hyperref[tab:F6]{19.1\%} of the images (Table~\hyperref[tab:F6]{F.6}).

\subsection{Structure Failures Also Occur}\label{sec:4.3}

\textbf{Structure failures are rare, but not negligible.} Structure failures account for only 274 (0.35\%) of the 79,308 pairs, and only 14 of them have a passing metric $\delta_1$ (3 on DDAD, 11 on DIODE); under MoGe's solver there are about 300 such pairs (Appendix~\hyperref[app:C]{C}). The other 260 are also metric failures, only \hyperref[tab:G1]{1.7\%} of all metric failures, and 73\% of these 260 structure failures also have a wrong scale, which is also more dispersed: the standard deviation of $\ln s^{*}$ is 0.71 for these 260 pairs, against 0.50 for the 11,888 metric failures with intact structure. With thresholds between 0.3 and 0.7, this share stays a minority (\hyperref[tab:G1]{0.1\%} to \hyperref[tab:G1]{8.8\%}, Table~\hyperref[tab:G1]{G.1}). The pairs with flawed structure (scale-aligned $\delta_1$ $<$ 0.9), however, vary strongly across datasets: 3.4\% to 6.0\% on NYUv2, ETH3D, KITTI, HAMMER and iBims-1, 17\% on DIODE and 35\% on DDAD (Table~\hyperref[tab:G2]{G.2} gives the share of metric failures with intact structure per dataset). Metric3D-v2 and the two MetricAnything models, whose training data include DDAD, have fewer images with flawed structure on DDAD (12\% to 23\%), but DA3METRIC and DepthVLM, also trained on DDAD, do not (50\% and 78\%); the models not trained on DDAD have 22\% to 59\%, so the structural errors on DDAD are not only a matter of training overlap (Section~\hyperref[sec:3.1]{3.1}). Structure failures are most common among the metric failures of DIODE and HAMMER, yet only 1.9\% and 2.0\%; all 105 on HAMMER come from DepthVLM. DepthVLM is the exception: 41\% of its images have flawed structure (49\% pooled over all images) and only 17\% of its metric failures have intact structure, so its failures come mainly from structure.

\textbf{When structure is flawed, scale alone is not enough.} In Figure~\hyperref[fig:2]{2}, structural errors appear as points off the dashed line, mostly from DDAD; even when the scale is right, DDAD has a mean scale-aligned Rel of 0.107 to 0.116 for the three models of Figure~\hyperref[fig:2]{2}, against 0.028 to 0.078 on the other 6 datasets (Appendix~\hyperref[app:G]{G}). On these images the metric error cannot be explained by scale alone.

Pure structure failures, i.e. structure failures with a correct scale, number only 69 (0.5\% of the 15,085 metric failures). Metric metrics alone cannot reveal which kind a failure is, so we measure them separately.

\section{Shared and Model-Specific Failures: Two Parts of the Error in a Mean}\label{sec:5}

\subsection{Grading Failures by How Many Architectures Fail}\label{sec:5.1}

\begin{figure}[t]
\centering
\renewcommand{\thefigure}{3}

\includegraphics[width=0.55\linewidth]{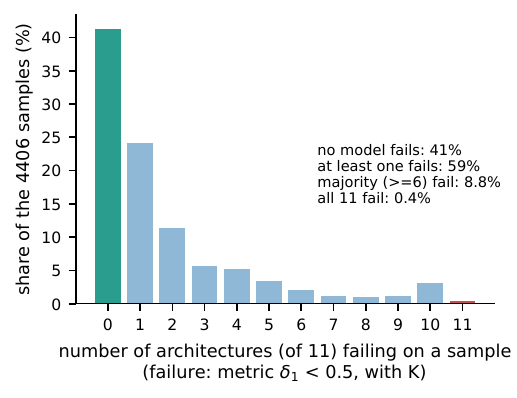}
\caption{Number of architectures failing on an image (out of 11, with intrinsics).}
\label{fig:3}
\end{figure}

This section uses only the results of the 11 architectures with intrinsics, the only condition common to all 11. For each image, we count the architectures with a metric failure on it (Section~\hyperref[sec:3.3]{3.3}) and denote the count by $k$. We grade failures into three tiers by $k$: $k \geq 6$ (most architectures fail) is a \textbf{shared failure}, $k$ between 1 and 2 is a \textbf{model-specific failure}, and $k$ between 3 and 5 is the \textbf{intermediate} tier. The 11 architectures have 7,750 failures in total (one failure is one architecture failing on one image). Grouped by the $k$ of the image, 42\%, 27\% and 32\% of the failures fall in the shared, specific and intermediate tiers. Figure~\hyperref[fig:3]{3} shows the distribution of $k$, Table~\hyperref[tab:5]{5} gives it per dataset, and Figures~\hyperref[fig:F10]{F.10} to F.13 show, for each dataset, the images failed by the most architectures. With thresholds $k \geq 5$ or $k \geq 7$, and $k \leq 1$, the qualitative conclusions below are unchanged (Appendix~\hyperref[app:G]{G}, Table~\hyperref[tab:G3]{G.3} also varies the failure threshold from 0.3 to 0.7).

\textbf{The 11 architectures are not independent.} MoGe-2, MoGe-3, MA-PointMap and FoundationGeo share the MoGe solver, UniDepthV2 and UniK3D come from the same authors, and MA-DepthMap has the network closest to Depth Pro and reuses its focal-length formula (\citealp{ma2026metricanything}; Appendix~\hyperref[app:D]{D}). Merging them into 6 families and requiring at least 4 failing families, 206 of the 218 HAMMER shared failures remain, but only 95 of the 157 on DIODE (Appendix~\hyperref[app:G]{G}). The HAMMER shared failures are therefore not repeated counts of related models; part of the DIODE ones are. Metric3D-v2 fails on 99.9\% of DIODE images; without it, DIODE has 108 shared failures.

\textbf{Shared failures are not only a matter of hard scenes.} Shared failures come partly from scenes: those on HAMMER come from only 4 of its 7 scenes (Section~\hyperref[sec:5.2]{5.2}). But scenes are not the whole cause. For each architecture, we randomly permute its failure flags within each recorded scene, keeping its failure rate in every scene, and compare the pairwise correlation ($\varphi$ coefficient) between the failure flags of the architectures: it is 0.23 on HAMMER and 0.24 on DIODE, against about 0 after permutation (95\% intervals within about $\pm$0.05), and still 0.20 on HAMMER when empty-table and object images are separated. Within a scene, most models thus tend to fail on the same images; on HAMMER, shared failures concentrate on specific images within a few scenes. We do not use the number of images with $k \geq 6$ for this comparison, because it depends on how high the failure rates are: when a hard scene has failure rates near 0.6, perfectly synchronised models can give fewer images with $k \geq 6$ than independent ones (Appendix~\hyperref[app:G]{G}).

\begin{table}[t]
\centering
\renewcommand{\thetable}{5}
\caption{Number of architectures failing on each image ($k$), per dataset (11 architectures with intrinsics; failure means metric $\delta_1$ $<$ 0.5). The first four columns split the images into four disjoint groups by $k$, summing to 100\%; the last column is the share failed by all 11 architectures. This table counts images; Table~\hyperref[tab:6]{6} counts failures.}
\label{tab:5}
\footnotesize
\setlength{\tabcolsep}{4pt}
\begin{tabular*}{\linewidth}{@{\extracolsep{\fill}}lrrrrrrr@{}}
\toprule
Dataset & Images & $k=0$ & Specific (1--2) & Intermediate (3--5) & Shared ($\geq$ 6) & Shared images & $k=11$ \\
\midrule
NYUv2 & 654 & 92.2\% & 6.6\% & 0.9\% & 0.3\% & 2 & 0\% \\
KITTI & 652 & 92.8\% & 7.1\% & 0.2\% & 0\% & 0 & 0\% \\
iBims-1 & 100 & 80.0\% & 16.0\% & 3.0\% & 1.0\% & 1 & 0\% \\
DDAD & 1,000 & 38.0\% & 59.8\% & 1.9\% & 0.3\% & 3 & 0\% \\
ETH3D & 454 & 16.7\% & 67.4\% & 14.5\% & 1.3\% & 6 & 0\% \\
DIODE & 771 & 0\% & 38.7\% & 41.0\% & 20.4\% & 157 & 0.8\% \\
HAMMER & 775 & 9.8\% & 33.0\% & 29.0\% & 28.1\% & 218 & 1.3\% \\
\midrule
All & 4,406 & 41.3\% & 35.5\% & 14.4\% & 8.8\% & 387 & 0.4\% \\
\bottomrule
\end{tabular*}
\end{table}

The intermediate tier is a transition from shared to model-specific failures: on the three datasets counted by scene, 92\% of its images come from scenes that also contain shared failures, and they are failed mostly by Metric3D-v2, DepthVLM, UniDepthV2 and Depth Pro.

\subsection{Shared Failures Concentrate in a Few Scenes}\label{sec:5.2}

\begin{figure}[p]
\centering
\renewcommand{\thefigure}{4}
\captionsetup{font=footnotesize}
\includegraphics[width=1.00\linewidth,height=0.84\textheight,keepaspectratio]{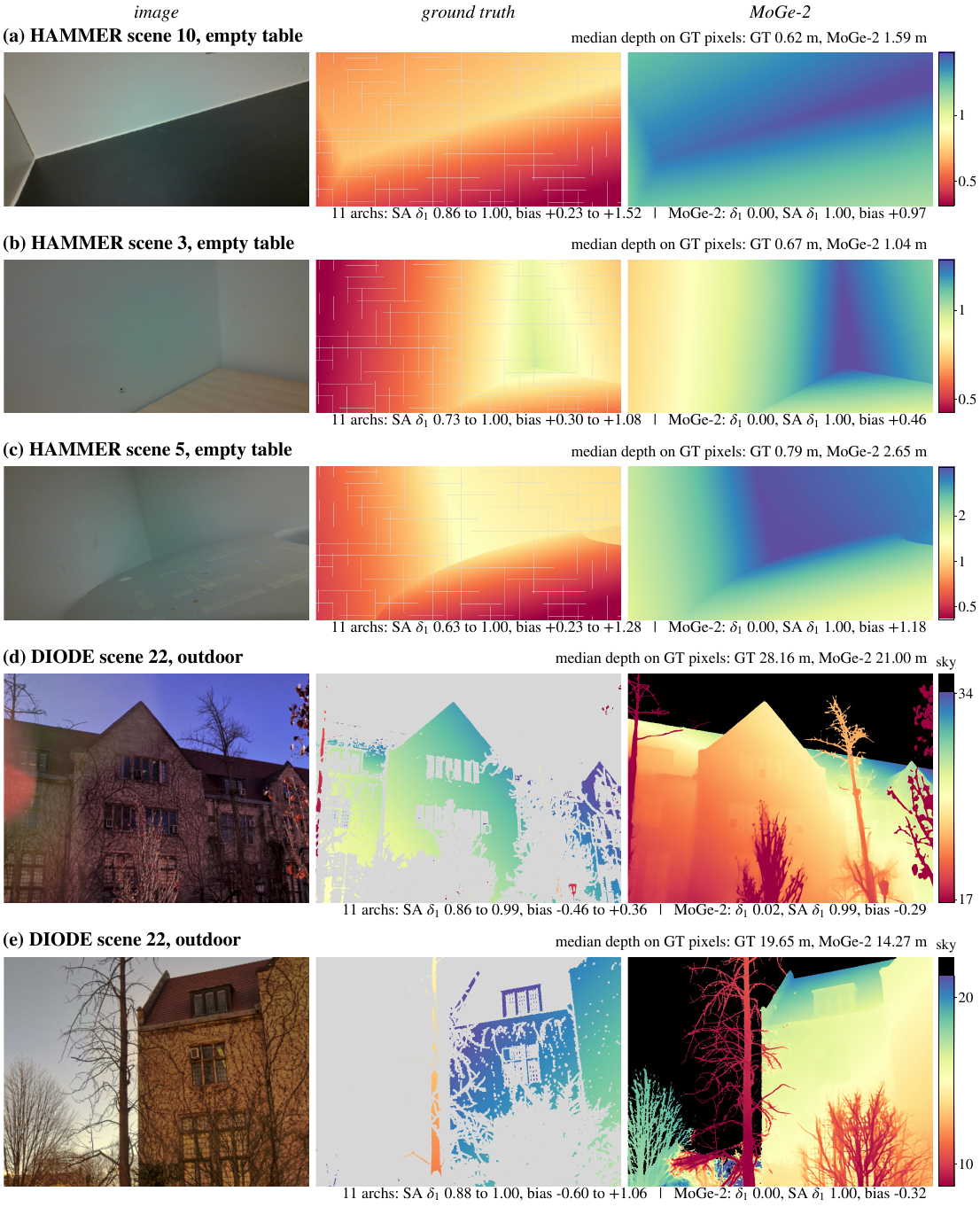}
\caption{Images failed by all 11 architectures with intact structure (pure scale failures, with intrinsics). (a) to (c) are the 3 HAMMER empty-table scenes, one image per scene; (d) and (e) come from DIODE. Each row shows the input image, the ground-truth depth and the prediction of MoGe-2 (with intrinsics); ground truth and prediction share a logarithmic colour scale (right, in metres), so a global colour shift indicates a scale error and a different spatial pattern of colours indicates a structural error. Grey in the ground truth marks pixels without ground truth; black in the prediction marks pixels that MoGe-2's own mask classifies as sky, i.e. the farthest region, labelled ``sky'' at the top of the colour bar and excluded from the colour scale. The top right of each row gives the median depth of the ground truth and of MoGe-2's prediction, both over pixels with valid ground truth; below each row are the ranges of the scale-aligned $\delta_1$ (SA) and of the bias ($\operatorname{mean}(\ln (1/s^{*}))$, positive: depth overestimated) of the 11 architectures on that image, then the metric $\delta_1$, scale-aligned $\delta_1$ and bias of MoGe-2.}
\label{fig:4}
\end{figure}

There are 387 shared-failure images, almost all from HAMMER (218) and DIODE (157), and none from KITTI. 90\% come from 9 recorded scenes: 4 of the 7 HAMMER scenes (scene10, scene3, scene5, scene9) and 5 of the 6 DIODE scenes. The four HAMMER scenes hold 510 of its 775 images; on DIODE, shared failures occur in every scene. ETH3D and DDAD have many images failed by at least one architecture (83\% and 62\%), but most are model-specific (Section~\hyperref[sec:5.3]{5.3}).

Shared failures mostly point in one direction: on a shared-failure image, 92\% of the failing architectures err in the same direction on average. On HAMMER almost all overestimate depth (99.5\%, with K), and 87\% of the failures have intact structure. On DIODE only 55\% overestimate depth, the direction varies with the scene, and only 65\% have intact structure, so structural error plays a larger role.

The 16 images failed by all 11 architectures come from 5 scenes. The 10 HAMMER images are all empty-table frames (Figure~\hyperref[fig:4]{4}(a) to (c) shows one per scene): on each, the median scale-aligned $\delta_1$ over the 11 architectures is 0.88 to 1.00, every bias is positive (+0.22 to +1.52), and MoGe-2 overestimates depth by a factor of 1.6 to 4.0, so they are mainly pure scale failures. Of the 6 DIODE images, two have intact structure (Figure~\hyperref[fig:4]{4}(d) and (e)) but the architectures err in different directions; the other 4 have flawed structure to different degrees: on two outdoor images most models get the relative distance between foreground trees and background buildings wrong, another outdoor image has only a slight structural error (scale-aligned $\delta_1$ of 0.65 to 0.95 over the 11 architectures), and the indoor one is a large white wall (Appendix~\hyperref[app:F]{F}, Figure~\hyperref[fig:F9]{F.9}).

Shared failures have one exception. Of the 165 HAMMER images failed by 9 or more architectures, MA-DepthMap gets 154 right, although its success rate over all HAMMER images is only 73\%. It is the only architecture that underestimates depth on HAMMER (bias -0.14), and it succeeds exactly on the images where the other architectures overestimate depth.

\textbf{Two kinds of shared failure on HAMMER.} By the labels of HAMMER, the shared failures fall into two kinds. The first are empty-table shots that lack objects whose size reveals the distance: 169 of the 218, including all 10 HAMMER images failed by all 11 architectures; the mean failure rate of the 11 architectures on the 371 empty-table images is 49\%. The second contain objects and are still failed by most models: 49 of the 404 images with objects, 40 of them from scene10; as on the empty table, they are almost all pure scale failures in one direction (99.5\% overestimate depth, 89\% with intact structure). Over the 404 images with objects, the mean failure rate of the 11 architectures is 28\%. Missing objects thus make scale harder to estimate (Section~\hyperref[sec:6.3]{6.3} tests this with paired images), but objects alone do not guarantee the right scale. DIODE has no such label; its shared failures contain cues such as building windows and floors.

\begin{table}[t]
\centering
\renewcommand{\thetable}{6}
\caption{Metric failures of each architecture (with intrinsics) by the number of architectures $k$ failing on the image. ``Main location of specific failures'' is given only for the 4 models with many specific failures: the two datasets with the most specific failures, the share of the model's specific failures on each and the bias there (positive: overestimated).}
\label{tab:6}
\footnotesize
\setlength{\tabcolsep}{4pt}
\begin{tabularx}{\linewidth}{@{}lrrrr>{\raggedright\arraybackslash}X@{}}
\toprule
Model & Failures & $k\leq 2$ (specific) & $k$ = 3 to 5 & $k\geq 6$ (shared) & Main location of specific failures (share, bias) \\
\midrule
Metric3D-v2 & 1,220 & 30\% & 39\% & 31\% & DIODE (81\%, +0.39); HAMMER (12\%, +0.28) \\
DA3METRIC & 454 & 3\% & 26\% & 71\% & — \\
Depth Pro & 1,551 & 59\% & 20\% & 21\% & DDAD (58\%, -0.61); ETH3D (25\%, -0.81) \\
MA-DepthMap & 506 & 41\% & 43\% & 16\% & HAMMER (52\%, -0.27); ETH3D (23\%, -0.18) \\
DepthVLM & 1,116 & 32\% & 36\% & 31\% & HAMMER (52\%, +0.62); DDAD (36\%, -0.24) \\
MoGe-2 & 384 & 3\% & 18\% & 79\% & — \\
MoGe-3 & 376 & 4\% & 22\% & 74\% & — \\
MA-PointMap & 416 & 9\% & 18\% & 73\% & — \\
FoundationGeo & 487 & 13\% & 27\% & 60\% & — \\
UniDepthV2 & 732 & 6\% & 49\% & 44\% & — \\
UniK3D & 508 & 5\% & 37\% & 58\% & — \\
\bottomrule
\end{tabularx}
\end{table}

\textbf{The DIODE shared failures are not unsolvable; those on HAMMER must be read separately.} Of the 387 shared failures, only 16 are failed by all 11 architectures. On DIODE, at least one architecture succeeds on 151 of the 157, and on 148 when MA-DepthMap is left out; on images failed by 6 to 10 architectures, the median number of succeeding architectures is 4. These successes are spread over 10 architectures, and only 22\% of them come from an architecture whose bias points opposite to that of the failing majority, unlike HAMMER, where MA-DepthMap alone accounts for half of the successes. These DIODE images are therefore hard for current models but solved by several of them, and reflect a shortcoming common to this generation of models. HAMMER differs: at least one architecture succeeds on 208 of the 218, but on 123 of them only MA-DepthMap does; without it, 85 remain, and the median number of succeeding architectures is only 1. As explained above (``Shared failures have one exception'', Section~\hyperref[sec:5.2]{5.2}), MA-DepthMap's successes do not show that it obtains the scale from the image. Whether the scale of the HAMMER empty-table images can be recovered from the image is therefore left open by these results. Section~\hyperref[sec:6]{6} uses comparisons that change only the content to test which image content relates to the scale error.

\subsection{Model-Specific Failures Are Systematic Biases}\label{sec:5.3}

Model-specific failures come mainly from 4 models. Table~\hyperref[tab:6]{6} lists, per architecture, on which kind of image each failure falls. Of all failures, 27\% are model-specific, of which 24 percentage points come from Depth Pro, Metric3D-v2, DepthVLM and MA-DepthMap, and only 3 percentage points from the other 7 architectures together. These 7 architectures have almost no failures of their own: 94\% of their failures fall on images failed by at least 3 architectures, and 58\% to 79\% on shared-failure images (44\% for UniDepthV2).

The specific failures of these 4 models concentrate on one or two datasets, with a constant direction within each: the 761 of Depth Pro on DDAD and ETH3D almost all underestimate depth (on ETH3D 91\% have intact structure, but on DDAD only 40\% do, so Depth Pro's DDAD failures combine a consistent scale bias with structural error), the 298 of Metric3D-v2 on DIODE and the 189 of DepthVLM on HAMMER all overestimate it, and the 108 of MA-DepthMap on HAMMER all underestimate it. They are thus systematic scale biases rather than random errors, and cannot be found from the 7-dataset mean: Depth Pro has the lowest mean metric $\delta_1$, but on NYUv2 and HAMMER it is on par with most models (Appendix~\hyperref[app:F]{F}). The exception is DDAD: only 8\% of the 130 specific failures of DepthVLM and 40\% of the 532 of Depth Pro have intact structure; DDAD is in the training data of DepthVLM.

All 4 models use a canonical camera space (Table~\hyperref[tab:1]{1}), but DA3METRIC, of the same class, has only 3\% specific failures, so we report this only as an observation and infer no causal relation.

\subsection{How Shared Failures and Other Images Affect the Mean and Ranking}\label{sec:5.4}

Readers use a mean for two things: to judge how accurate a model is overall, and to rank models. Table~\hyperref[tab:7]{7} splits the 7-benchmark mean error of each model into four parts by the $k$ of the images (the three failure tiers, and the shortfall of $\delta_1$ below 1 on images with $k$ = 0); the parts sum to the error in Table~\hyperref[tab:3]{3}.

\begin{table}[t]
\centering
\renewcommand{\thetable}{7}
\caption{Mean error of each model by $k$ tier (11 architectures with intrinsics). The columns ``$k$ = 0'', ``Specific'', ``Intermediate'' and ``Shared'' give the share of the metric $\delta_1$ error $1 - \delta_1$ falling on each tier, summing to 100\%; images with $k$ = 0 have no failing architecture, but the part of $\delta_1$ below 1 still counts as error. ``Without shared'' is the mean recomputed on images with $k < 6$ only. Weighting datasets equally, the four tiers $k$ = 0, specific, intermediate and shared make up 47.1\%, 32.6\%, 12.9\% and 7.3\% of all images.}
\label{tab:7}
\scriptsize
\setlength{\tabcolsep}{2.6pt}
\begin{tabular*}{\linewidth}{@{\extracolsep{\fill}}lrrrrrrrrrrrrr@{}}
\toprule
& \multicolumn{4}{c}{Share of $1-\delta_1$ error by tier} & \multicolumn{4}{c}{Metric $\delta_1$} & \multicolumn{5}{c}{Metric Rel} \\ \cmidrule(lr){2-5}\cmidrule(lr){6-9}\cmidrule(lr){10-14} Model & $k=0$ & Spec. & Interm. & Shared & All & Rank & w/o shared & Rank & All & Rank & On shared & w/o shared & Rank \\
\midrule
Metric3D-v2 & 7\% & 32\% & 35\% & 27\% & 0.739 & 9 & 0.763 & 9 & 0.183 & 8 & 23\% & 0.165 & 9 \\
DA3METRIC & 13\% & 20\% & 24\% & 42\% & 0.865 & 5 & 0.914 & 5 & 0.128 & 2 & 27\% & 0.103 & 4 \\
Depth Pro & 10\% & 53\% & 19\% & 17\% & 0.661 & 11 & 0.700 & 11 & 0.224 & 10 & 19\% & 0.193 & 10 \\
MA-DepthMap & 14\% & 45\% & 29\% & 12\% & 0.846 & 7 & 0.842 & 8 & 0.126 & 1 & 11\% & 0.126 & 7 \\
DepthVLM & 18\% & 32\% & 27\% & 23\% & 0.728 & 10 & 0.747 & 10 & 0.295 & 11 & 31\% & 0.247 & 11 \\
MoGe-2 & 18\% & 18\% & 18\% & 46\% & 0.877 & 1 & 0.928 & 1 & 0.136 & 5 & 34\% & 0.097 & 1 \\
MoGe-3 & 20\% & 19\% & 19\% & 42\% & 0.875 & 2 & 0.921 & 3 & 0.131 & 3 & 31\% & 0.098 & 2 \\
MA-PointMap & 16\% & 19\% & 20\% & 45\% & 0.875 & 3 & 0.926 & 2 & 0.136 & 6 & 33\% & 0.099 & 3 \\
FoundationGeo & 15\% & 25\% & 23\% & 38\% & 0.860 & 6 & 0.901 & 6 & 0.138 & 7 & 28\% & 0.111 & 6 \\
UniDepthV2 & 8\% & 16\% & 41\% & 35\% & 0.826 & 8 & 0.864 & 7 & 0.185 & 9 & 30\% & 0.148 & 8 \\
UniK3D & 10\% & 16\% & 31\% & 43\% & 0.874 & 4 & 0.915 & 4 & 0.134 & 4 & 30\% & 0.104 & 5 \\
\bottomrule
\end{tabular*}
\end{table}

\textbf{Shared failures set the level of the error.} Shared-failure images make up 7.3\% of the data with datasets weighted equally (\hyperref[tab:5]{8.8\%} pooled over all images, Table~\hyperref[tab:5]{5}). They account for 38\% to 46\% of the $1-\delta_1$ error of the top 6 models by metric $\delta_1$ (MoGe-2, MoGe-3, MA-PointMap, UniK3D, DA3METRIC, FoundationGeo) and 27\% to 34\% of their Rel error, against 14\% to 18\% of the $1-\delta_1$ error for the same number of random images drawn from the same datasets (2.3 to 3.2 times). On individual datasets the effect is stronger: 89\% of the $1-\delta_1$ error and 80\% of the Rel of MoGe-2 on HAMMER come from these images, and 63\% and 40\% on DIODE. This is a circular estimate, since the shared-failure images are selected by these models' own failures. A held-out test removes the circularity: for each model, we consider only the architectures outside its own family, rank the images by how many of them fail, and take per dataset as many images as there are original shared failures (rule in Appendix~\hyperref[app:G]{G}); these images account for 34\% to 40\% of its $1-\delta_1$ error and overlap the original shared failures by 81\% to 93\%, well above the 14\% to 18\% of random images. Counting by 6 families gives the same conclusion (Appendix~\hyperref[app:G]{G}). Without them, the $1-\delta_1$ error of the top 6 falls by 29\% to 42\% and their Rel by 19\% to 29\% (Table~\hyperref[tab:7]{7}), so how low these means are is largely set by images on which most models fail together.

\textbf{The ranking is decided by images other than shared failures, but the effect of shared failures on the ranking depends on the metric.} Under metric $\delta_1$, removing the shared-failure images leaves the ranking almost unchanged, with only two pairs of adjacent ranks swapped; the errors of the models at the bottom come mainly from specific failures (Depth Pro, MA-DepthMap) or the intermediate tier (UniDepthV2). Under metric Rel, 9 of the 11 models change rank: MA-DepthMap drops from 1st to 7th and MoGe-2 rises from 5th to 1st. The reason is that $\delta_1$ is bounded: it costs a failed image at most 1, and most models have a low $\delta_1$ on these images, which therefore cannot separate them; Rel is unbounded, so an image with a scale off by a factor of two can outweigh many normal images; MA-DepthMap gets right exactly the HAMMER images that the other models fail together (Section~\hyperref[sec:5.2]{5.2}). Under $\delta_1$ this agrees with \citet{meding2022trivial} on ImageNet: model differences show only on the images that not all models get right or all get wrong; under Rel the shared-failure images are not failed by every model (MA-DepthMap gets most HAMMER shared failures right), so they still separate models.

A model's mean can thus be read as two parts: the part on shared-failure images lowers the means of all models, and the part on the remaining images decides its gap to other models. We therefore recommend reporting the shared-failure images alongside the mean, or the mean stratified by $k$, and, when ranking with an unbounded metric such as Rel, stating whether the ranking depends on these images.

\section{What in the Image Relates to the Scale Error}\label{sec:6}

The 7 benchmarks differ at once in scenes, sensors, depth ranges and training overlap, so we use two sources that change the content and keep everything else as fixed as possible (Section~\hyperref[sec:6.1]{6.1}). No evaluated model lists TUM RGB-D in its training data.

\subsection{Data}\label{sec:6.1}

\textbf{HAMMER, paired.} The HAMMER camera is mounted on a robot arm, and each trajectory is replayed twice by the arm, once with objects on the table and once without; the two passes have exactly the same viewpoints and background \citep{jung2023hammer}. We pair the 371 empty-table images with their counterparts with objects by file name and compare each pair, so that differences in viewpoint, camera distance and background cancel in the paired difference (example and background check in Appendix~\hyperref[app:H]{H}).

\textbf{TUM-Family.} The Freiburg 3 structure and texture sequences of TUM RGB-D \citep{sturm2012benchmark} vary, with the same sensor, the geometry (a flat board or a zigzag board, ``structure'') and the surface appearance (plain or covered with posters, ``texture''), each at two camera distances (near and far, with median depths of about 1.0 m and 2.0 m). Each sequence is a separate handheld recording, so viewpoint and trajectory also differ between sequences; in the two sequences with structure and texture, posters are also laid on the floor in front of the board. We use all 8 sequences and sample frames with rules fixed in advance that depend only on poses and depth coverage, giving 514 images (30 to 131 per sequence), with the official intrinsics. Appendix~\hyperref[app:H]{H} gives the sampling rules, a frame of each sequence and our reprojection check of the depth against the poses.

\textbf{TUM-Ref.} As a control recorded with the same camera, intrinsics and ground-truth pipeline, we use an office with desks, chairs, monitors and a teddy bear (\texttt{freiburg3\_long\_office\_household}, 215 images selected with the same rules). Its median valid depth is 1.67 m, with 1st to 99th percentiles of 0.61 to 7.49 m, a wider range than the family sequences. It differs from the family sequences in content, depth range and layout, so it shows only that the TUM-Family failures are not caused by the sensor, the intrinsics or the ground truth.

\subsection{The Error Direction Mainly Follows the Sequence, Not the Model}\label{sec:6.2}

Unlike the scenes of Section~\hyperref[sec:5]{5}, TUM-Family and TUM-Ref share one camera and its intrinsics, so the error direction can be compared across content. We use the 11 models with ground-truth intrinsics; the ``family average'' averages over the frames of each sequence, then over the 8 sequences.

\textbf{The error is in scale, not in structure.} On TUM-Family, the family-average metric $\delta_1$ of the 11 models is only 0.16 to 0.40, with failure rates of 60\% to 84\%; after scale correction, the scale-aligned $\delta_1$ of the 10 models other than DepthVLM (0.901) lies between 0.976 and 0.992. The failures here are thus almost all pure scale failures: the models recover the structure of the board correctly and only place it at the wrong distance. On TUM-Ref, the same 11 models reach a metric $\delta_1$ of 0.84 to 0.96, with biases between $-$0.12 and +0.02, all within the range that passes $\delta_1$ ($\pm$ln 1.25 $\approx$ $\pm$0.22).

\begin{figure}[t]
\centering
\renewcommand{\thefigure}{5}

\includegraphics[width=1.00\linewidth]{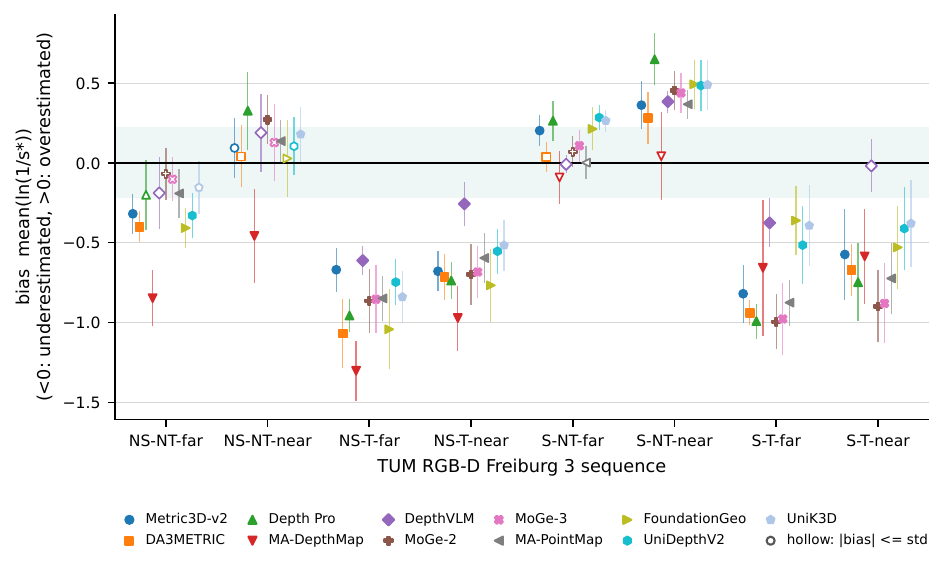}
\caption{Bias $\operatorname{mean}(\ln(1/s^{*}))$ of the 11 models (with ground-truth intrinsics) on each TUM-Family sequence; negative means underestimated and positive overestimated depth. Vertical bars show $\pm$ dispersion $\operatorname{std}(\ln s^{*})$, computed over all frames of the sequence. Each model has its own colour and marker; filled markers are direction-stable (|bias| $>$ dispersion) and hollow markers direction-uncertain. The shaded band is $\pm$ln 1.25, the range within which a structurally perfect prediction passes $\delta_1$. S = structure, NS = no structure, T = texture, NT = no texture.}
\label{fig:5}
\end{figure}

\textbf{The error direction is set by the sequence, not by the model.} We call a model \textbf{direction-stable} on a sequence when the absolute value of its bias exceeds its dispersion, and \textbf{direction-uncertain} otherwise (both computed over all frames of the sequence). Of the 88 cells of Figure~\hyperref[fig:5]{5}, 70 are direction-stable. On the 4 sequences with printed texture, all 44 cells underestimate depth, 43 of them direction-stable (median biases of $-$0.59 to $-$0.85); on the near sequence with structure and without texture, 10 of the 11 models stably overestimate depth (median +0.44). The two NS-NT sequences show only a flat, plain surface whose image contains no scale information; the bias there reflects the model's prior and is not used to compare models.

We decompose the 11 $\times$ 8 bias matrix (rows are models, columns are sequences) into a sequence main effect, a model main effect and an interaction, which account for 84\%, 7\% and 9\% of the total variance; the sequence effect includes surface, distance and trajectory together, and is far larger than the model effect. This holds for most models: the ordering of a model's biases over the 8 sequences correlates with that of the mean bias of the other 10 models with a median Spearman correlation of 0.93. Between sequences, the biases of most models also shift in the same direction (Appendix~\hyperref[app:H]{H}). The shared failures of Section~\hyperref[sec:5]{5} are the same phenomenon: on the same scenes, the scale of most models errs in the same direction.

\textbf{TUM-Family is a diagnostic set, not a ranking benchmark.} Model rankings on two halves of its 8 sequences have a median Spearman correlation of -0.03, and it covers only short distances (99th percentile about 3.9 m), so we use it only for directions and cue comparisons.

\subsection{Adding Objects (HAMMER)}\label{sec:6.3}

Objects of known size are a classic cue to absolute scale, and the HAMMER pairs test it directly (Section~\hyperref[sec:5.2]{5.2}). We compare the absolute log scale error $\lvert\ln s^{*}\rvert$ of each model, taking the change as ``with objects minus empty table'' (negative means the error shrinks), averaged over the 371 pairs. With only 6 scenes, we use an exact scene-level sign-flip test (64 sign combinations); per-model bootstrap intervals are given in Appendix~\hyperref[app:H]{H}.

\begin{figure}[t]
\centering
\renewcommand{\thefigure}{6}

\includegraphics[width=0.95\linewidth]{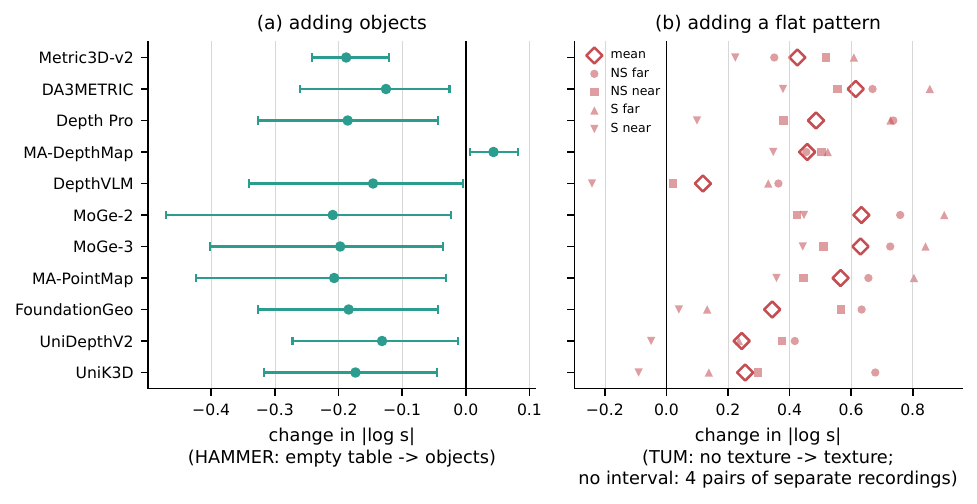}
\caption{Change in the absolute log scale error $\lvert\ln s^{*}\rvert$ when (a) adding objects (HAMMER, 371 paired images; error bars are 95\% bootstrap intervals resampling the 6 scenes, indicative only with 6 scenes; the test is in Section~\hyperref[sec:6.3]{6.3} and Appendix~\hyperref[app:H]{H}) and (b) adding a planar printed pattern (TUM, four paired sequences; small dots are individual pairs, diamonds the mean over the four pairs; the four pairs are separate recordings, so no interval is given); negative values mean that the scale error shrinks when the cue is added.}
\label{fig:6}
\end{figure}

Adding objects reduces the scale error of 10 of the 11 models (Figure~\hyperref[fig:6]{6}a); the mean paired change of these ten models is between $-$0.21 and $-$0.13. The change averaged over the 11 models is negative in 5 of the 6 scenes (scene-level sign-flip test, one-sided p = 0.03); individually, 5 models reach p $\leq$ 0.05 (Table~\hyperref[tab:H1]{H.1}). The empty-table images are dominated by bare walls and table tops, so ``adding objects'' also adds the texture and structure that come with them. Solving the scale on the shared background pixels only (about 73\%) gives almost the same change (Table~\hyperref[tab:H1]{H.1}), so objects change the scale of the whole scene, not only the depth of the objects. Failure rates fall accordingly: averaged over the 11 models on the 371 pairs, 30\% with objects against 49\% without. With objects, the bias of every model decreases by 0.12 to 0.21. The effect is largest where shared failures concentrate (scene5, scene3, scene10).

MA-DepthMap is the exception (+0.04): it underestimates depth on HAMMER (Section~\hyperref[sec:5.2]{5.2}), and objects make its depths smaller still (bias from $-$0.07 to $-$0.20), so its error grows. Objects thus act in the same direction for all models; whether the error grows or shrinks depends on the original bias.

\subsection{Adding a Printed Pattern (TUM)}\label{sec:6.4}

A planar printed pattern differs from an object. It adds texture and thus more visual information than a plain surface, and its text and graphics may carry some scale information; but a printed pattern can be of any physical size, so it carries much weaker scale information than real objects and can even mislead the scale estimate. The textured and untextured TUM-Family sequences share geometry and distance class and differ in whether the surface is printed; they are separate recordings (Section~\hyperref[sec:6.1]{6.1}), but can test whether current models obtain scale from such content. In the group without structure, the texture is a conference poster laid on the plane; in the group with structure, it comes from the coloured wrapping of the board and the posters on the floor in front of it. The change is again taken as $\lvert\ln s^{*}\rvert$ ``with pattern minus without pattern''.

There are four pairs (two geometries, each at near and far distance). On the sequences with a printed pattern, the mean scale error of all 11 models is larger (Figure~\hyperref[fig:6]{6}b). Here the change of each pair is the difference of the means of $\lvert\ln s^{*}\rvert$ over all frames of the two sequences, and the ``mean'' of each model is the average over the four pairs; 41 of the 44 ``model $\times$ pair'' cells are larger. The 3 cells that are smaller are all on the near pair with structure: one each of DepthVLM, UniDepthV2 and UniK3D.

HAMMER pairs share a pose and TUM pairs are different recordings, so we compare only directions: objects reduce the scale error and patterns increase it; more content alone does not give a better scale.

\textbf{Printed patterns barely affect structure.} On the same four pairs, the scale-aligned $\delta_1$ changes little: in the two groups without structure, adding posters leaves structure unchanged or better (median +0.002 at far and +0.057 at near distance); in the two groups with structure it gets slightly worse (medians $-$0.009 and $-$0.007), and apart from a drop of 0.12 for DepthVLM on the near pair, no drop exceeds 0.04. The error caused by printed patterns is thus almost entirely in scale.

These posters were not optimised against any model, yet on the 4 textured sequences the 11 models almost all stably underestimate depth (43 of 44 cells, Section~\hyperref[sec:6.2]{6.2}), corresponding to underestimating depth by a factor of about 1.8 to 2.3. The untextured sequences show no such consistent underestimation: on NS-NT-far all 11 models also underestimate depth, but with a median of only $-$0.20, and on the other 3 untextured sequences most models overestimate depth. Prior work changes the local depth of monocular depth models with optimised adversarial patches \citep{cheng2022physical}; our results suggest that an unoptimised planar pattern may already be associated with a shift in the global scale of metric models, while structure is unaffected and scale-aligned metrics alone would not reveal it. Confirming causality requires comparing the same scene from the same viewpoint with and without the pattern.

\section{Ground-Truth Intrinsics and Model Size}\label{sec:7}

\begin{figure}[!ht]
\centering
\renewcommand{\thefigure}{7}

\includegraphics[width=0.90\linewidth]{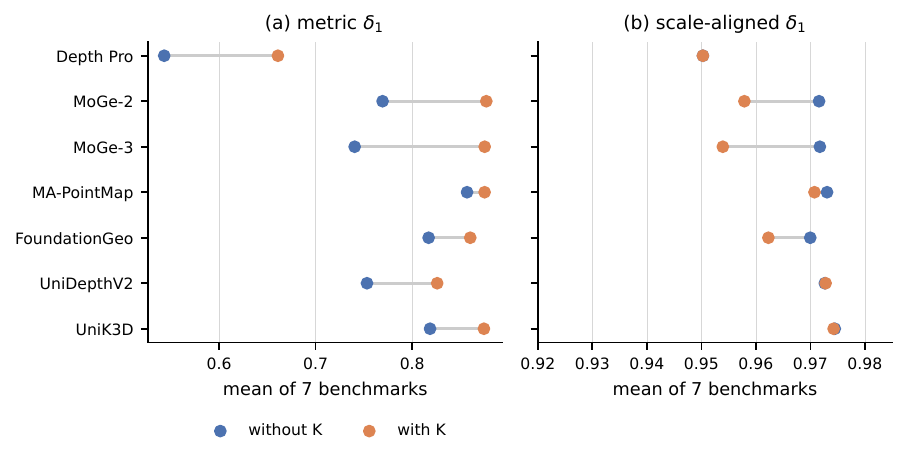}
\caption{Effect of ground-truth intrinsics. (a) Metric $\delta_1$; (b) scale-aligned $\delta_1$.}
\label{fig:7}
\end{figure}

\textbf{Ground-truth intrinsics raise metric accuracy but do not improve structure.} Supplying ground-truth intrinsics raises the 7-benchmark mean metric $\delta_1$ of all 7 architectures that can be evaluated without them, by +0.018 (MA-PointMap) to +0.135 (MoGe-3) (Figure~\hyperref[fig:7]{7}a), but does not raise the scale-aligned $\delta_1$. For the four models sharing the MoGe solver it \textbf{drops} (by 0.014 for MoGe-2, 0.018 for MoGe-3, \hyperref[tab:F1]{0.008} for FoundationGeo and 0.002 for MA-PointMap), while the ray-input UniDepthV2 and UniK3D are unchanged to three decimals. This is consistent with the design: in the post-hoc solve the intrinsics affect the ratios between pixel depths only through the shift $t$, and with the focal length fixed, $t$ must also satisfy the projection constraints under the true focal length, so the structure may be fitted less well (Appendix~\hyperref[app:D]{D}). The scale can worsen too: on HAMMER, the post-hoc solve models have an overall bias near 0 ($-$0.05 to +0.10) without intrinsics and about +0.2 with them. On single datasets the metric $\delta_1$ can even drop: on HAMMER for MA-PointMap, FoundationGeo and also the ray-input UniK3D, and on NYUv2 for MA-PointMap (Tables~\hyperref[tab:F1]{F.1}, F.7). Ground-truth intrinsics are thus no upper bound on metric accuracy, especially for post-hoc solve models.

\textbf{In $\delta_1$ points, the gain from size is mainly in scale, but structure errors shrink by a similar fraction.} Within the five families with several public sizes (Metric3D-v2, MoGe-2, MoGe-3, UniDepthV2 and UniK3D), we compare adjacent sizes in 16 steps (Table~\hyperref[tab:I1]{I.1}, Figure~\hyperref[fig:I1]{I.1}). The scale-aligned $\delta_1$ rises at every step, but by at most \hyperref[tab:F1]{0.022}, because it is already above 0.92 at the smallest size; in 14 steps the metric $\delta_1$ rises as well, by 2.7 to 12.0 times the scale-aligned gain (median 4.9 times). In two steps (Appendix~\hyperref[app:I]{I}), the metric $\delta_1$ drops instead even though the scale-aligned $\delta_1$ rises. Measured with unbounded errors, from the smallest to the largest size the scale-aligned Rel falls by 26\% to 29\% and the absolute scale error $\operatorname{mean}(\lvert\ln s^{*}\rvert)$ by 18\% to 47\% in four of the five families; MoGe-3 (ViT-L to ViT-G) improves structure by 11\% but not scale (+14\% with intrinsics; Table~\hyperref[tab:I5]{I.5}).

\textbf{Failures remain substantial with both intrinsics and the largest size.} Taking each architecture with intrinsics at its largest public size (for MoGe-2 ViT-L, both checkpoints), the failure rate over the 7 benchmarks is \hyperref[tab:I3]{8.0\%} to \hyperref[tab:I3]{31.4\%}, on DIODE \hyperref[tab:I3]{16.3\%} to \hyperref[tab:I3]{98.6\%} and on HAMMER \hyperref[tab:I3]{23.9\%} to \hyperref[tab:I3]{81.5\%} (Table~\hyperref[tab:I3]{I.3}).

\textbf{The shared-failure images stay hard when the family under test takes no part in selecting them.} The other sizes and the runs without intrinsics are close relatives of the 11 evaluations that selected the shared failures, so their failure rates on these images are biased by the selection (Table~\hyperref[tab:I2]{I.2}); we therefore rebuild the shared set without the family under test (Table~\hyperref[tab:I4]{I.4}). On HAMMER, every size of these families, with or without intrinsics, fails on 43\% to 100\% of the images so selected, against 1\% to 78\% of the other images, and the largest size is not clearly better. On DIODE the selected images remain harder than the rest except for Metric3D-v2 and MoGe-3 without intrinsics; the latter shows that the DIODE shared failures of the MoGe family depend partly on its own models having selected them. For MoGe-2, UniDepthV2 and UniK3D, larger sizes lower the DIODE rate (Table~\hyperref[tab:I4]{I.4}). The FoundationGeo paper \citep{liu2026foundationgeo} attributes the remaining error on DDAD, ETH3D, HAMMER and DIODE mainly to out-of-distribution focal lengths. For the canonical-camera and ray-input classes, the true focal length enters the scale conversion directly or conditions the network, so the role of a wrong focal estimate can be tested: these models still fail heavily with intrinsics on the hard images selected by other architectures (on HAMMER 91\% for DA3METRIC and 91\% for UniK3D, against 9\% and 27\% on the other images; Appendix~\hyperref[app:G]{G} and Table~\hyperref[tab:I4]{I.4}), so for them a wrong focal estimate cannot explain the failures; whether limited focal coverage in training contributes is not tested here. The test does not apply to the post-hoc solve class (including FoundationGeo): its scale head acts after the solve, independently of the intrinsics (Appendix~\hyperref[app:D]{D}).

\section{Limitations and Reproducibility}\label{sec:8}

\paragraph{Limitations}

\begin{itemize}
\item \emph{Independence.} The images of DIODE, HAMMER and ETH3D come from a limited number of recorded scenes (6, 7 and 13), with overlapping viewpoints within a scene, so the images are not independent. Section~\hyperref[sec:6.3]{6.3} has only 6 scenes and therefore uses a scene-level sign-flip test; the per-model intervals (Appendix~\hyperref[app:H]{H}) are themselves unstable; those of Appendix~\hyperref[app:I]{I} treat each image as an independent sample (Section~\hyperref[sec:3.3]{3.3}) and are therefore too narrow; differences of a few hundredths between models should not be read as a ranking. The shared failures of Section~\hyperref[sec:5]{5} come from 9 recorded scenes (Section~\hyperref[sec:5.2]{5.2}), so their share in the mean error is set by these scenes; the 11 architectures are not independent either (Appendix~\hyperref[app:G]{G}).
\item \emph{Mechanism.} This paper measures where scale errors occur and in which direction, not why models err this way internally (e.g. on the printed-texture sequences, or MA-DepthMap on HAMMER). Training data may contribute, but do not map simply onto HAMMER performance (Appendix~\hyperref[app:E]{E}).
\item \emph{Protocol.} The thresholds for failure, intact structure and structure failure were not tuned on the results. Appendix~\hyperref[app:G]{G} varies them and Appendix~\hyperref[app:B]{B} checks them with the threshold-free log-error decomposition and with $\delta_2$ and $\delta_3$: the qualitative conclusions do not change, but the exact proportions do. The motion thresholds of TUM-Family were compared only in terms of sample size. The benchmarks, their intrinsics and the protocol all follow MoGe-2, which may favour the MoGe series.
\item \emph{Training overlap.} Five models were trained on DDAD (Section~\hyperref[sec:3.1]{3.1}). Overlap is checked only at the dataset level. Without DDAD the conclusions are unchanged (Appendix~\hyperref[app:G]{G}).
\item \emph{Ground truth.} We take the released ground truth of each benchmark and of TUM as given, without independent checks; the outdoor DIODE ground truth covers only part of the pixels (4\% to 30\% in Appendix~\hyperref[app:F]{F}), so the metrics there reflect only that region.
\item \emph{Model size.} Sizes are compared within families only (Appendix~\hyperref[app:I]{I}).
\end{itemize}

\textbf{Reproducibility.} We will release the evaluation code, the model adapters (all models are used through their public weights) and instructions for obtaining each third-party repository at the commits we used. The two TUM-derived sets will be released as processed data under CC BY 4.0 (the licence of TUM RGB-D) with the selection rules, source frames and intrinsics, and the results of all 36 model evaluations (including the size evaluations of Appendix~\hyperref[app:I]{I}) as per-image files, together with the shared-failure label of every image (the number $k$ of failing architectures and its tier).

\section{Conclusion}\label{sec:9}

Recent zero-shot metric depth models have not yet solved metric depth estimation in everyday scenes: the top 4 model evaluations by metric $\delta_1$ still have failure rates of 7.7\% to 9.8\%; even taking the lowest failure rate of any of the 18 evaluations on each dataset, DIODE and HAMMER reach 15.2\% and 23.9\%, whereas the other five datasets are at most 0.2\%; these failures are invisible in the means. The remaining difficulty lies mainly in scale: 78.8\% of metric failures have intact structure and are wrong only in scale, after scale correction 98.3\% no longer fail, flawed structure occurs mainly on the DDAD dataset, and structure failures mainly in the DepthVLM model. The largest part of the failures (42\%) is shared by most models: these images cover only 7.3\% of the data, concentrate on HAMMER (4 of its 7 scenes) and DIODE, still account for 34\% to 40\% of the $1-\delta_1$ error of the top 6 models when selected by models of other families (against 14\% to 18\% for as many random images from the same datasets), and point in the same direction for most models; on HAMMER they are mostly empty-table shots. In the HAMMER pairs, adding objects reduces the scale error of most models, and the TUM sequences with printed patterns (separate recordings) have larger scale errors; ground-truth intrinsics mainly affect the scale, and from the smallest to the largest size, 4 of the 5 families reduce both structure and scale errors; with intrinsics and the largest size, every architecture still fails on at least 8.0\% of images, and on 16.3\% and 23.9\% of DIODE and HAMMER (the two minima come from the two public MoGe-2 checkpoints). For users: a reliable scale reference (such as a known length) can largely correct most failures, and the metric scale should not be trusted on views without size cues, such as an empty table or a blank wall. We recommend that future metric depth evaluations (points (i) to (iii) answer the three gaps): (i) report the failure rate and its distribution over datasets besides the mean; (ii) report metric metrics, scale-aligned metrics and the bias, dispersion and absolute error of $s^{*}$ side by side; (iii) evaluate several models on the same images and report the images that most models fail separately, or stratify means by the number of failing models; and (iv) report ground-truth intrinsics as a separate condition, not an upper bound on performance (Section~\hyperref[sec:7]{7}).

\bibliography{main}
\bibliographystyle{tmlr}

\clearpage
\appendix

\section{Model and Data Selection}\label{app:A}

\textbf{Model selection criteria.} We evaluate recent state-of-the-art models that perform monocular metric depth estimation across indoor and outdoor scenes with a single set of weights. This excludes four groups: (i) metric models with separate indoor and outdoor weights, such as ZoeDepth \citep{bhat2023zoedepth} and Depth-Anything-V2-Metric \citep{yang2024depthv2}; (ii) models that output only relative (affine-invariant) depth, such as MiDaS \citep{ranftl2020towards}, Depth Anything v1 and v2 \citep{yang2024depthanything,yang2024depthv2}, and Marigold v1 and v2 \citep{ke2024repurposing,pavlovic2026marigoldv2}, which have no metric scale to evaluate; (iii) models that cannot output a dense depth map, such as DepthLM \citep{cai2025depthlm}, which answers the depth of one pixel per generation and cannot be evaluated with our per-pixel protocol, so among vision-language models we use DepthVLM, which outputs dense depth maps; and (iv) variants of MoGe-2 specialised for long-range scenes: AerialMetric \citep{song2026aerialmetric}, patched for aerial views, and WildMoGe \citep{xiangli2026metricscenes}, MoGe-2 fine-tuned on MetricScenes to recover the scale of distant landmarks. On the 7 benchmarks used here (same protocol), its paper reports a metric $\delta_1$ of 0.734 without and 0.849 with intrinsics, against 0.768 and 0.874 for MoGe-2 (its Table 1), so it does not improve on MoGe-2 in everyday scenes; we evaluate MoGe-2 directly.

\textbf{Datasets considered but not used.} MetricScenes \citep{xiangli2026metricscenes}, which includes a subset of Stereo4D, is a training corpus: its partial depth comes from SfM/MVS with the scale recovered from geo-tagged metadata or known stereo baselines, and the missing regions are filled by a Poisson completion guided by MoGe-2, which also sets the scale of the foreground. Its completed ground truth thus depends on MoGe-2, and WildMoGe, a fine-tuned MoGe-2, is trained on it. Datasets that target only transparent and specular objects are specialised challenges, and further indoor datasets would add to an already large indoor share. Aerial and medical scenes fall outside everyday scenes and their depth ranges lie far outside common training data: the aerial dataset AerialMetric \citep{song2026aerialmetric} was captured at relative flight heights of 80 m and 120 m, with 70 to 300 m in its synthetic part; endoscopic data have working distances of millimetres to centimetres, for example depths of about 15 to 246 mm in SCARED \citep{allan2021stereo}, and ground-truth depth clipped to 0 to 100 mm in C3VD \citep{bobrow2023colonoscopy}. Failures on these scenes would readily be attributed to distribution shift rather than to scale recovery itself, so we do not use them.

\section{An Alternative Decomposition of the Metric Error}\label{app:B}

Section~\hyperref[sec:3.3]{3.3} measures structure with scale-aligned metrics and scale with $s^{*}$. The metric error can also be split into a structure term and a scale term directly on the log error. For the valid pixels of an image, let $d_i = \ln p_i - \ln g_i$ and $\mu = \frac{1}{N}\sum_i d_i$. Then

\[\frac{1}{N}\sum_i d_i^2 = \mu^2 + \frac{1}{N}\sum_i (d_i - \mu)^2 .\]

The second term on the right is the scale-invariant error of \citet{eigen2014depth} (with $\lambda = 1$); it does not change when the prediction is multiplied by any constant, and so measures structure. The first term, $\mu^2$, is the part removed by scaling the prediction by $e^{-\mu}$, and so measures scale. $e^{-\mu}$ is also the best global scale for the log error, since $\frac{1}{N}\sum_i (d_i - c)^2$ is smallest at $c = \mu$. The log-MSE can thus be split into ``scale + structure'' without any threshold. These quantities are computed from the per-pixel log error and released with the per-image results (field \texttt{log\_stats}).

\textbf{The two splits agree.} On metric failures, the scale term accounts for 93\% of the log-MSE; over all images it accounts for 80\%, since images that pass already have small scale errors. Across datasets, the share on failures is lowest on DDAD (85.8\%), consistent with Section~\hyperref[sec:4.3]{4.3}, where flawed structure occurs mainly on DDAD. This split needs no threshold: with failure judged by $\delta_2$ or $\delta_3$, 99.7\% and 99.9\% of the failures no longer fail after scale correction, above the 98.3\% under $\delta_1$. $s^{*}$ and $e^{-\mu}$ are the median and the mean of the same per-pixel log errors, with a median per-image ratio of 1.000 (Spearman correlation 0.98), so on our data the two splits are nearly equivalent; on a few images the two differ widely, and Appendix~\hyperref[app:C]{C} compares them with several other alignment scales. Neither is the scale that maximises $\delta_1$, so the scale-aligned $\delta_1$ is occasionally slightly below the metric $\delta_1$, but by little (Appendix~\hyperref[app:C]{C}). We use $s^{*}$ because multiplying the prediction by one global scale before computing a metric is the usual practice in depth evaluation (scale-aligned metrics) and applies to any metric; the cost is that it is not an exact additive split. Since $\frac{1}{N}\sum_i (d_i + \ln s)^2 = \frac{1}{N}\sum_i (d_i - \mu)^2 + (\mu + \ln s)^2$, the RMSElog after alignment with any scale $s$ equals the scale-invariant error plus the squared deviation of $s$ from $e^{-\mu}$. The extra second term from using $s^{*}$ instead of $e^{-\mu}$ is, over the 79,308 pairs, only 1.4\% of the structure term at the median (9.3\% at the 90th percentile).

\textbf{Limitations of the log decomposition.} It has three limitations. First, the exact additive split holds only for the log-MSE: the identity relies on the ``mean plus variance'' property of the squared error in log space, and Rel, $\delta_1$ and RMSE have no such exact split; multiplying the prediction by $e^{-\mu}$ before computing another metric instead returns to the approach of Section~\hyperref[sec:3.3]{3.3}, differing only in how the scale is estimated, and the two estimates nearly coincide (median ratio 1.000). Second, $\mu^2$ carries no sign ($\mu$ itself is signed and is our bias): overestimating depth by a factor of two gives the same $\mu^2$ as underestimating it by a factor of two. Third, it does not separate systematic bias from instability. $\mu^2$ is one number per image, and averaging it over images mixes two different errors. For the $M$ images of a dataset, let $\overline{\mu^2} = \frac{1}{M}\sum_j \mu_j^2$ be the mean scale term and $\bar\mu$ the mean of the $\mu_j$. Applying the same decomposition once more gives

\[\overline{\mu^2} = \bar\mu^2 + \frac{1}{M}\sum_j (\mu_j - \bar\mu)^2 ,\]

where the first term is the systematic bias (all images off in one direction, fixed by one global constant) and the second the instability (images off either way, beyond global correction); $\overline{\mu^2}$ is their sum.

\textbf{Relation to our statistics.} Since $\ln s^* \approx -\mu$, the bias $\operatorname{mean}(\ln (1/s^{*}))$ of Section~\hyperref[sec:3.3]{3.3} is approximately $\bar\mu$ and keeps its sign, the dispersion $\operatorname{std}(\ln s^{*})$ is approximately the square root of the second term, and the absolute error $\operatorname{mean}(\lvert\ln s^{*}\rvert)$ is approximately the mean of $|\mu|$. The scale term of the log decomposition is thus a merger of our three statistics: it tells how large the scale error is, but not in which direction it goes or whether one constant can correct it. In practice their proportions differ widely: over the 126 cells of 18 model evaluations $\times$ 7 datasets, the systematic bias $\bar\mu^2$ accounts for a median of 14\% of $\overline{\mu^2}$, ranging from 0\% to 98\%. On HAMMER, MoGe-2 (without K) and UniK3D (without K) have almost the same $\overline{\mu^2}$ (0.101 and 0.108), but the former is almost unbiased (bias +0.005, dispersion 0.32), so its error is almost all instability, while the latter has a bias of +0.223 and a dispersion of 0.24, and 47\% of its $\overline{\mu^2}$ is $\bar\mu^2$, a systematic bias that one constant would correct. By $\overline{\mu^2}$ alone the two models look almost the same; only the split above shows that one is unstable and the other systematically biased.

\section{Solving for the Alignment Scale}\label{app:C}

The $s^{*}$ of Section~\hyperref[sec:3.3]{3.3} is the median of $g/p$ over the valid pixels. This appendix compares it with four other alignment scales: the log mean $e^{-\mu}$ of Appendix~\hyperref[app:B]{B}, the ratio of medians of early self-supervised depth evaluation, and the solutions of the MoGe evaluation pipeline on the 64$\times$64 grid and at full resolution. All numbers come from the per-image results of the 18 model evaluations. The five scales use the same valid pixels, are all stored in the per-image results (field $scales$) and can be checked with the comparison script released with the code.

\textbf{Objectives of the five scales.} Let $d_i = \ln p_i - \ln g_i$.

\begin{itemize}
\item Per-pixel median (ours): $\ln s^* = -\mathrm{med}_i\, d_i$, the solution of $\min_s \sum_i |d_i + \ln s|$, i.e. the L1-optimal scale for the per-pixel log error.
\item Log mean: $\ln s = -\frac{1}{N}\sum_i d_i = -\mu$, the solution of $\min_s \sum_i (d_i + \ln s)^2$, i.e. the L2-optimal scale, and the scale term of the log-MSE decomposition of Appendix~\hyperref[app:B]{B}.
\item Ratio of medians: $\ln s = \mathrm{med}(\ln g) - \mathrm{med}(\ln p)$, i.e. $median(g)/median(p)$ (the evaluation code of \citealp{zhou2017unsupervised} and \citealp{godard2019digging}).
\item MoGe: $\min_s \sum_i |s p_i - g_i| / g_i = \sum_i (p_i / g_i)\,|s - g_i / p_i|$, whose solution is the median of $g_i / p_i$ weighted by $p_i / g_i$, solved on the 64$\times$64 grid or on all valid pixels.
\end{itemize}

\paragraph{Why the per-pixel median}

\begin{enumerate}
\item \textbf{Consistent with per-pixel metrics.} $\delta_1$ and Rel are per-pixel errors. The per-pixel median and the log mean both minimise the per-pixel log error directly, one in L1 and one in L2.
\item \textbf{Robust to outlying pixels.} The median has a breakdown point of 50\%: as long as fewer than half of the pixels are wrong, $s^{*}$ is set by the others. The mean has a breakdown point of 0: a few pixels with very large log errors can push $e^{-\mu}$ away. So the log mean nearly equals the median on most images, but on images with outlying pixels it can turn intact structure into a structure failure.
\item \textbf{Uses pixel correspondence.} The ratio of medians compares only the distributions of the two depth maps. It uses no pixel correspondence and corresponds to no per-pixel objective: shuffling the predicted pixels within the image leaves it unchanged. If the prediction preserves the depth order and the log error varies monotonically with depth (e.g. $p = c\,g^k$, a global compression or stretch of the depth range), the two coincide exactly. They differ only when a local region is wrong: then the medians of $p$ and $g$ fall on different pixels, and the ratio of medians folds the depth difference between the two regions into $s$.
\item \textbf{Symmetric in log.} The per-pixel median, the log mean and the ratio of medians are symmetric in $p$ and $g$ (swapping the two turns $s$ into $1/s$): overestimating by a factor of two and underestimating by a factor of two give values of $\ln s$ that differ only in sign. The criterion $\max(p/g, g/p) < 1.25$ of $\delta_1$ and the statistics of $\ln s^{*}$ are symmetric in the same way, so the alignment scale should be too. MoGe's weighted median is not: points predicted too far ($p/g$ large) get more weight and pull $s$ down, towards judging the depth overestimated. Solving at full resolution instead of on the 64$\times$64 grid does not remove this tendency.
\end{enumerate}

\textbf{On most images the five nearly coincide.} The share of pairs within 5\% of the per-pixel median is 94.1\% for the log mean, 91.5\% for the ratio of medians, 94.8\% for MoGe 64$\times$64 and 97.7\% for full resolution (Table~\hyperref[tab:C1]{C.1}). The log mean and the ratio of medians deviate from the per-pixel median in no particular direction, with about half of the pairs on each side; MoGe 64$\times$64 is below the median in 86.3\% of pairs, and full resolution is never above it. Per image, the ratio of $s^{*}$ to $e^{-\mu}$ has a median of 1.000 and a Spearman correlation of 0.98.

\textbf{The differences concentrate on a few images.} The number of scale-aligned failures (scale-aligned $\delta_1$ $<$ 0.5) is 274 for the per-pixel median, 603 for the log mean, 534 for the ratio of medians, 872 for MoGe 64$\times$64 and 486 at full resolution. Pairs that pass the metric $\delta_1$ but become structure failures through the alignment number 14, 127, 186, 309 and 108 in the same order, mostly on DDAD and DIODE (Table~\hyperref[tab:C1]{C.1}). A few outliers can pull the MoGe solution close to 0: $s$ $<$ 0.05 occurs in 61 pairs on the 64$\times$64 grid, 11 at full resolution and none for the other three. The log mean has the highest mean scale-aligned $\delta_1$ (0.956, against 0.955 for the per-pixel median), but 2.2 times as many scale-aligned failures. The main conclusions hold under all five scales: among metric failures, intact structure ranges from 78.3\% to 81.5\% and passing after scale correction from 96.3\% to 98.3\%, against 78.8\% and 98.3\% under the per-pixel median. The median is not the scale that maximises $\delta_1$: the scale-aligned $\delta_1$ is below the metric $\delta_1$ in 23.2\% of pairs, but the median drop is only 0.0015 and exceeds 0.05 in only 0.85\% (0.63\% under the log mean, 1.1\% to 1.9\% under the other three).

\begin{table}[htbp]
\centering
\renewcommand{\thetable}{C.1}
\caption{Comparison of the five alignment scales, 18 model evaluations. ``Within 5\%'' is the share of pairs whose scale is within 5\% of the per-pixel median, for log mean / ratio of medians / MoGe 64$\times$64 / MoGe full resolution; the last two columns are for per-pixel median (ours) / log mean / ratio of medians / MoGe 64$\times$64 / MoGe full resolution.}
\label{tab:C1}
\small
\setlength{\tabcolsep}{4pt}
\begin{tabular}{lrrr}
\toprule
Dataset & Within 5\% & SA failures & \makecell{Metric pass, SA failure} \\
\midrule
NYUv2 & 97.8\% / 95.8\% / 99.8\% / 99.8\% & 0 / 2 / 1 / 1 / 1 & 0 / 2 / 1 / 1 / 1 \\
KITTI & 98.4\% / 94.6\% / 97.3\% / 99.6\% & 3 / 4 / 3 / 10 / 3 & 0 / 1 / 0 / 7 / 0 \\
ETH3D & 97.6\% / 96.2\% / 96.2\% / 98.9\% & 18 / 51 / 37 / 78 / 42 & 0 / 3 / 5 / 19 / 2 \\
iBims-1 & 95.9\% / 95.3\% / 98.7\% / 98.7\% & 1 / 1 / 1 / 2 / 2 & 0 / 0 / 0 / 1 / 1 \\
DDAD & 89.6\% / 82.1\% / 88.9\% / 95.9\% & 43 / 165 / 194 / 175 / 98 & 3 / 53 / 127 / 92 / 43 \\
DIODE & 86.3\% / 92.4\% / 92.0\% / 96.2\% & 104 / 253 / 172 / 470 / 204 & 11 / 68 / 50 / 189 / 61 \\
HAMMER & 98.7\% / 93.5\% / 97.5\% / 97.5\% & 105 / 127 / 126 / 136 / 136 & 0 / 0 / 3 / 0 / 0 \\
\midrule
All & 94.1\% / 91.5\% / 94.8\% / 97.7\% & 274 / 603 / 534 / 872 / 486 & 14 / 127 / 186 / 309 / 108 \\
\bottomrule
\end{tabular}
\end{table}

\textbf{Dispersion and bias.} Per model, the dispersion $\operatorname{std}(\ln s)$ under the log mean and the ratio of medians differs from the per-pixel median by at most 0.009, and the bias $\operatorname{mean}(\ln (1/s))$ by at most 0.006. MoGe 64$\times$64 has the largest dispersion: Depth Pro rises from 0.305 to 0.460 and MA-DepthMap from 0.109 to 0.234; these models have predictions clipped at 1000 m (the Depth Pro family) or outliers on DDAD and DIODE. Both MoGe solutions shift the bias upwards, by at most 0.045 above the median (Depth Pro), consistent with their tendency to judge the depth overestimated (Table~\hyperref[tab:C2]{C.2}).

\begin{table}[htbp]
\centering
\renewcommand{\thetable}{C.2}
\caption{Dispersion $\operatorname{std}(\ln s)$, bias $\operatorname{mean}(\ln (1/s))$ and scale-aligned failure rate (scale-aligned $\delta_1$ $<$ 0.5) under the five alignments, 7 benchmarks weighted equally. Each cell gives per-pixel median (ours) / log mean / ratio of medians / MoGe 64$\times$64 / MoGe full resolution.}
\label{tab:C2}
\scriptsize
\setlength{\tabcolsep}{3.6pt}
\begin{tabular}{llrrr}
\toprule
Model & Intrinsics & Dispersion & Bias & \makecell{SA failure rate} \\
\midrule
Metric3D-v2 & with K & 0.084/0.085/0.090/0.094/0.094 & ${+0.107}/{+0.106}/{+0.107}/{+0.111}/{+0.111}$ & 0.0\%/0.0\%/0.1\%/0.0\%/0.0\% \\
DA3METRIC & with K & 0.106/0.108/0.114/0.133/0.117 & ${-0.004}/{-0.004}/{+0.001}/{+0.008}/{+0.002}$ & 0.1\%/0.4\%/0.5\%/0.8\%/0.2\% \\
Depth Pro & w/o K & 0.305/0.304/0.307/0.460/0.325 & ${-0.162}/{-0.165}/{-0.159}/{-0.117}/{-0.150}$ & 0.4\%/1.5\%/0.9\%/1.7\%/1.0\% \\
Depth Pro & with K & 0.280/0.280/0.283/0.439/0.300 & ${-0.137}/{-0.140}/{-0.134}/{-0.091}/{-0.124}$ & 0.4\%/1.5\%/0.9\%/1.6\%/1.0\% \\
MA-DepthMap & with K & 0.109/0.108/0.115/0.234/0.119 & ${-0.042}/{-0.042}/{-0.039}/{-0.010}/{-0.037}$ & 0.1\%/0.1\%/0.2\%/1.1\%/0.2\% \\
DepthVLM & with K & 0.149/0.148/0.158/0.158/0.154 & ${+0.092}/{+0.089}/{+0.097}/{+0.119}/{+0.115}$ & 2.7\%/3.7\%/4.1\%/3.9\%/3.5\% \\
MoGe-2 & w/o K & 0.165/0.164/0.168/0.170/0.169 & ${-0.069}/{-0.070}/{-0.069}/{-0.061}/{-0.064}$ & 0.1\%/0.3\%/0.1\%/0.5\%/0.2\% \\
MoGe-2 & with K & 0.133/0.130/0.137/0.141/0.138 & ${+0.032}/{+0.031}/{+0.029}/{+0.047}/{+0.040}$ & 0.2\%/0.5\%/0.2\%/0.8\%/0.4\% \\
MoGe-3 & w/o K & 0.160/0.158/0.163/0.166/0.163 & ${-0.083}/{-0.085}/{-0.084}/{-0.075}/{-0.078}$ & 0.1\%/0.3\%/0.2\%/0.5\%/0.2\% \\
MoGe-3 & with K & 0.125/0.121/0.129/0.134/0.130 & ${+0.033}/{+0.032}/{+0.029}/{+0.050}/{+0.042}$ & 0.2\%/0.6\%/0.2\%/1.0\%/0.6\% \\
MA-PointMap & w/o K & 0.155/0.154/0.159/0.163/0.162 & ${-0.019}/{-0.017}/{-0.017}/{-0.012}/{-0.014}$ & 0.1\%/0.2\%/0.2\%/0.5\%/0.2\% \\
MA-PointMap & with K & 0.129/0.128/0.135/0.139/0.137 & ${+0.001}/{+0.002}/{+0.002}/{+0.009}/{+0.006}$ & 0.2\%/0.3\%/0.2\%/0.6\%/0.3\% \\
FoundationGeo & w/o K & 0.152/0.153/0.155/0.155/0.154 & ${-0.026}/{-0.032}/{-0.028}/{-0.018}/{-0.021}$ & 0.2\%/0.3\%/0.3\%/0.4\%/0.2\% \\
FoundationGeo & with K & 0.136/0.136/0.141/0.141/0.140 & ${+0.040}/{+0.036}/{+0.038}/{+0.051}/{+0.047}$ & 0.2\%/0.5\%/0.4\%/0.6\%/0.3\% \\
UniDepthV2 & w/o K & 0.180/0.181/0.182/0.192/0.187 & ${+0.064}/{+0.066}/{+0.067}/{+0.073}/{+0.069}$ & 0.1\%/0.3\%/0.2\%/0.7\%/0.3\% \\
UniDepthV2 & with K & 0.145/0.148/0.150/0.162/0.157 & ${+0.074}/{+0.077}/{+0.078}/{+0.083}/{+0.079}$ & 0.1\%/0.3\%/0.2\%/0.7\%/0.3\% \\
UniK3D & w/o K & 0.149/0.150/0.153/0.160/0.155 & ${+0.046}/{+0.049}/{+0.050}/{+0.056}/{+0.051}$ & 0.1\%/0.3\%/0.3\%/0.7\%/0.2\% \\
UniK3D & with K & 0.111/0.113/0.117/0.129/0.122 & ${+0.047}/{+0.050}/{+0.051}/{+0.057}/{+0.052}$ & 0.1\%/0.3\%/0.4\%/0.7\%/0.2\% \\
\bottomrule
\end{tabular}
\end{table}

\begin{figure}[htbp]
\centering
\renewcommand{\thefigure}{C.1}

\includegraphics[width=1.00\linewidth]{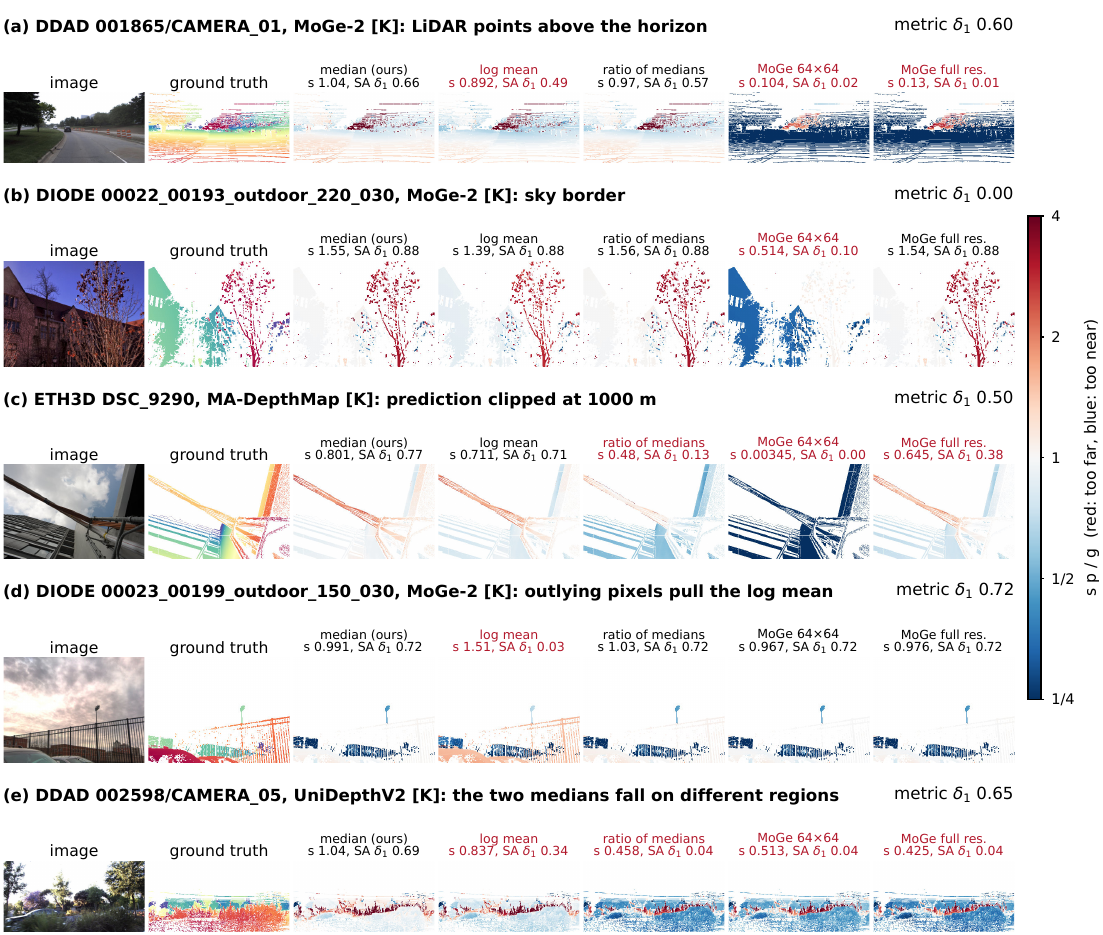}
\caption{Special cases of the five alignments. Each row shows the image, the ground-truth depth, and the error $s p / g$ under the per-pixel median (ours), log mean, ratio of medians, MoGe 64$\times$64 and MoGe full-resolution $s$ (red: predicted too far, blue: too near, white: agreement). Column titles give each $s$ and scale-aligned $\delta_1$ (SA); titles of structure failures (SA $\delta_1$ $<$ 0.5) are in red. The metric $\delta_1$ is at the right of each row title.}
\label{fig:C1}
\end{figure}

\textbf{Special cases.} Figure~\hyperref[fig:C1]{C.1} shows five examples; each row gives the image, the ground truth and the error map $s p / g$ under the five values of $s$. Each group of numbers below is in the order per-pixel median / log mean / ratio of medians / MoGe 64$\times$64 / MoGe full resolution.

\begin{itemize}
\item (a) DDAD 001865/CAMERA\_01, MoGe-2 with K: among the LiDAR points above the horizon lie some very near noisy points. $s$ is 1.04 / 0.892 / 0.970 / 0.104 / 0.130 and the scale-aligned $\delta_1$ is 0.66 / 0.49 / 0.57 / 0.02 / 0.01. On this image all 18 evaluations are structure failures under the 64$\times$64 solution, and none is under the per-pixel median.
\item (b) DIODE 00022\_00193\_outdoor\_220\_030, MoGe-2 with K: ground-truth points near the sky border are enlarged on the 64$\times$64 grid. $s$ is 1.55 / 1.39 / 1.56 / 0.514 / 1.54, and the scale-aligned $\delta_1$ is 0.88 / 0.88 / 0.88 / 0.10 / 0.88. Here full resolution alone fixes the problem.
\item (c) ETH3D electro/DSC\_9290, MA-DepthMap with K: 0.1\% of the pixels are clipped at 1000 m. $s$ is 0.801 / 0.711 / 0.480 / 0.0034 / 0.645 and the scale-aligned $\delta_1$ is 0.77 / 0.71 / 0.13 / 0.00 / 0.38. Full resolution reduces the problem but does not solve it.
\item (d) DIODE 00023\_00199\_outdoor\_150\_030, MoGe-2 with K: a few pixels with very large log errors pull the mean away. $s$ is 0.991 / 1.51 / 1.03 / 0.967 / 0.976, and the scale-aligned $\delta_1$ is 0.72 / 0.03 / 0.72 / 0.72 / 0.72. On the same image FoundationGeo, MoGe-3 and DepthVLM (all with K) are also structure failures only under the log mean.
\item (e) DDAD 002598/CAMERA\_05, UniDepthV2 with K: the prediction is too far around the distant trees and slightly too near in the foreground below, so the medians of $p$ and $g$ fall on different regions, and the ratio of medians judges the whole image more than twice too far. $s$ is 1.04 / 0.837 / 0.458 / 0.513 / 0.425, the scale-aligned $\delta_1$ is 0.69 / 0.34 / 0.04 / 0.04 / 0.04, and the metric $\delta_1$ is 0.65. The same holds for all 4 evaluations of UniK3D and UniDepthV2 with and without K.
\end{itemize}

We do not modify the ground truth of these images; instead, we use an alignment scale that is robust to a few outlying pixels.

The images that are structure failures in at least 9 evaluations under the 64$\times$64 solution and in fewer under the median number 25: 22 from DIODE (18 of them from scene\_00022), 2 from DDAD and 1 from ETH3D; the comparison script prints the list of these images.

\section{How Models Use the Focal Length}\label{app:D}

Table~\hyperref[tab:1]{1} groups the 11 architectures into four classes by how the intrinsics enter the model (Section~\hyperref[sec:3.1]{3.1}). For each class we describe three things: where the intrinsics enter the model, where the metric scale comes from, and where the focal length comes from when no intrinsics are given. The descriptions below are based on the papers and the inference code of each model; $f$ and $W$ denote the focal length and image width in pixels.

\textbf{1. Canonical camera space, output conversion.} The same scene captured with cameras of different focal lengths shows objects at different sizes in the image while the depth is the same; a network that sees only the image cannot distinguish ``the object is near'' from ``the focal length is long''. Metric3D \citep{yin2023metric3d} proposed the canonical camera space transformation (CSTM) to remove this ambiguity: during training, all samples are converted to a canonical camera with focal length $f_{c}$, so the network learns depth in the canonical camera; at inference, the output is converted back using the real focal length. It offers two variants: transforming the depth labels (CSTM label) or transforming the input image (CSTM image). Earlier camera-aware methods such as CAM-Convs \citep{facil2019camconvs} feed camera parameters as extra input channels, and Metric3D compares against this in an ablation. The four models of this class follow the idea of CSTM label: the network does not receive the focal length, which appears only at the output:
\begin{itemize}
\item Metric3D-v2 \citep{hu2024metric3dv2}: $\text{depth} = \text{network output} \times f / 1000$ (canonical focal length 1000 px).
\item DA3METRIC \citep{lin2025da3}: the paper states that it follows the canonical camera space transformation of Metric3D-v2; at inference, $\text{depth} = \text{network output} \times f / 300$, where $f$ is the mean of fx and fy on the network input image.
\item Depth Pro \citep{bochkovskii2025depthpro}: the network predicts a canonical inverse depth $C$, and $\text{inverse depth} = C \times W / f$. The paper cites \citet{yin2023metric3d} for the focal length conversion. It differs from the previous two in two respects: the conversion acts on inverse depth, and the normalisation uses the image width $W$ rather than a fixed canonical focal length.
\item MetricAnything-DepthMap \citep{ma2026metricanything}: follows the formula of Depth Pro, $\text{inverse depth} = C \times W / f$, with the same canonical inverse depth and the same normalisation by the image width.
\end{itemize}

In this class, the metric scale is determined by the product of the network output and the focal length: an error by some factor in the focal length gives an error by the same factor in depth. Depth Pro has a field-of-view prediction head; without intrinsics, the predicted field of view gives $f = 0.5 W / \tan(\text{FoV}/2)$, so intrinsics are optional for it. The other three lack a focal length source and need intrinsics.

\textbf{2. Canonical camera space, input rescaling.} DepthVLM \citep{yu2026depthvlm} rescales the input image by a factor $f_{c} / f$ ($f_{c} = 1000$) so that the focal length of the rescaled image equals the canonical focal length. Depth values do not change with image rescaling, so the network outputs metric depth directly on the rescaled image, and the depth map is then resized back to the original size. This is the same transformation as CSTM image in Metric3D. The focal length only determines the rescaling factor; without intrinsics no rescaling is possible, so DepthVLM is evaluated only with intrinsics.

\textbf{3. Post-hoc geometric solve} (MoGe-2, MoGe-3, MA-PointMap, FoundationGeo). The network does not receive the focal length and outputs a point map, one 3D point $(x, y, z)$ per pixel. The point map differs from the true point cloud by a global scale and a shift $t$ along the optical axis. The focal length can be recovered from the point map because pinhole projection requires every pixel to satisfy $(u, v) = f \cdot (x, y) / (z + t)$, where $(u, v)$ is the image coordinate of the pixel (with the image centre as origin and the half-diagonal as unit); with the pixel positions known and the point map given by the network, $f$ and $t$ are constrained by this set of equations. The four models use the same solver code (FoundationGeo and MA-PointMap reuse the MoGe implementation): the point map is downsampled to 64$\times$64 and $\textstyle\sum  \|f \cdot (x, y) / (z + t) - (u, v)\|^{2}$ is minimised. For a given $t$ the optimal $f$ has a closed form, and $t$ is then solved with the Levenberg–Marquardt method. Without intrinsics, $f$ and $t$ are solved jointly; with intrinsics, $f$ is fixed to the true focal length and only $t$ is solved. The depth is then $z + t$, scaled by the scale head (below).

The metric scale does not come from this solve. MoGe-2, MoGe-3 and MA-PointMap use a separate scale head that predicts a scalar from the global image feature (cls token), which is multiplied onto the depth after the solve; FoundationGeo predicts a per-pixel scale field, likewise multiplied after the solve, and also corrects the ray direction of each pixel by at most 3$^\circ$. In this class, the intrinsics therefore affect depth only through the shift $t$: $t$ changes the ratios between pixel depths in $z + t$ as well as their overall size, but the output of the scale head is not affected by the intrinsics. Before the solve, MoGe-3 refines the log depth of the point map in 3 iterations, which do not involve the focal length.

\textbf{4. Rays as network input} (UniDepthV2, UniK3D). A ray is the direction from the camera centre through each pixel. Given intrinsics $K$, the ray of pixel $(u, v)$ is $r = K^{-1} [u, v, 1]^{\top}$, normalised to a unit vector; both models compute it this way. The rays are converted to polar and azimuthal angles, encoded as Fourier features, and fed to the depth decoder as conditioning. The decoder predicts a distance $\rho$ along each ray, the 3D point is $\rho \cdot r$, and depth is its z component. Without intrinsics, the rays are predicted by the camera head of the network: the camera head of UniDepthV2 predicts $f_x, f_y, c_x, c_y$ and obtains the rays with the same formula; that of UniK3D predicts the field of view and a set of spherical harmonic coefficients from which it generates the rays, which can represent non-pinhole cameras.

In this class, the metric scale comes from $\rho$; the rays determine directions and influence the prediction of $\rho$ through the conditioning, rather than converting depth multiplicatively as in class 1.

\textbf{Which elements of the intrinsics ``with K'' uses.} The classes use different elements of the intrinsics, so ``with K'' is not the same intervention for all of them: the four MoGe-family models use only fx (converted from fov\_x) and take the principal point at the image centre; UniDepthV2 and UniK3D use the full K; Metric3D-v2, DA3METRIC, Depth Pro and MA-DepthMap use only the focal length; DepthVLM uses the focal length only to set its resize factor. The MoGe evaluation loader first reprojects every image to a view with the principal point at the image centre, so every model receives a centred principal point and the models that use only the focal length lose no information.

\section{Comparison with Published Results}\label{app:E}

The purpose of this appendix is to confirm that the results we integrated and evaluated ourselves carry no bias introduced by our implementation. The input resolution of each model follows the MoGe-2 evaluation settings, and the other models use their default configurations. The pipeline uses the data loading and metric code shipped with MoGe, unmodified, and other models are connected through thin adapters; the configuration of the 7 benchmarks (resolution, split files, depth units, segmentation files) is identical item by item to the official \texttt{moge2.json}, and ground-truth intrinsics are passed through the interface of the official \texttt{--oracle} option.

\textbf{Comparison with the MoGe-2 paper (same protocol).} Table B.4 in the supplementary material of the MoGe-2 paper \citep{wang2025moge2} was produced by its authors with the same data and metric code on the same 7 benchmarks, and reports the metric depth (without scale alignment) of MoGe-2 as well as Depth Pro, UniDepthV2 and Metric3D-v2. The reference values of these four models in Table~\hyperref[tab:E1]{E.1} are all taken from that table, not from the models' own papers. We convert the percentages of the MoGe-2 paper to fractions.

\begin{table}[htbp]
\centering
\renewcommand{\thetable}{E.1}
\caption{Comparison with Table B.4 of the MoGe-2 paper. ``Mean'' is the average over images within each dataset followed by the equal-weight average over the 7 datasets, given as Rel / $\delta_1$; ``max. difference'' is the largest absolute difference between our values and those of the MoGe-2 paper over the 7 datasets, in the same format.}
\label{tab:E1}
\small
\setlength{\tabcolsep}{4pt}
\begin{tabular}{llrrr}
\toprule
Model & K & Mean (ours) & Mean (MoGe-2 paper) & Max.\ difference on one dataset \\
\midrule
Metric3D-v2 & with K & 0.183 / 0.739 & 0.183 / 0.739 & 0.001 / 0.001 \\
Depth Pro & w/o K & 0.276 / 0.543 & 0.276 / 0.544 & 0.000 / 0.002 \\
MoGe-2 & w/o K & 0.157 / 0.769 & 0.157 / 0.768 & 0.002 / 0.007 \\
MoGe-2 & with K & 0.136 / 0.877 & 0.136 / 0.874 & 0.004 / 0.008 \\
UniDepthV2 & w/o K & 0.213 / 0.753 & 0.213 / 0.753 & 0.001 / 0.001 \\
UniDepthV2 & with K & 0.185 / 0.826 & 0.185 / 0.826 & 0.001 / 0.002 \\
\bottomrule
\end{tabular}
\end{table}

All four models agree with the MoGe-2 paper on every dataset: over the 42 ``condition $\times$ dataset'' cells of the 6 rows of Table~\hyperref[tab:E1]{E.1}, the differences in Rel and $\delta_1$ do not exceed 0.01.

\textbf{Comparison of MoGe-3 with its paper (same protocol).} MoGe-3 \citep{kong2026moge3} uses the same evaluation code and benchmarks as MoGe-2. Our MoGe-3 results, obtained with the official evaluation script, the public weights and the default settings, disagree with the metric depth values in Table C.1 of its appendix; for example, the metric $\delta_1$ on KITTI without intrinsics is 0.553 for ViT-L against 0.851 in the paper, while the same pipeline reproduces the Depth Pro and UniK3D values of that table. We reported the discrepancy to the authors of MoGe-3, who confirmed it, are investigating it, and recommend using the reproducible results of the released models as the basis for comparison. All MoGe-3 results in this paper come from the public weights.

\textbf{Comparison with the models' own papers (protocols may differ).} The other models are not in Table B.4. For them, we compare with their papers on the test sets shared with this paper (Table~\hyperref[tab:E2]{E.2}). The evaluation protocols of these papers (data preprocessing and resampling, input resolution, valid-pixel masks, depth clipping, splits) may differ from the MoGe-2 benchmarks, so we require only broad agreement, not cell-by-cell equality.

\begin{table}[htbp]
\centering
\renewcommand{\thetable}{E.2}
\caption{Comparison with the models' own papers, all for metric depth. The ``Source'' column gives the table number in the model's paper, not in this paper. Each cell is the result on one dataset (not a cross-dataset average); our value is the mean over all images of that dataset. ``—'' means that the paper does not report Rel. For UniK3D, DIODE is compared only on its indoor part, because its paper reports only indoor results. The MetricAnything paper does not state the focal length used for its Table 4.}
\label{tab:E2}
\small
\setlength{\tabcolsep}{4pt}
\begin{tabular*}{\linewidth}{@{\extracolsep{\fill}}llllrrrr@{}}
\toprule
Model & K & Source & Dataset & \makecell{Rel\\(paper)} & \makecell{Rel\\(ours)} & \makecell{$\delta_1$\\(paper)} & \makecell{$\delta_1$\\(ours)} \\
\midrule
DA3METRIC & with K & Depth Anything 3 Table 11 & NYUv2 & 0.070 & 0.071 & 0.963 & 0.970 \\
DA3METRIC & with K & Depth Anything 3 Table 11 & KITTI & 0.086 & 0.073 & 0.953 & 0.960 \\
DA3METRIC & with K & Depth Anything 3 Table 11 & ETH3D & 0.104 & 0.125 & 0.917 & 0.855 \\
DA3METRIC & with K & Depth Anything 3 Table 11 & DIODE & 0.128 & 0.152 & 0.838 & 0.766 \\
DepthVLM & with K & DepthVLM Table 2 & DDAD & — & 0.182 & 0.818 & 0.721 \\
DepthVLM & with K & DepthVLM Table 2 & ETH3D & — & 0.085 & 0.924 & 0.940 \\
DepthVLM & with K & DepthVLM Table 2 & iBims-1 & — & 0.112 & 0.912 & 0.875 \\
DepthVLM & with K & DepthVLM Table 2 & NYUv2 & — & 0.087 & 0.908 & 0.940 \\
MA-PointMap & w/o K & MetricAnything Table 5 & KITTI & 0.032 & 0.091 & 0.977 & 0.931 \\
MA-PointMap & w/o K & MetricAnything Table 5 & ETH3D & 0.108 & 0.107 & 0.902 & 0.896 \\
MA-PointMap & w/o K & MetricAnything Table 5 & iBims-1 & 0.113 & 0.147 & 0.866 & 0.848 \\
MA-PointMap & w/o K & MetricAnything Table 5 & DIODE & 0.139 & 0.165 & 0.798 & 0.728 \\
MA-PointMap & with K & MetricAnything Table 6 & KITTI & 0.055 & 0.074 & 0.996 & 0.966 \\
MA-PointMap & with K & MetricAnything Table 6 & ETH3D & 0.098 & 0.099 & 0.924 & 0.912 \\
MA-PointMap & with K & MetricAnything Table 6 & iBims-1 & 0.095 & 0.105 & 0.920 & 0.910 \\
MA-PointMap & with K & MetricAnything Table 6 & DIODE & 0.141 & 0.155 & 0.743 & 0.784 \\
MA-DepthMap & with K & MetricAnything Table 4 & ETH3D & 0.147 & 0.158 & 0.799 & 0.781 \\
FoundationGeo & w/o K & FoundationGeo Table 2 & NYUv2 & 0.102 & 0.078 & 0.930 & 0.959 \\
FoundationGeo & w/o K & FoundationGeo Table 2 & KITTI & 0.079 & 0.072 & 0.949 & 0.961 \\
FoundationGeo & w/o K & FoundationGeo Table 2 & ETH3D & 0.178 & 0.145 & 0.740 & 0.803 \\
FoundationGeo & w/o K & FoundationGeo Table 2 & iBims-1 & 0.104 & 0.103 & 0.900 & 0.917 \\
FoundationGeo & w/o K & FoundationGeo Table 2 & DDAD & 0.176 & 0.181 & 0.747 & 0.744 \\
FoundationGeo & w/o K & FoundationGeo Table 2 & DIODE & 0.175 & 0.186 & 0.697 & 0.617 \\
FoundationGeo & w/o K & FoundationGeo Table 2 & HAMMER & 0.224 & 0.181 & 0.696 & 0.717 \\
UniK3D & w/o K & UniK3D Tables 18–22 & NYUv2 & 0.074 & 0.099 & 0.965 & 0.928 \\
UniK3D & w/o K & UniK3D Tables 18–22 & KITTI & 0.174 & 0.095 & 0.812 & 0.958 \\
UniK3D & w/o K & UniK3D Tables 18–22 & iBims-1 & 0.104 & 0.093 & 0.919 & 0.936 \\
UniK3D & w/o K & UniK3D Tables 18–22 & ETH3D & 0.236 & 0.149 & 0.687 & 0.837 \\
UniK3D & w/o K & UniK3D Tables 18–22 & DIODE (indoor) & 0.161 & 0.148 & 0.713 & 0.760 \\
\bottomrule
\end{tabular*}
\end{table}

In 12 of the 29 cells, $\delta_1$ is within 0.02 of the paper, and the gaps have no common direction: for example, we are about 0.15 higher for UniK3D on KITTI and ETH3D, about 0.07 to 0.08 lower for FoundationGeo and DA3METRIC on DIODE, and about 0.10 lower for DepthVLM on DDAD (DDAD is in the training data of DepthVLM, Section~\hyperref[sec:3.1]{3.1}). Since the protocols of these papers differ from ours, we do not interpret these gaps as integration errors, but they show that the models' own numbers are not directly comparable with ours.

\textbf{The anomaly of DepthVLM on HAMMER.} DepthVLM overestimates depth most strongly on HAMMER (bias +0.68, median scale factor 0.52, i.e. it overestimates the depth of the table scenes by about a factor of two). This is not a problem of our integration code: with the same integration code, DepthVLM is almost unbiased on NYUv2, ETH3D and iBims-1 (bias $-$0.00, $-$0.01 and $-$0.02), and its results are close to those reported in its own paper (Table~\hyperref[tab:E2]{E.2}). Its failures on HAMMER worsen as the scene gets nearer: the failure rate is 11\% for images with a median ground-truth depth above 1 m, 77\% for 0.8 to 1 m, 90\% for 0.6 to 0.8 m, and 100\% below 0.6 m, while the median scale factor falls from 0.84 to 0.42. We therefore regard this as the behaviour of the model itself. The training data of DepthVLM contain only room-scale scans and driving scenes \citep{yu2026depthvlm}, with no close-range tabletop data. Training data do not simply explain the differences on HAMMER, however: the training data of UniDepthV2 and UniK3D include object-level and handheld-interaction close-range data (HOI4D, WildRGBD), yet their failure rates on HAMMER (53\% and 43\%) are higher than that of MoGe-2 (24\%), whose training data (\citealp{wang2025moge2}, Table A.1) contain no close-range tabletop dataset.

\section{Per-Dataset Results}\label{app:F}

Table~\hyperref[tab:3]{3} is an equal-weight average over the 7 benchmarks and does not show what happens within each dataset, while Sections~\hyperref[sec:4]{4} and \hyperref[sec:5]{5} cite many single-dataset numbers. Tables~\hyperref[tab:F1]{F.1} to F.7 give the metrics of Table~\hyperref[tab:3]{3} per dataset (the scale-aligned failure rate, mostly below 1\% on every dataset, is omitted and replaced by the share of $s^{*}$ within [0.8, 1.25]), computed as in Table~\hyperref[tab:3]{3} but without averaging over datasets: each value is computed over all images of that dataset. The arithmetic mean of one column of one model over the 7 tables gives the corresponding value in Table~\hyperref[tab:3]{3}. The last column is the share of images with $s^{*}$ within [0.8, 1.25], i.e. with a scale inside the range allowed by the $\delta_1$ criterion (Section~\hyperref[sec:4.2]{4.2}).

\begin{table}[htbp]
\centering
\renewcommand{\thetable}{F.1}
\caption{NYUv2 (indoor rooms, 654 images). $\uparrow$/$\downarrow$: higher/lower is better; $\to 0$: closer to zero is better.}
\label{tab:F1}
\footnotesize
\setlength{\tabcolsep}{2.6pt}
\begin{tabular*}{\linewidth}{@{\extracolsep{\fill}}llrrrrrrrrr@{}}
\toprule
& & \multicolumn{3}{c}{Metric} & \multicolumn{2}{c}{Structure} & \multicolumn{4}{c}{Scale ($s^{*}$)} \\ \cmidrule(lr){3-5}\cmidrule(lr){6-7}\cmidrule(lr){8-11} Model & K & $\delta_1\uparrow$ & Rel$\downarrow$ & Fail.$\downarrow$ & $\delta_1\uparrow$ & Rel$\downarrow$ & Bias ($\to 0$) & Disp.$\downarrow$ & Abs.\ err.$\downarrow$ & $s^{*}\in[0.8,1.25]\uparrow$ \\
\midrule
Metric3D-v2 & with K & 0.965 & 0.071 & 0.5\% & 0.974 & 0.047 & ${-0.001}$ & 0.063 & 0.047 & 99.5\% \\
\addlinespace[1.5pt]
DA3METRIC & with K & 0.970 & 0.071 & 0.3\% & 0.981 & 0.042 & ${-0.043}$ & 0.058 & 0.057 & 99.7\% \\
\addlinespace[1.5pt]
Depth Pro & w/o K & 0.919 & 0.107 & 3.5\% & 0.976 & 0.044 & ${+0.053}$ & 0.098 & 0.086 & 96.6\% \\
Depth Pro & with K & 0.935 & 0.093 & 2.6\% & 0.976 & 0.044 & ${+0.008}$ & 0.098 & 0.074 & 97.4\% \\
\addlinespace[1.5pt]
MA-DepthMap & with K & 0.969 & 0.078 & 0.6\% & 0.985 & 0.037 & ${-0.045}$ & 0.071 & 0.068 & 99.4\% \\
\addlinespace[1.5pt]
DepthVLM & with K & 0.940 & 0.087 & 0.2\% & 0.956 & 0.072 & ${-0.001}$ & 0.058 & 0.046 & 99.8\% \\
\addlinespace[1.5pt]
MoGe-2 & w/o K & 0.961 & 0.073 & 1.7\% & 0.983 & 0.034 & ${-0.006}$ & 0.080 & 0.059 & 98.3\% \\
MoGe-2 & with K & 0.968 & 0.064 & 1.1\% & 0.983 & 0.034 & ${-0.004}$ & 0.067 & 0.049 & 98.9\% \\
\addlinespace[1.5pt]
MoGe-3 & w/o K & 0.967 & 0.069 & 1.1\% & 0.983 & 0.034 & ${-0.010}$ & 0.072 & 0.055 & 98.9\% \\
MoGe-3 & with K & 0.973 & 0.061 & 0.8\% & 0.983 & 0.033 & ${-0.003}$ & 0.060 & 0.046 & 99.2\% \\
\addlinespace[1.5pt]
MA-PointMap & w/o K & 0.948 & 0.085 & 2.9\% & 0.984 & 0.035 & ${-0.004}$ & 0.095 & 0.072 & 97.1\% \\
MA-PointMap & with K & 0.921 & 0.104 & 3.7\% & 0.978 & 0.038 & ${-0.086}$ & 0.076 & 0.097 & 96.3\% \\
\addlinespace[1.5pt]
FoundationGeo & w/o K & 0.959 & 0.078 & 1.8\% & 0.985 & 0.035 & ${+0.022}$ & 0.080 & 0.062 & 98.2\% \\
FoundationGeo & with K & 0.971 & 0.067 & 0.5\% & 0.983 & 0.037 & ${-0.011}$ & 0.067 & 0.052 & 99.5\% \\
\addlinespace[1.5pt]
UniDepthV2 & w/o K & 0.928 & 0.106 & 4.0\% & 0.983 & 0.036 & ${+0.080}$ & 0.072 & 0.086 & 96.0\% \\
UniDepthV2 & with K & 0.960 & 0.078 & 1.7\% & 0.983 & 0.034 & ${+0.048}$ & 0.063 & 0.060 & 98.3\% \\
\addlinespace[1.5pt]
UniK3D & w/o K & 0.928 & 0.099 & 4.3\% & 0.984 & 0.034 & ${+0.066}$ & 0.082 & 0.079 & 95.7\% \\
UniK3D & with K & 0.970 & 0.065 & 1.1\% & 0.984 & 0.033 & ${+0.027}$ & 0.061 & 0.048 & 98.9\% \\
\bottomrule
\end{tabular*}
\end{table}

\begin{table}[htbp]
\centering
\renewcommand{\thetable}{F.2}
\caption{KITTI (outdoor driving, 652 images).}
\label{tab:F2}
\footnotesize
\setlength{\tabcolsep}{2.6pt}
\begin{tabular*}{\linewidth}{@{\extracolsep{\fill}}llrrrrrrrrr@{}}
\toprule
& & \multicolumn{3}{c}{Metric} & \multicolumn{2}{c}{Structure} & \multicolumn{4}{c}{Scale ($s^{*}$)} \\ \cmidrule(lr){3-5}\cmidrule(lr){6-7}\cmidrule(lr){8-11} Model & K & $\delta_1\uparrow$ & Rel$\downarrow$ & Fail.$\downarrow$ & $\delta_1\uparrow$ & Rel$\downarrow$ & Bias ($\to 0$) & Disp.$\downarrow$ & Abs.\ err.$\downarrow$ & $s^{*}\in[0.8,1.25]\uparrow$ \\
\midrule
Metric3D-v2 & with K & 0.980 & 0.052 & 0.0\% & 0.985 & 0.040 & ${-0.025}$ & 0.034 & 0.033 & 100.0\% \\
\addlinespace[1.5pt]
DA3METRIC & with K & 0.960 & 0.073 & 0.0\% & 0.969 & 0.056 & ${-0.036}$ & 0.041 & 0.045 & 100.0\% \\
\addlinespace[1.5pt]
Depth Pro & w/o K & 0.383 & 0.235 & 62.1\% & 0.964 & 0.055 & ${-0.265}$ & 0.154 & 0.268 & 38.0\% \\
Depth Pro & with K & 0.888 & 0.108 & 6.6\% & 0.964 & 0.055 & ${-0.020}$ & 0.137 & 0.088 & 93.6\% \\
\addlinespace[1.5pt]
MA-DepthMap & with K & 0.963 & 0.071 & 0.0\% & 0.979 & 0.055 & ${-0.020}$ & 0.055 & 0.044 & 100.0\% \\
\addlinespace[1.5pt]
DepthVLM & with K & 0.916 & 0.099 & 0.2\% & 0.928 & 0.074 & ${+0.041}$ & 0.050 & 0.053 & 99.8\% \\
\addlinespace[1.5pt]
MoGe-2 & w/o K & 0.636 & 0.179 & 35.0\% & 0.980 & 0.041 & ${-0.198}$ & 0.103 & 0.199 & 65.2\% \\
MoGe-2 & with K & 0.936 & 0.087 & 0.6\% & 0.947 & 0.078 & ${-0.000}$ & 0.050 & 0.036 & 99.7\% \\
\addlinespace[1.5pt]
MoGe-3 & w/o K & 0.553 & 0.197 & 43.1\% & 0.981 & 0.040 & ${-0.218}$ & 0.085 & 0.218 & 56.9\% \\
MoGe-3 & with K & 0.926 & 0.094 & 0.3\% & 0.939 & 0.086 & ${-0.005}$ & 0.047 & 0.035 & 99.8\% \\
\addlinespace[1.5pt]
MA-PointMap & w/o K & 0.931 & 0.091 & 3.8\% & 0.980 & 0.048 & ${-0.057}$ & 0.085 & 0.079 & 96.2\% \\
MA-PointMap & with K & 0.966 & 0.074 & 0.0\% & 0.976 & 0.052 & ${-0.044}$ & 0.046 & 0.053 & 100.0\% \\
\addlinespace[1.5pt]
FoundationGeo & w/o K & 0.961 & 0.072 & 0.6\% & 0.979 & 0.043 & ${-0.019}$ & 0.068 & 0.054 & 99.4\% \\
FoundationGeo & with K & 0.961 & 0.094 & 0.2\% & 0.969 & 0.059 & ${+0.061}$ & 0.054 & 0.069 & 99.8\% \\
\addlinespace[1.5pt]
UniDepthV2 & w/o K & 0.954 & 0.086 & 1.8\% & 0.981 & 0.042 & ${+0.038}$ & 0.073 & 0.064 & 98.2\% \\
UniDepthV2 & with K & 0.977 & 0.060 & 0.2\% & 0.980 & 0.042 & ${+0.025}$ & 0.033 & 0.034 & 99.8\% \\
\addlinespace[1.5pt]
UniK3D & w/o K & 0.958 & 0.095 & 1.2\% & 0.981 & 0.040 & ${+0.058}$ & 0.070 & 0.075 & 98.8\% \\
UniK3D & with K & 0.978 & 0.064 & 0.0\% & 0.981 & 0.041 & ${+0.035}$ & 0.032 & 0.040 & 100.0\% \\
\bottomrule
\end{tabular*}
\end{table}

\begin{table}[htbp]
\centering
\renewcommand{\thetable}{F.3}
\caption{ETH3D (building interiors and exteriors, reliefs and terrains, 454 images).}
\label{tab:F3}
\footnotesize
\setlength{\tabcolsep}{2.6pt}
\begin{tabular*}{\linewidth}{@{\extracolsep{\fill}}llrrrrrrrrr@{}}
\toprule
& & \multicolumn{3}{c}{Metric} & \multicolumn{2}{c}{Structure} & \multicolumn{4}{c}{Scale ($s^{*}$)} \\ \cmidrule(lr){3-5}\cmidrule(lr){6-7}\cmidrule(lr){8-11} Model & K & $\delta_1\uparrow$ & Rel$\downarrow$ & Fail.$\downarrow$ & $\delta_1\uparrow$ & Rel$\downarrow$ & Bias ($\to 0$) & Disp.$\downarrow$ & Abs.\ err.$\downarrow$ & $s^{*}\in[0.8,1.25]\uparrow$ \\
\midrule
Metric3D-v2 & with K & 0.888 & 0.118 & 7.9\% & 0.986 & 0.038 & ${+0.061}$ & 0.113 & 0.098 & 92.3\% \\
\addlinespace[1.5pt]
DA3METRIC & with K & 0.855 & 0.125 & 10.8\% & 0.981 & 0.048 & ${-0.087}$ & 0.136 & 0.117 & 89.6\% \\
\addlinespace[1.5pt]
Depth Pro & w/o K & 0.328 & 0.385 & 67.4\% & 0.966 & 0.072 & ${-0.446}$ & 0.548 & 0.534 & 32.8\% \\
Depth Pro & with K & 0.368 & 0.349 & 63.0\% & 0.966 & 0.073 & ${-0.491}$ & 0.525 & 0.522 & 37.2\% \\
\addlinespace[1.5pt]
MA-DepthMap & with K & 0.781 & 0.158 & 16.5\% & 0.979 & 0.058 & ${-0.027}$ & 0.169 & 0.141 & 83.5\% \\
\addlinespace[1.5pt]
DepthVLM & with K & 0.940 & 0.085 & 0.0\% & 0.950 & 0.075 & ${-0.006}$ & 0.048 & 0.037 & 100.0\% \\
\addlinespace[1.5pt]
MoGe-2 & w/o K & 0.912 & 0.103 & 6.6\% & 0.987 & 0.035 & ${-0.022}$ & 0.121 & 0.097 & 93.6\% \\
MoGe-2 & with K & 0.930 & 0.103 & 5.3\% & 0.985 & 0.038 & ${+0.007}$ & 0.113 & 0.091 & 95.2\% \\
\addlinespace[1.5pt]
MoGe-3 & w/o K & 0.882 & 0.100 & 9.7\% & 0.989 & 0.034 & ${-0.028}$ & 0.128 & 0.094 & 90.3\% \\
MoGe-3 & with K & 0.927 & 0.093 & 5.5\% & 0.986 & 0.038 & ${+0.015}$ & 0.110 & 0.080 & 94.5\% \\
\addlinespace[1.5pt]
MA-PointMap & w/o K & 0.896 & 0.107 & 8.6\% & 0.988 & 0.033 & ${-0.034}$ & 0.131 & 0.105 & 91.6\% \\
MA-PointMap & with K & 0.912 & 0.099 & 7.0\% & 0.988 & 0.035 & ${-0.047}$ & 0.115 & 0.096 & 93.4\% \\
\addlinespace[1.5pt]
FoundationGeo & w/o K & 0.803 & 0.145 & 18.3\% & 0.982 & 0.038 & ${-0.006}$ & 0.216 & 0.143 & 81.9\% \\
FoundationGeo & with K & 0.864 & 0.128 & 11.5\% & 0.982 & 0.039 & ${-0.007}$ & 0.206 & 0.127 & 88.5\% \\
\addlinespace[1.5pt]
UniDepthV2 & w/o K & 0.695 & 0.207 & 29.3\% & 0.992 & 0.031 & ${+0.092}$ & 0.202 & 0.179 & 70.7\% \\
UniDepthV2 & with K & 0.852 & 0.150 & 12.1\% & 0.992 & 0.030 & ${+0.062}$ & 0.156 & 0.133 & 87.9\% \\
\addlinespace[1.5pt]
UniK3D & w/o K & 0.837 & 0.149 & 14.8\% & 0.992 & 0.029 & ${+0.040}$ & 0.164 & 0.133 & 85.2\% \\
UniK3D & with K & 0.926 & 0.112 & 5.7\% & 0.992 & 0.029 & ${+0.033}$ & 0.118 & 0.100 & 94.3\% \\
\bottomrule
\end{tabular*}
\end{table}

\begin{table}[htbp]
\centering
\renewcommand{\thetable}{F.4}
\caption{iBims-1 (indoor, 100 images).}
\label{tab:F4}
\footnotesize
\setlength{\tabcolsep}{2.6pt}
\begin{tabular*}{\linewidth}{@{\extracolsep{\fill}}llrrrrrrrrr@{}}
\toprule
& & \multicolumn{3}{c}{Metric} & \multicolumn{2}{c}{Structure} & \multicolumn{4}{c}{Scale ($s^{*}$)} \\ \cmidrule(lr){3-5}\cmidrule(lr){6-7}\cmidrule(lr){8-11} Model & K & $\delta_1\uparrow$ & Rel$\downarrow$ & Fail.$\downarrow$ & $\delta_1\uparrow$ & Rel$\downarrow$ & Bias ($\to 0$) & Disp.$\downarrow$ & Abs.\ err.$\downarrow$ & $s^{*}\in[0.8,1.25]\uparrow$ \\
\midrule
Metric3D-v2 & with K & 0.941 & 0.100 & 1.0\% & 0.976 & 0.043 & ${+0.043}$ & 0.086 & 0.078 & 99.0\% \\
\addlinespace[1.5pt]
DA3METRIC & with K & 0.967 & 0.075 & 0.0\% & 0.976 & 0.039 & ${+0.002}$ & 0.078 & 0.057 & 100.0\% \\
\addlinespace[1.5pt]
Depth Pro & w/o K & 0.815 & 0.159 & 14.0\% & 0.975 & 0.041 & ${+0.033}$ & 0.195 & 0.132 & 86.0\% \\
Depth Pro & with K & 0.878 & 0.116 & 9.0\% & 0.975 & 0.041 & ${+0.033}$ & 0.124 & 0.102 & 91.0\% \\
\addlinespace[1.5pt]
MA-DepthMap & with K & 0.928 & 0.088 & 5.0\% & 0.987 & 0.037 & ${-0.050}$ & 0.099 & 0.080 & 95.0\% \\
\addlinespace[1.5pt]
DepthVLM & with K & 0.875 & 0.112 & 3.0\% & 0.924 & 0.087 & ${-0.023}$ & 0.087 & 0.073 & 98.0\% \\
\addlinespace[1.5pt]
MoGe-2 & w/o K & 0.827 & 0.138 & 15.0\% & 0.983 & 0.032 & ${-0.058}$ & 0.171 & 0.127 & 85.0\% \\
MoGe-2 & with K & 0.930 & 0.098 & 6.0\% & 0.979 & 0.038 & ${+0.004}$ & 0.124 & 0.079 & 94.0\% \\
\addlinespace[1.5pt]
MoGe-3 & w/o K & 0.770 & 0.151 & 22.0\% & 0.984 & 0.034 & ${-0.093}$ & 0.174 & 0.148 & 78.0\% \\
MoGe-3 & with K & 0.927 & 0.103 & 4.0\% & 0.967 & 0.040 & ${-0.009}$ & 0.130 & 0.086 & 96.0\% \\
\addlinespace[1.5pt]
MA-PointMap & w/o K & 0.848 & 0.147 & 14.0\% & 0.981 & 0.031 & ${-0.058}$ & 0.197 & 0.130 & 86.0\% \\
MA-PointMap & with K & 0.910 & 0.105 & 7.0\% & 0.980 & 0.036 & ${-0.031}$ & 0.131 & 0.093 & 93.0\% \\
\addlinespace[1.5pt]
FoundationGeo & w/o K & 0.917 & 0.103 & 5.0\% & 0.982 & 0.035 & ${-0.072}$ & 0.096 & 0.098 & 95.0\% \\
FoundationGeo & with K & 0.963 & 0.076 & 1.0\% & 0.978 & 0.039 & ${-0.000}$ & 0.078 & 0.059 & 99.0\% \\
\addlinespace[1.5pt]
UniDepthV2 & w/o K & 0.932 & 0.095 & 2.0\% & 0.981 & 0.035 & ${-0.036}$ & 0.117 & 0.082 & 98.0\% \\
UniDepthV2 & with K & 0.955 & 0.077 & 1.0\% & 0.981 & 0.034 & ${-0.003}$ & 0.099 & 0.059 & 99.0\% \\
\addlinespace[1.5pt]
UniK3D & w/o K & 0.936 & 0.093 & 3.0\% & 0.981 & 0.034 & ${+0.023}$ & 0.114 & 0.074 & 97.0\% \\
UniK3D & with K & 0.962 & 0.076 & 1.0\% & 0.981 & 0.033 & ${+0.013}$ & 0.095 & 0.058 & 99.0\% \\
\bottomrule
\end{tabular*}
\end{table}

\begin{table}[htbp]
\centering
\renewcommand{\thetable}{F.5}
\caption{DDAD (outdoor driving, multiple cameras, 1,000 images).}
\label{tab:F5}
\footnotesize
\setlength{\tabcolsep}{2.6pt}
\begin{tabular*}{\linewidth}{@{\extracolsep{\fill}}llrrrrrrrrr@{}}
\toprule
& & \multicolumn{3}{c}{Metric} & \multicolumn{2}{c}{Structure} & \multicolumn{4}{c}{Scale ($s^{*}$)} \\ \cmidrule(lr){3-5}\cmidrule(lr){6-7}\cmidrule(lr){8-11} Model & K & $\delta_1\uparrow$ & Rel$\downarrow$ & Fail.$\downarrow$ & $\delta_1\uparrow$ & Rel$\downarrow$ & Bias ($\to 0$) & Disp.$\downarrow$ & Abs.\ err.$\downarrow$ & $s^{*}\in[0.8,1.25]\uparrow$ \\
\midrule
Metric3D-v2 & with K & 0.938 & 0.091 & 0.4\% & 0.948 & 0.074 & ${-0.008}$ & 0.058 & 0.043 & 99.6\% \\
\addlinespace[1.5pt]
DA3METRIC & with K & 0.875 & 0.123 & 0.1\% & 0.880 & 0.116 & ${-0.022}$ & 0.039 & 0.035 & 100.0\% \\
\addlinespace[1.5pt]
Depth Pro & w/o K & 0.351 & 0.334 & 62.5\% & 0.846 & 0.139 & ${-0.391}$ & 0.360 & 0.400 & 38.6\% \\
Depth Pro & with K & 0.411 & 0.323 & 55.4\% & 0.846 & 0.139 & ${-0.375}$ & 0.390 & 0.384 & 45.3\% \\
\addlinespace[1.5pt]
MA-DepthMap & with K & 0.871 & 0.138 & 3.2\% & 0.933 & 0.098 & ${-0.093}$ & 0.065 & 0.098 & 97.0\% \\
\addlinespace[1.5pt]
DepthVLM & with K & 0.721 & 0.182 & 14.0\% & 0.828 & 0.144 & ${-0.069}$ & 0.129 & 0.119 & 89.6\% \\
\addlinespace[1.5pt]
MoGe-2 & w/o K & 0.730 & 0.157 & 19.6\% & 0.919 & 0.088 & ${-0.056}$ & 0.158 & 0.122 & 80.5\% \\
MoGe-2 & with K & 0.858 & 0.130 & 1.1\% & 0.880 & 0.108 & ${+0.013}$ & 0.074 & 0.057 & 99.0\% \\
\addlinespace[1.5pt]
MoGe-3 & w/o K & 0.719 & 0.163 & 21.1\% & 0.918 & 0.088 & ${-0.051}$ & 0.168 & 0.128 & 79.2\% \\
MoGe-3 & with K & 0.852 & 0.133 & 1.9\% & 0.875 & 0.110 & ${+0.019}$ & 0.075 & 0.059 & 98.8\% \\
\addlinespace[1.5pt]
MA-PointMap & w/o K & 0.900 & 0.123 & 1.1\% & 0.929 & 0.087 & ${+0.005}$ & 0.088 & 0.069 & 99.0\% \\
MA-PointMap & with K & 0.913 & 0.113 & 0.6\% & 0.926 & 0.090 & ${+0.006}$ & 0.065 & 0.050 & 99.7\% \\
\addlinespace[1.5pt]
FoundationGeo & w/o K & 0.744 & 0.181 & 17.0\% & 0.920 & 0.084 & ${+0.055}$ & 0.154 & 0.143 & 84.1\% \\
FoundationGeo & with K & 0.872 & 0.126 & 2.3\% & 0.905 & 0.097 & ${+0.025}$ & 0.080 & 0.064 & 98.2\% \\
\addlinespace[1.5pt]
UniDepthV2 & w/o K & 0.777 & 0.183 & 13.8\% & 0.924 & 0.109 & ${-0.061}$ & 0.137 & 0.129 & 86.5\% \\
UniDepthV2 & with K & 0.895 & 0.140 & 0.9\% & 0.926 & 0.108 & ${-0.073}$ & 0.056 & 0.078 & 99.1\% \\
\addlinespace[1.5pt]
UniK3D & w/o K & 0.757 & 0.202 & 15.7\% & 0.926 & 0.115 & ${-0.027}$ & 0.157 & 0.135 & 84.5\% \\
UniK3D & with K & 0.907 & 0.142 & 0.6\% & 0.926 & 0.116 & ${-0.060}$ & 0.055 & 0.067 & 99.4\% \\
\bottomrule
\end{tabular*}
\end{table}

\begin{table}[htbp]
\centering
\renewcommand{\thetable}{F.6}
\caption{DIODE (mixed indoor and outdoor, 771 images).}
\label{tab:F6}
\footnotesize
\setlength{\tabcolsep}{2.6pt}
\begin{tabular*}{\linewidth}{@{\extracolsep{\fill}}llrrrrrrrrr@{}}
\toprule
& & \multicolumn{3}{c}{Metric} & \multicolumn{2}{c}{Structure} & \multicolumn{4}{c}{Scale ($s^{*}$)} \\ \cmidrule(lr){3-5}\cmidrule(lr){6-7}\cmidrule(lr){8-11} Model & K & $\delta_1\uparrow$ & Rel$\downarrow$ & Fail.$\downarrow$ & $\delta_1\uparrow$ & Rel$\downarrow$ & Bias ($\to 0$) & Disp.$\downarrow$ & Abs.\ err.$\downarrow$ & $s^{*}\in[0.8,1.25]\uparrow$ \\
\midrule
Metric3D-v2 & with K & 0.020 & 0.491 & 99.9\% & 0.984 & 0.032 & ${+0.400}$ & 0.061 & 0.400 & 0.3\% \\
\addlinespace[1.5pt]
DA3METRIC & with K & 0.766 & 0.152 & 20.9\% & 0.949 & 0.063 & ${-0.061}$ & 0.200 & 0.142 & 79.6\% \\
\addlinespace[1.5pt]
Depth Pro & w/o K & 0.377 & 0.319 & 62.9\% & 0.936 & 0.065 & ${-0.337}$ & 0.427 & 0.413 & 37.2\% \\
Depth Pro & with K & 0.431 & 0.296 & 57.7\% & 0.936 & 0.064 & ${-0.289}$ & 0.418 & 0.374 & 42.7\% \\
\addlinespace[1.5pt]
MA-DepthMap & with K & 0.735 & 0.179 & 23.7\% & 0.961 & 0.059 & ${+0.075}$ & 0.168 & 0.149 & 76.5\% \\
\addlinespace[1.5pt]
DepthVLM & with K & 0.546 & 0.265 & 44.0\% & 0.842 & 0.129 & ${+0.028}$ & 0.267 & 0.211 & 59.9\% \\
\addlinespace[1.5pt]
MoGe-2 & w/o K & 0.660 & 0.175 & 32.2\% & 0.957 & 0.050 & ${-0.145}$ & 0.204 & 0.186 & 68.6\% \\
MoGe-2 & with K & 0.774 & 0.161 & 19.1\% & 0.941 & 0.056 & ${+0.005}$ & 0.205 & 0.143 & 81.3\% \\
\addlinespace[1.5pt]
MoGe-3 & w/o K & 0.684 & 0.170 & 30.4\% & 0.955 & 0.051 & ${-0.133}$ & 0.199 & 0.176 & 70.6\% \\
MoGe-3 & with K & 0.816 & 0.149 & 15.2\% & 0.937 & 0.060 & ${+0.027}$ & 0.183 & 0.125 & 85.7\% \\
\addlinespace[1.5pt]
MA-PointMap & w/o K & 0.728 & 0.165 & 24.1\% & 0.956 & 0.048 & ${-0.081}$ & 0.205 & 0.166 & 76.4\% \\
MA-PointMap & with K & 0.784 & 0.155 & 18.0\% & 0.954 & 0.049 & ${+0.002}$ & 0.195 & 0.139 & 82.4\% \\
\addlinespace[1.5pt]
FoundationGeo & w/o K & 0.617 & 0.186 & 37.6\% & 0.954 & 0.053 & ${-0.163}$ & 0.221 & 0.202 & 63.0\% \\
FoundationGeo & with K & 0.751 & 0.162 & 21.9\% & 0.929 & 0.065 & ${+0.007}$ & 0.212 & 0.141 & 79.2\% \\
\addlinespace[1.5pt]
UniDepthV2 & w/o K & 0.517 & 0.430 & 46.6\% & 0.956 & 0.055 & ${+0.042}$ & 0.429 & 0.310 & 54.0\% \\
UniDepthV2 & with K & 0.670 & 0.410 & 31.3\% & 0.954 & 0.054 & ${+0.168}$ & 0.385 & 0.261 & 69.5\% \\
\addlinespace[1.5pt]
UniK3D & w/o K & 0.730 & 0.181 & 24.5\% & 0.962 & 0.048 & ${-0.058}$ & 0.213 & 0.162 & 75.7\% \\
UniK3D & with K & 0.807 & 0.162 & 17.6\% & 0.960 & 0.047 & ${+0.047}$ & 0.184 & 0.129 & 83.0\% \\
\bottomrule
\end{tabular*}
\end{table}

\begin{table}[htbp]
\centering
\renewcommand{\thetable}{F.7}
\caption{HAMMER (close-range tabletop, 775 images).}
\label{tab:F7}
\footnotesize
\setlength{\tabcolsep}{2.6pt}
\begin{tabular*}{\linewidth}{@{\extracolsep{\fill}}llrrrrrrrrr@{}}
\toprule
& & \multicolumn{3}{c}{Metric} & \multicolumn{2}{c}{Structure} & \multicolumn{4}{c}{Scale ($s^{*}$)} \\ \cmidrule(lr){3-5}\cmidrule(lr){6-7}\cmidrule(lr){8-11} Model & K & $\delta_1\uparrow$ & Rel$\downarrow$ & Fail.$\downarrow$ & $\delta_1\uparrow$ & Rel$\downarrow$ & Bias ($\to 0$) & Disp.$\downarrow$ & Abs.\ err.$\downarrow$ & $s^{*}\in[0.8,1.25]\uparrow$ \\
\midrule
Metric3D-v2 & with K & 0.443 & 0.357 & 52.4\% & 0.990 & 0.042 & ${+0.282}$ & 0.176 & 0.282 & 47.6\% \\
\addlinespace[1.5pt]
DA3METRIC & with K & 0.660 & 0.275 & 31.1\% & 0.994 & 0.040 & ${+0.219}$ & 0.190 & 0.219 & 68.9\% \\
\addlinespace[1.5pt]
Depth Pro & w/o K & 0.630 & 0.391 & 35.4\% & 0.989 & 0.044 & ${+0.220}$ & 0.353 & 0.255 & 64.6\% \\
Depth Pro & with K & 0.717 & 0.281 & 25.4\% & 0.989 & 0.044 & ${+0.179}$ & 0.270 & 0.203 & 74.6\% \\
\addlinespace[1.5pt]
MA-DepthMap & with K & 0.677 & 0.170 & 26.7\% & 0.992 & 0.051 & ${-0.136}$ & 0.139 & 0.178 & 73.3\% \\
\addlinespace[1.5pt]
DepthVLM & with K & 0.160 & 1.233 & 81.5\% & 0.707 & 0.183 & ${+0.677}$ & 0.406 & 0.677 & 18.5\% \\
\addlinespace[1.5pt]
MoGe-2 & w/o K & 0.660 & 0.271 & 33.3\% & 0.992 & 0.040 & ${+0.005}$ & 0.322 & 0.230 & 66.7\% \\
MoGe-2 & with K & 0.740 & 0.308 & 23.9\% & 0.990 & 0.042 & ${+0.199}$ & 0.300 & 0.204 & 76.1\% \\
\addlinespace[1.5pt]
MoGe-3 & w/o K & 0.608 & 0.248 & 39.1\% & 0.993 & 0.041 & ${-0.049}$ & 0.294 & 0.233 & 60.9\% \\
MoGe-3 & with K & 0.705 & 0.284 & 26.3\% & 0.991 & 0.042 & ${+0.188}$ & 0.267 & 0.198 & 73.7\% \\
\addlinespace[1.5pt]
MA-PointMap & w/o K & 0.746 & 0.246 & 24.0\% & 0.994 & 0.032 & ${+0.099}$ & 0.288 & 0.190 & 76.0\% \\
MA-PointMap & with K & 0.719 & 0.304 & 26.8\% & 0.993 & 0.039 & ${+0.207}$ & 0.278 & 0.213 & 73.2\% \\
\addlinespace[1.5pt]
FoundationGeo & w/o K & 0.717 & 0.181 & 27.2\% & 0.988 & 0.044 & ${-0.001}$ & 0.225 & 0.161 & 72.8\% \\
FoundationGeo & with K & 0.638 & 0.315 & 30.7\% & 0.990 & 0.040 & ${+0.207}$ & 0.258 & 0.239 & 69.3\% \\
\addlinespace[1.5pt]
UniDepthV2 & w/o K & 0.468 & 0.381 & 53.8\% & 0.992 & 0.035 & ${+0.290}$ & 0.227 & 0.291 & 46.2\% \\
UniDepthV2 & with K & 0.471 & 0.377 & 53.4\% & 0.993 & 0.034 & ${+0.289}$ & 0.220 & 0.290 & 46.6\% \\
\addlinespace[1.5pt]
UniK3D & w/o K & 0.583 & 0.305 & 41.4\% & 0.995 & 0.032 & ${+0.223}$ & 0.241 & 0.230 & 58.6\% \\
UniK3D & with K & 0.570 & 0.315 & 42.8\% & 0.995 & 0.031 & ${+0.235}$ & 0.230 & 0.240 & 57.2\% \\
\bottomrule
\end{tabular*}
\end{table}

\begin{figure}[htbp]
\centering
\renewcommand{\thefigure}{F.8}

\includegraphics[width=0.90\linewidth]{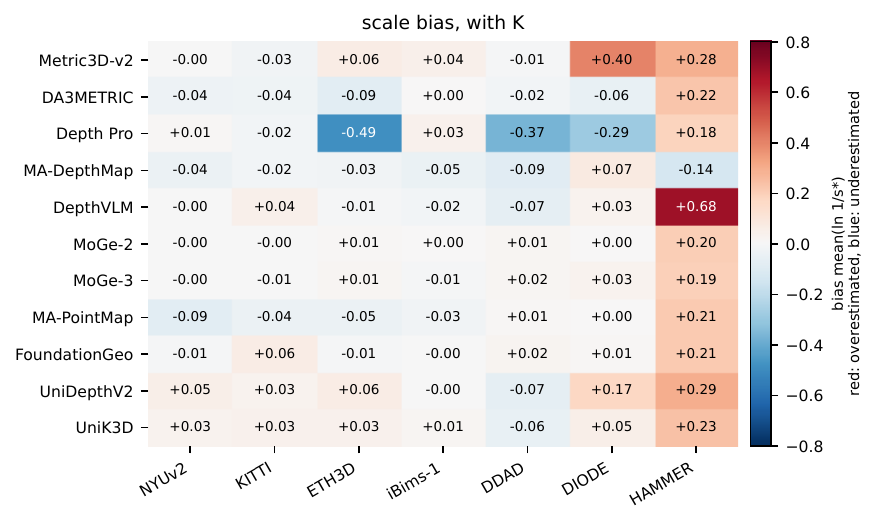}
\caption{Scale bias $\operatorname{mean}(\ln (1/s^{*}))$ (Section~\hyperref[sec:3.3]{3.3}) of each architecture (with intrinsics) on each benchmark; the values equal the ``Bias'' columns of the tables in Appendix~\hyperref[app:F]{F}. Red: overestimated depth (positive); blue: underestimated depth (negative).}
\label{fig:F8}
\end{figure}

\textbf{Scale bias by dataset.} Figure~\hyperref[fig:F8]{F.8} summarises the per-benchmark bias of each architecture (with intrinsics) from Tables~\hyperref[tab:F1]{F.1} to F.7, showing which directions most architectures share.

\begin{figure}[tbp]
\centering
\renewcommand{\thefigure}{F.9}
\captionsetup{font=footnotesize}
\includegraphics[width=1.00\linewidth,height=0.76\textheight,keepaspectratio]{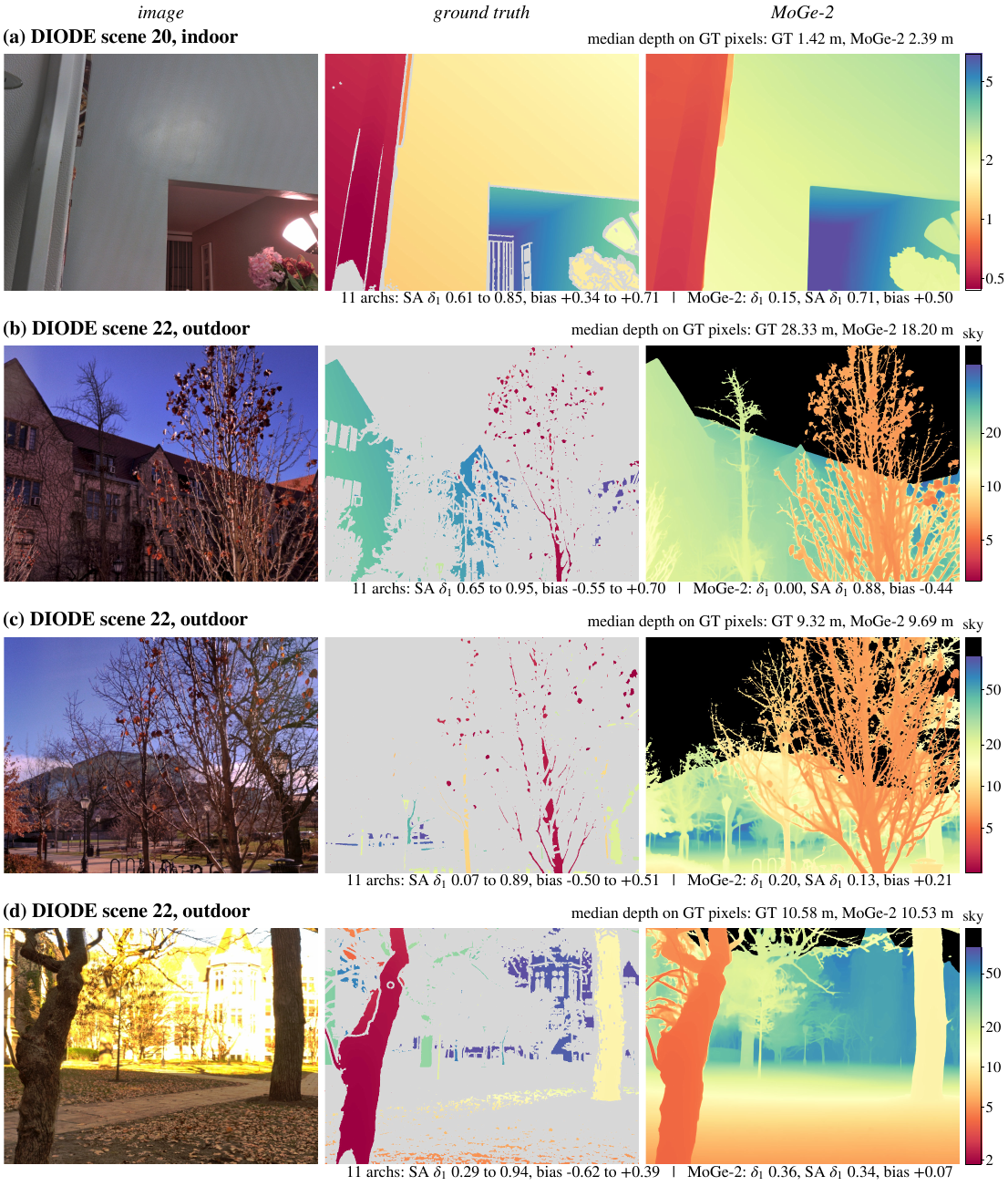}
\caption{Images failed by all 11 architectures with flawed structure (with intrinsics), all from DIODE. Same layout as Figure~\hyperref[fig:4]{4}; (b) has only a slight structural error (scale-aligned $\delta_1$ of MoGe-2 0.88).}
\label{fig:F9}
\end{figure}

\begin{figure}[tbp]
\centering
\renewcommand{\thefigure}{F.10}
\captionsetup{font=footnotesize}
\includegraphics[width=1.00\linewidth,height=0.89\textheight,keepaspectratio]{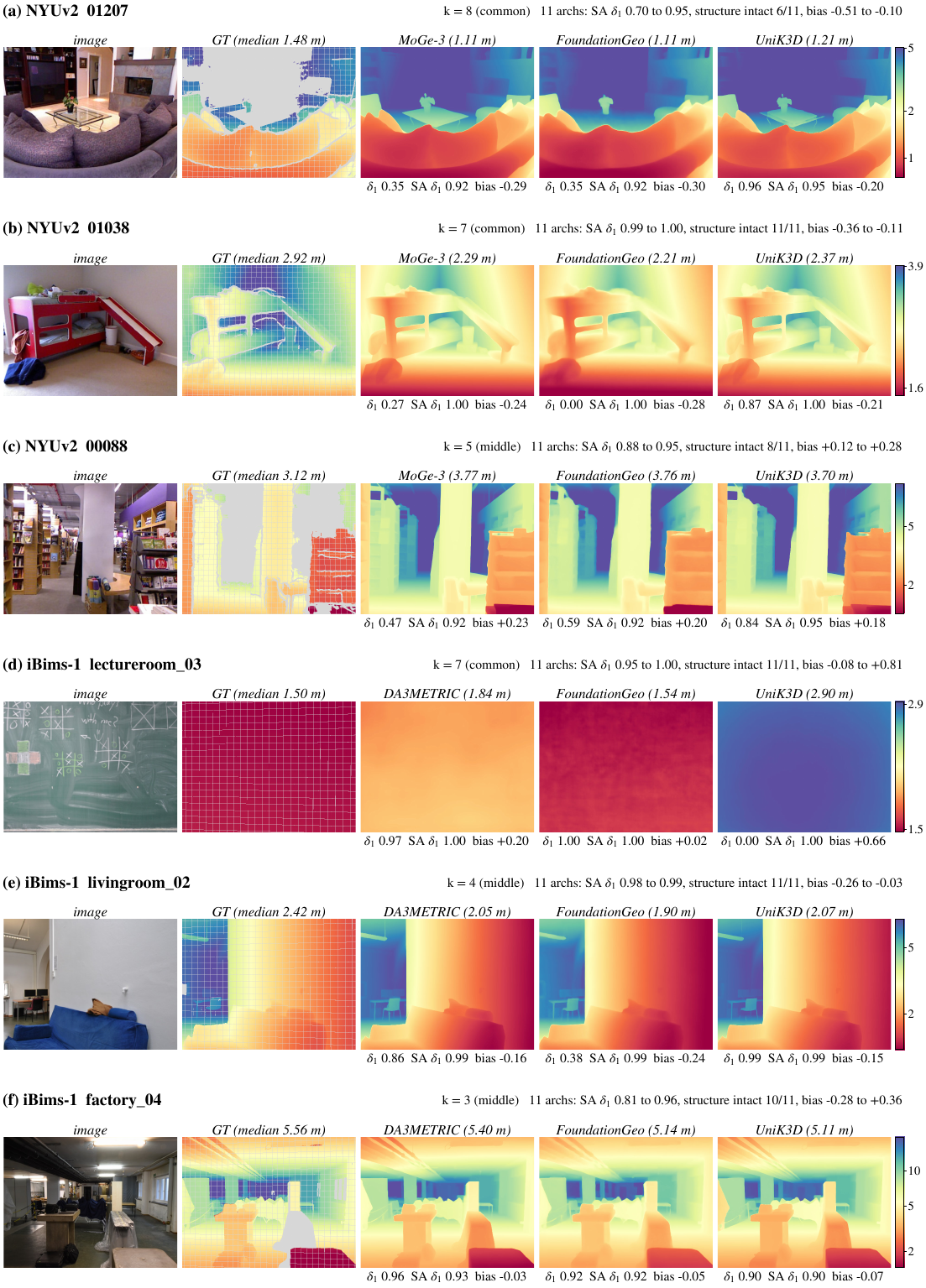}
\caption{Images of NYUv2 and iBims-1 failed by the most architectures (with intrinsics). Each row shows the image, the ground truth and the predictions of the 3 architectures with the highest mean metric $\delta_1$ on that dataset, sharing a logarithmic colour scale per row (in metres); parentheses give the median depth over pixels with valid ground truth. The right side of each row title gives the number of failing architectures $k$, and the range of scale-aligned $\delta_1$, the number of architectures with intact structure (scale-aligned $\delta_1$ $\geq$ 0.9) and the range of bias over the 11 architectures; below each prediction are its metric $\delta_1$, scale-aligned $\delta_1$ (SA) and bias. Grey: no ground truth; black: sky predicted by the model. (c), (e), (f): intermediate tier.}
\label{fig:F10}
\end{figure}

\FloatBarrier
\textbf{Hard images shared by most models, per dataset.} Tables~\hyperref[tab:F1]{F.1} to F.7 give only means. To show where the difficulty of each dataset lies, Figures~\hyperref[fig:F10]{F.10} to F.13 select, in each dataset, the images failed by the most architectures, together with the predictions of the 3 architectures with the highest mean metric $\delta_1$ on that dataset (with intrinsics). The selection rule is as follows. We take 3 images per dataset; ETH3D and DIODE contain both indoor and outdoor scenes, so we take 3 indoor and 3 outdoor images from each; KITTI has no shared failure and only 1 image in the intermediate tier, so it contributes 1 image. Within each group, images are taken from the shared failures ($k$ $\geq$ 6, Section~\hyperref[sec:5]{5}) in order of decreasing $k$ and increasing mean metric $\delta_1$ over the 11 architectures, at most one per scene, and filled from the intermediate tier ($k$ from 3 to 5) when needed. Three adjustments were made. The HAMMER empty-table images contain no object that reveals size and are already shown in Figure~\hyperref[fig:4]{4}, so here we select only from the 404 images with objects. The DIODE images failed by all 11 architectures come from indoor scene\_00020 and outdoor scene\_00022 and are already shown in Figure~\hyperref[fig:4]{4} and Figure~\hyperref[fig:F9]{F.9}, so here we select from the other scenes; only 2 indoor scenes remain, so the third indoor image comes from another scan of one of them. The 3 shared-failure images of DDAD are adjacent clips of the same open lot, all taken by CAMERA\_06, so for DDAD we treat each camera as a scene and take the other two images from the intermediate tier.

\textbf{The 4 all-failed DIODE images with flawed structure.} Of the 6 DIODE images failed by all 11 architectures (Section~\hyperref[sec:5.2]{5.2}), 4 have flawed structure (Figure~\hyperref[fig:F9]{F.9}), each in a different way. (a) is an indoor scene dominated by a large textureless white wall, with a small structural error: the scale-aligned $\delta_1$ of MoGe-2 is 0.71, and its predicted depth range (ratio of the 90th to the 10th percentile depth) is 7.5, close to the ground-truth 9.1. (b) to (d) are outdoor scenes with buildings in the background and bare trees in the foreground. Figure~\hyperref[fig:F9]{F.9} shows that MoGe-2 predicts the outlines of both the trees and the buildings accurately; what it gets wrong is the relative distance between trees and buildings: it places the buildings too close to the trees, compressing the depth range of the whole image, with ground-truth depth ranges of 9.6, 16.9 and 37.5 against only 3.3, 4.6 and 11.8 for MoGe-2. In (b) the ground truth covers only 16\% of the pixels, mostly on the building, so this compression matters little for $\delta_1$: the scale-aligned $\delta_1$ is 0.88 for MoGe-2 and 0.65 to 0.95 over the 11 architectures, a slight structural error. In such scenes trees can be of any size, there is no directly comparable reference between the foreground trees and the background buildings, and the background of (d) is also strongly overexposed; but none of these image factors alone explains the failure: Figure~\hyperref[fig:4]{4}(d) and (e) also show trees in front of buildings without visible ground, yet MoGe-2 gets their structure right, while Figure~\hyperref[fig:F9]{F.9}(d) shows a fair amount of ground. The failure lies in the relative distance between foreground and background. Scale alignment multiplies by a single global factor and cannot compensate for such an error, so the scale-aligned $\delta_1$ of (c) and (d) is very low (0.13 and 0.34). Images on which most models get the relative depth wrong do exist, but they are rare: structure failures make up only 1.7\% of metric failures (Section~\hyperref[sec:4.3]{4.3}). The images that all models fail are therefore hard for different reasons: these DIODE images are hard in structure, while the HAMMER empty-table images have intact structure and are hard because the image contains almost no object whose size reveals the distance.

\begin{figure}[tbp]
\centering
\renewcommand{\thefigure}{F.11}
\captionsetup{font=footnotesize}
\includegraphics[width=1.00\linewidth,height=0.95\textheight,keepaspectratio]{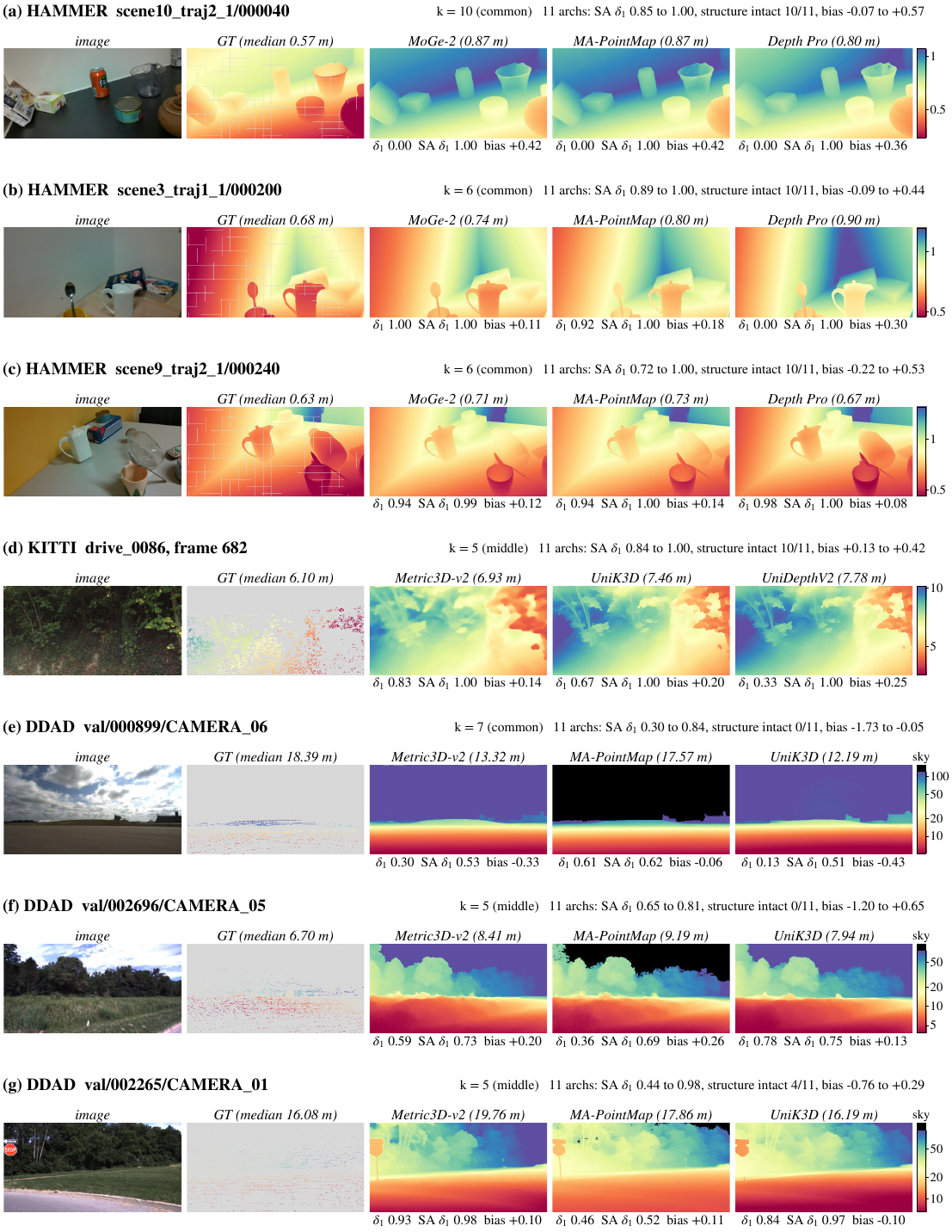}
\caption{Images of HAMMER (only images with objects), KITTI and DDAD failed by the most architectures; same layout as Figure~\hyperref[fig:F10]{F.10}. KITTI has no shared failure; (d) of KITTI and (f), (g) of DDAD come from the intermediate tier.}
\label{fig:F11}
\end{figure}

\begin{figure}[tbp]
\centering
\renewcommand{\thefigure}{F.12}
\captionsetup{font=footnotesize}
\includegraphics[width=1.00\linewidth,height=0.95\textheight,keepaspectratio]{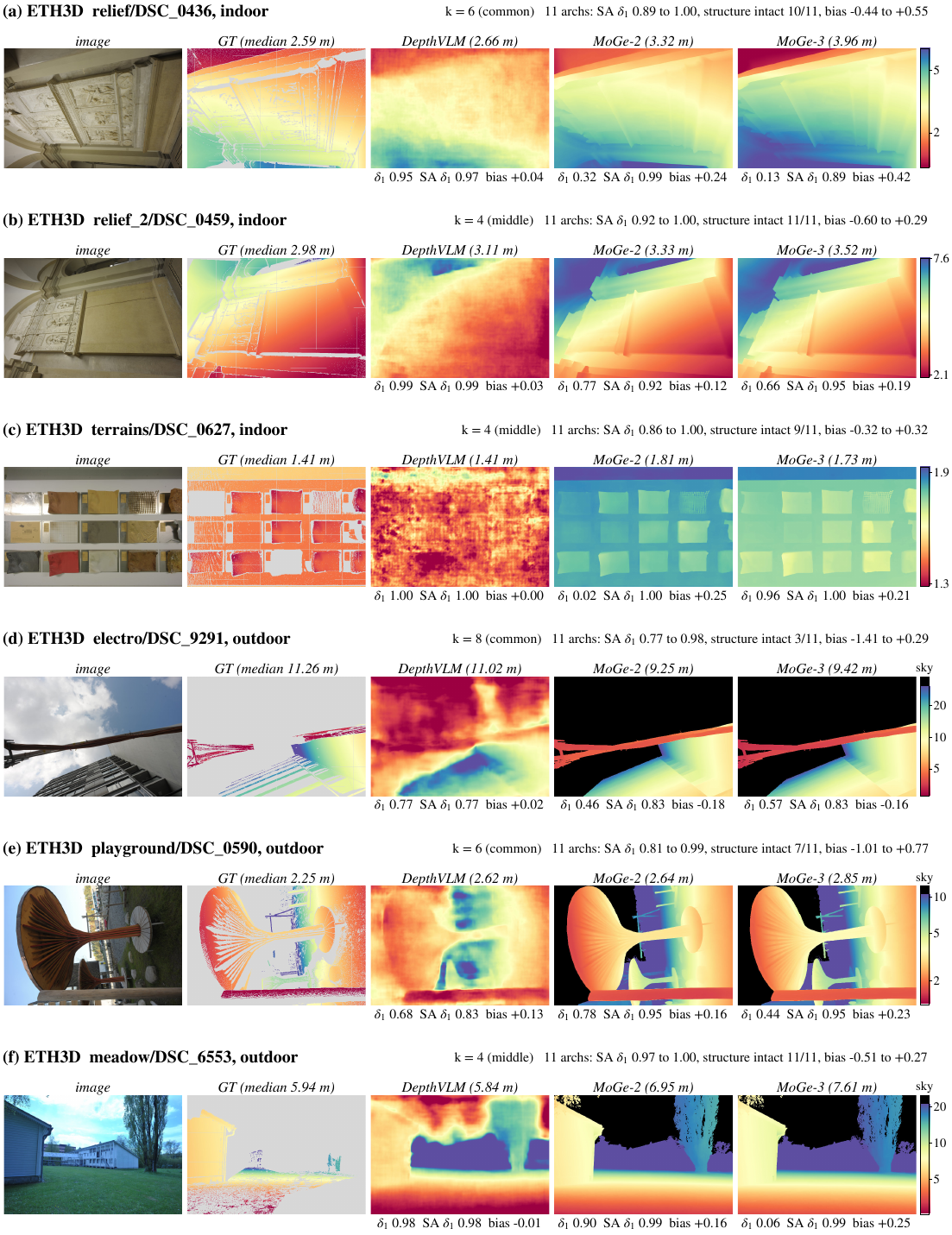}
\caption{Images of ETH3D failed by the most architectures, 3 indoor ((a) to (c)) and 3 outdoor ((d) to (f)); same layout as Figure~\hyperref[fig:F10]{F.10}. (b), (c) and (f) come from the intermediate tier.}
\label{fig:F12}
\end{figure}

\begin{figure}[tbp]
\centering
\renewcommand{\thefigure}{F.13}
\captionsetup{font=footnotesize}
\includegraphics[width=1.00\linewidth,height=0.93\textheight,keepaspectratio]{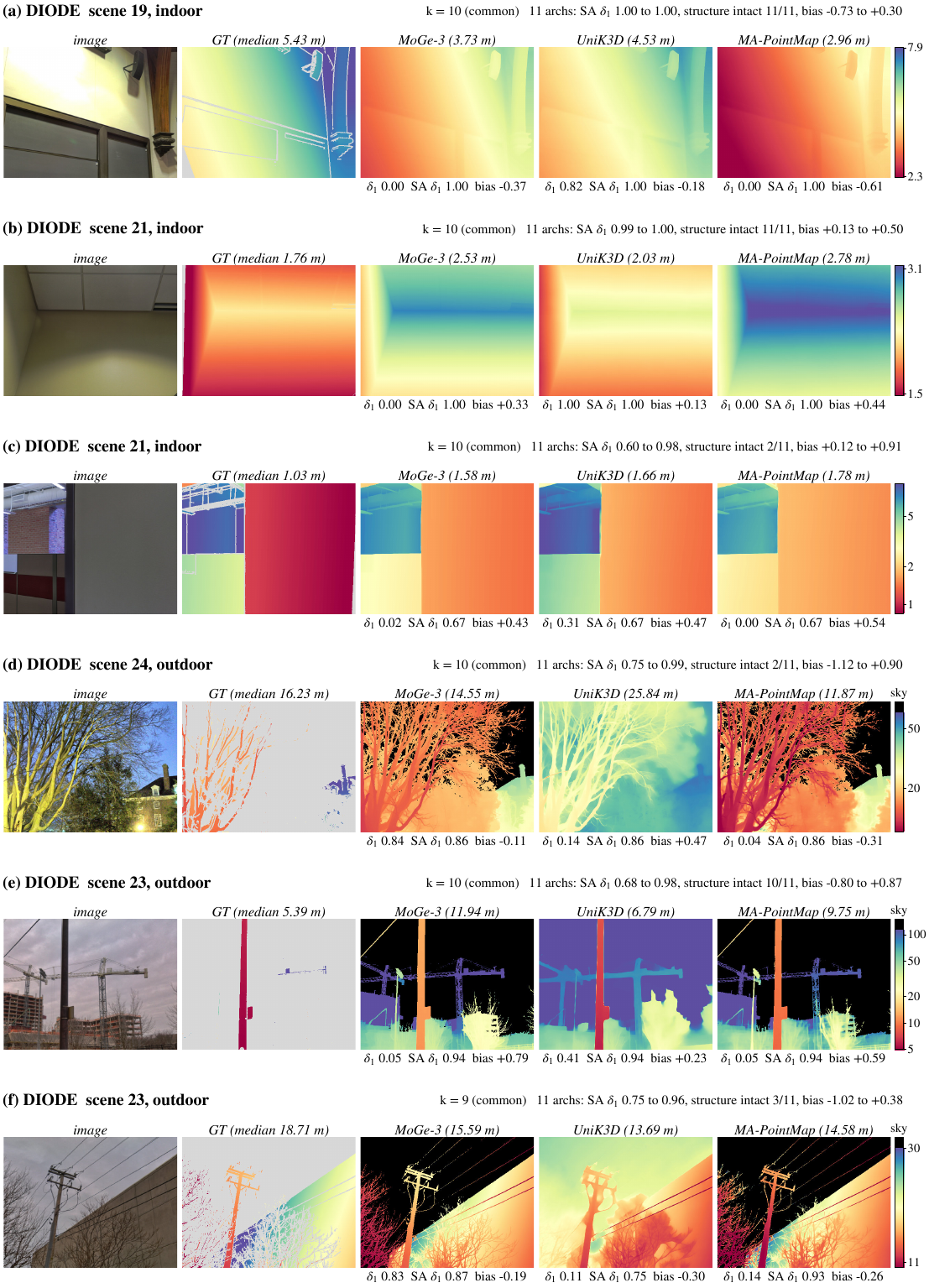}
\caption{Images of DIODE failed by the most architectures, 3 indoor ((a) to (c)) and 3 outdoor ((d) to (f)); same layout as Figure~\hyperref[fig:F10]{F.10}, all from scenes other than those of Figure~\hyperref[fig:4]{4} and Figure~\hyperref[fig:F9]{F.9}.}
\label{fig:F13}
\end{figure}

The difficulty differs between datasets.

\begin{itemize}
\item \emph{Intact structure, wrong scale only.} NYUv2, iBims-1, HAMMER and KITTI belong to this kind, as do most hard images of the indoor parts of ETH3D and DIODE. On these images most of the 11 architectures have intact structure (at least 6, except the third indoor DIODE image), and on each image most or all architectures err in the same direction.
\begin{itemize}
\item NYUv2 (Figure~\hyperref[fig:F10]{F.10}(a) to (c)): on the first two images all 11 architectures underestimate depth (bias -0.51 to -0.10 and -0.36 to -0.11); the third is a bookstore aisle on which all overestimate depth (+0.12 to +0.28). Even MoGe-3, with the highest mean on NYUv2, underestimates their depth by a factor of about 1.3.
\item iBims-1 (Figure~\hyperref[fig:F10]{F.10}(d) to (f)): at least 10 of the 11 architectures have intact structure on each. lectureroom\_03 is a blackboard facing the camera, showing only one plane and chalk writing, with biases from -0.08 to +0.81; livingroom\_02 shows a white wall and a sofa, with all architectures underestimating depth; factory\_04 is a deep factory hall, with 9 architectures underestimating depth and biases from -0.28 to +0.36.
\item HAMMER (Figure~\hyperref[fig:F11]{F.11}(a) to (c)): all three are tables with everyday objects such as cans, tins, mugs and teapots, failed by 10, 6 and 6 architectures; on each, 10 architectures have intact structure and 10 overestimate depth. Of the shared failures, 49 have objects on the table, 40 of them from scene10. So even with objects of known size in view, most architectures still overestimate the depth of the close-range HAMMER tables.
\item KITTI (Figure~\hyperref[fig:F11]{F.11}(d)): bushes close to the road; 10 architectures have intact structure and all overestimate depth (+0.13 to +0.42); the ground truth covers only 13\% of the pixels.
\item ETH3D indoor (Figure~\hyperref[fig:F12]{F.12}(a) to (c)): relief and relief\_2 are relief walls and terrains is a display of material samples; 9 to 11 architectures have intact structure, most overestimate depth, and the biases lie between -0.60 and +0.55. On these three images DepthVLM reaches a metric $\delta_1$ of at least 0.95, while MoGe-2 and MoGe-3 fail on some.
\item DIODE indoor (Figure~\hyperref[fig:F13]{F.13}(a) to (c)): each is failed by 10 architectures. The scene\_00019 image shows a wall and window frames at an oblique angle, with all 11 architectures structurally intact and 10 underestimating depth; the two scene\_00021 images show a room corner and a nearby wall, with all architectures overestimating depth. On the wall image ((c)) only 2 architectures have intact structure: the distance between the near wall and the far brick wall and ceiling is estimated wrongly.
\end{itemize}
\item \emph{Flawed structure.} The hard images of DDAD belong to this kind (Table~\hyperref[tab:4]{4}: flawed structure accounts for 48.1\% of the DDAD failures, against 5\% to 23\% elsewhere).
\begin{itemize}
\item DDAD (Figure~\hyperref[fig:F11]{F.11}(e) to (g)): all show open ground, with 0, 0 and 4 architectures structurally intact; on 000899 all 11 underestimate depth (bias $-$1.73 to $-$0.05), so the direction is shared but the structure is not. The ground truth covers only 2.3\% to 3.6\% of the pixels, all on the ground extending from near to far in the lower part of the image (for 000899, the 10th to 90th percentiles of ground-truth depth are 8.4 to 112.7 m). The ground occupies only a few compressed rows in the image yet spans an order of magnitude in distance, and all 11 architectures get the depth variation within this narrow band wrong.
\end{itemize}
\item \emph{Mixed.} The outdoor examples of ETH3D and DIODE are mixed: electro and two of the DIODE images have flawed structure, while playground (7 of 11 intact) and DIODE (e) (10 of 11 intact) do not; over all shared failures on outdoor DIODE, 72\% have intact structure for at least 6 architectures.
\begin{itemize}
\item ETH3D outdoor (Figure~\hyperref[fig:F12]{F.12}(d) to (f)): electro is an upward view of a building corner with a tilted frame; only 3 architectures have intact structure, 4 overestimate and 7 underestimate depth, with biases from -1.41 to +0.29. playground shows a sculpture in the foreground and a field behind; 7 architectures have intact structure, with biases from -1.01 to +0.77, so it is closer to the first kind. meadow shows houses and a lawn; all 11 architectures have intact structure and 9 overestimate depth, so it belongs to the first kind.
\item DIODE outdoor (Figure~\hyperref[fig:F13]{F.13}(d) to (f)): trees and utility poles in the foreground, buildings or cranes in the background. The ground truth covers only 4\% to 30\% of the pixels, mostly on the nearby branches and poles. 2, 10 ((e)) and 3 architectures have intact structure, and the bias range exceeds 1 on each image (-1.12 to +0.90, -0.80 to +0.87, -1.02 to +0.38).
\end{itemize}
\end{itemize}

The mean errors of different datasets therefore hide different difficulties. On most datasets, including the outdoor parts of ETH3D and DIODE for most of their hard images, the hard images have correct structure and are hard in scale; DDAD is the exception, where the hard images have flawed structure.

\section{Sensitivity to the Thresholds and Supplementary Analyses}\label{app:G}

The main text uses three thresholds: metric failure (metric $\delta_1$ $<$ 0.5), intact structure (scale-aligned $\delta_1$ $\geq$ 0.9) and structure failure (scale-aligned $\delta_1$ $<$ 0.5, the same value as for metric failure). None was tuned on the results. Here we vary them on the existing per-image results and recount, without new inference. The criterion we use to judge whether the conclusions hold is that the qualitative conclusions of the main text hold as long as pure scale failures are a majority of metric failures and structure failures a minority for every value, and DDAD and DIODE remain the two lowest datasets; the exact percentages vary with the thresholds, and the numbers cited in the main text correspond only to the values 0.5 and 0.9. A second way of splitting the error that does not depend on the $\delta_1$ threshold, and its results under $\delta_2$ and $\delta_3$, are given in Appendix~\hyperref[app:B]{B}.

\begin{table}[htbp]
\centering
\renewcommand{\thetable}{G.1}
\caption{Shares of metric failures with intact structure, with slightly flawed structure, and that are also structure failures (all ``image $\times$ model evaluation'' pairs of the 18 model evaluations pooled). The structure failure threshold takes the same value as the metric failure threshold; ``slightly flawed'': scale-aligned $\delta_1$ from that threshold to 0.9.}
\label{tab:G1}
\small
\setlength{\tabcolsep}{4pt}
\begin{tabular*}{\linewidth}{@{\extracolsep{\fill}}lrrrrrrr@{}}
\toprule
\makecell{Failure\\threshold} & Failures & \makecell{Intact\\$\geq$ 0.8} & \makecell{Intact\\$\geq$ 0.85} & \makecell{Intact\\$\geq$ 0.9} & \makecell{Intact\\$\geq$ 0.95} & \makecell{Slightly\\flawed} & \makecell{Also struct.\\failure} \\
\midrule
$<$ 0.3 & 12,767 & 89.4\% & 86.0\% & 80.8\% & 71.4\% & 19.0\% & 0.1\% \\
$<$ 0.4 & 13,871 & 88.6\% & 85.0\% & 79.7\% & 70.1\% & 19.5\% & 0.8\% \\
$<$ 0.5 & 15,085 & 87.9\% & 84.2\% & 78.8\% & 69.1\% & 19.5\% & 1.7\% \\
$<$ 0.6 & 16,675 & 87.0\% & 83.0\% & 77.3\% & 67.5\% & 18.5\% & 4.2\% \\
$<$ 0.7 & 18,805 & 85.2\% & 80.8\% & 74.8\% & 64.7\% & 16.4\% & 8.8\% \\
\bottomrule
\end{tabular*}
\end{table}

Over the 20 combinations, the share of metric failures with intact structure stays between 65\% and 89\%, always a majority, and the share that are also structure failures stays between 0.1\% and 8.8\%, always a minority. The latter is more sensitive to the threshold: when the metric failure threshold is relaxed to 0.7, the structure failure threshold rises with it, some images with slightly flawed structure are counted as structure failures, and the share with slightly flawed structure drops from 19.0\% to 16.4\%. The 78.8\% and 1.7\% in the main text correspond only to the values 0.5 and 0.9.

\begin{table}[htbp]
\centering
\renewcommand{\thetable}{G.2}
\caption{Share of metric failures with intact structure (scale-aligned $\delta_1$ $\geq$ 0.9) per dataset, pooling the 18 model evaluations (with and without intrinsics). Counting only the 11 model evaluations with intrinsics, the shares are 75\% to 87\% per dataset and 33\% for DDAD, different from here.}
\label{tab:G2}
\small
\setlength{\tabcolsep}{4pt}
\begin{tabular*}{\linewidth}{@{\extracolsep{\fill}}lrrrrrrr@{}}
\toprule
\makecell{Failure\\threshold} & NYUv2 & KITTI & ETH3D & iBims-1 & DDAD & DIODE & HAMMER \\
\midrule
$<$ 0.3 & 91\% & 95\% & 88\% & 92\% & 55\% & 81\% & 86\% \\
$<$ 0.4 & 87\% & 95\% & 87\% & 91\% & 54\% & 78\% & 86\% \\
$<$ 0.5 & 88\% & 95\% & 87\% & 91\% & 52\% & 77\% & 87\% \\
$<$ 0.6 & 87\% & 93\% & 87\% & 91\% & 48\% & 75\% & 87\% \\
$<$ 0.7 & 84\% & 90\% & 87\% & 79\% & 44\% & 74\% & 87\% \\
\bottomrule
\end{tabular*}
\end{table}

The ordering between datasets is roughly the same at every threshold: DDAD is always the lowest and DIODE always the second lowest. DDAD is the only dataset where pure scale failures are not a clear majority (44\% to 55\%), consistent with the conclusion of Section~\hyperref[sec:4.3]{4.3} that DDAD has the largest structural error.

\begin{table}[htbp]
\centering
\renewcommand{\thetable}{G.3}
\caption{Statistics of Section~\hyperref[sec:5]{5} (11 architectures with intrinsics) as the metric failure threshold varies, with the definitions of shared failure ($k \geq 6$) and specific failure ($k \leq 2$) unchanged. ``4 models'' are Depth Pro, Metric3D-v2, DepthVLM and MA-DepthMap; ``top 6'' are the 6 models with the highest metric $\delta_1$ in Table~\hyperref[tab:7]{7}. ``Scenes holding 90\% of shared failures'' groups the shared-failure images by recording scene, sorts the scenes by their number of such images, and counts how many scenes are needed to reach 90\% of all shared-failure images; it measures how concentrated shared failures are at the scene level.}
\label{tab:G3}
\small
\setlength{\tabcolsep}{4pt}
\begin{tabular*}{\linewidth}{@{\extracolsep{\fill}}lrrrrrrr@{}}
\toprule
\makecell{Failure\\threshold} & Failures & \makecell{Share\\shared} & \makecell{Share\\specific} & \makecell{Specific from\\the 4 models} & \makecell{Scenes with\\90\% of shared} & \makecell{HAMMER shared\\overest.} & \makecell{Top-6 $\delta_1$ error\\on shared} \\
\midrule
$<$ 0.3 & 6,570 & 40\% & 29\% & 90\% & 7 & 100.0\% & 32\% to 40\% \\
$<$ 0.4 & 7,082 & 41\% & 27\% & 90\% & 8 & 99.8\% & 34\% to 43\% \\
$<$ 0.5 & 7,750 & 42\% & 27\% & 90\% & 9 & 99.5\% & 38\% to 46\% \\
$<$ 0.6 & 8,661 & 43\% & 26\% & 88\% & 9 & 98.7\% & 41\% to 49\% \\
$<$ 0.7 & 9,961 & 43\% & 23\% & 85\% & 9 & 97.4\% & 45\% to 52\% \\
\bottomrule
\end{tabular*}
\end{table}

As the failure threshold varies from 0.3 to 0.7, shared failures always make up about 40\% of failures and concentrate in 7 to 9 scenes, and HAMMER failures almost all overestimate depth throughout; specific failures always come more than 85\% from these 4 models; the share of shared-failure images in the error of the top 6 rises as the threshold is relaxed (32\% to 52\%) but always remains a large part. Changing the definition of $k$ (with the failure threshold kept at 0.5) leads to the same conclusions: with shared failures defined as $k \geq 5$ or $k \geq 7$, they make up 52\% and 35\% of all failures, 90\% of them come from 9 and 7 scenes, the share of shared failures in the $\delta_1$ error of the top 6 is 46\% to 53\% and 32\% to 41\%, removing these images changes ranks under $\delta_1$ by at most 1, and under Rel MA-DepthMap drops from 1st to 7th in both cases (Table~\hyperref[tab:G4]{G.4}); with specific failures defined as $k \leq 1$, they make up 14\% of all failures, 95\% of them from these 4 models.

\begin{table}[htbp]
\centering
\renewcommand{\thetable}{G.4}
\caption{Ranks after removing the shared-failure images (11 architectures with intrinsics, 7 benchmarks weighted equally) under three thresholds on $k$. Each cell gives the rank by metric $\delta_1$ / the rank by metric Rel; ``All'' is the original rank of Table~\hyperref[tab:7]{7}.}
\label{tab:G4}
\small
\setlength{\tabcolsep}{4pt}
\begin{tabular}{lrrrr}
\toprule
Model & All & \makecell{Without $k \geq 5$} & \makecell{Without $k \geq 6$} & \makecell{Without $k \geq 7$} \\
\midrule
Metric3D-v2 & 9 / 8 & 9 / 9 & 9 / 9 & 9 / 9 \\
DA3METRIC & 5 / 2 & 5 / 5 & 5 / 4 & 5 / 4 \\
Depth Pro & 11 / 10 & 11 / 10 & 11 / 10 & 11 / 10 \\
MA-DepthMap & 7 / 1 & 8 / 7 & 8 / 7 & 8 / 7 \\
DepthVLM & 10 / 11 & 10 / 11 & 10 / 11 & 10 / 11 \\
MoGe-2 & 1 / 5 & 1 / 1 & 1 / 1 & 1 / 1 \\
MoGe-3 & 2 / 3 & 3 / 2 & 3 / 2 & 3 / 2 \\
MA-PointMap & 3 / 6 & 2 / 3 & 2 / 3 & 2 / 3 \\
FoundationGeo & 6 / 7 & 6 / 6 & 6 / 6 & 6 / 6 \\
UniDepthV2 & 8 / 9 & 7 / 8 & 7 / 8 & 7 / 8 \\
UniK3D & 4 / 4 & 4 / 4 & 4 / 5 & 4 / 5 \\
\bottomrule
\end{tabular}
\end{table}

\textbf{Binned values of the step relation.} Binning all ``image $\times$ model evaluation'' pairs of the 18 model evaluations by $s^{*}$, with bin edges at 0.8, 0.9, 1/0.9, 1.25 and their squares (symmetric about 1 on a log scale): for $s^{*}$ between 0.9 and 1.11 the mean metric $\delta_1$ is 0.95; between 0.8 and 0.9 and between 1.11 and 1.25 it is 0.86 and 0.85; just outside [0.8, 1.25], i.e. between 0.64 and 0.8 and between 1.25 and 1.56, it drops to 0.12 and 0.17, and further out to 0.01 and 0.02. For the three models of Figure~\hyperref[fig:2]{2}, 98\% to 99\% of failed images have $s^{*}$ outside [0.8, 1.25], and images outside this range have a mean metric $\delta_1$ of only 0.08 to 0.10. In the top row of Figure~\hyperref[fig:2]{2}, the metric Rel follows the reference curve $\lvert 1/s^{*} - 1\rvert$ monotonically: for each model evaluation, the Spearman correlation between per-image metric Rel and $\lvert 1/s^{*} - 1\rvert$ over its 4,406 images has a median of 0.89 over the 18 model evaluations (range 0.71 to 0.97). We use Spearman rather than Pearson correlation. Spearman compares only the ranks of the two variables, whereas Pearson uses the values; since Rel is unbounded, a few images with very large Rel dominate the Pearson correlation: DepthVLM has the largest structural error but the highest Pearson correlation (0.99; 398 of its images have Rel above 1).

\textbf{Points off the dashed line in Figure~\hyperref[fig:2]{2}.} For the three models of Figure~\hyperref[fig:2]{2}, the images whose metric Rel exceeds the reference curve by more than 0.15 number 123, 106 and 113 for DA3METRIC, MoGe-2 and UniK3D (2.8\%, 2.4\% and 2.6\% of the 4,406 images); all of them also have a scale-aligned Rel above 0.15, and most come from DDAD (103, 70 and 86). Of the points off the step in the bottom row, those with $s^{*}$ near the 0.8 or 1.25 boundary mostly have intact structure and deviate only because the scale lies right at the boundary; those away from the boundary almost all have flawed structure (see the next paragraph). On DDAD the structural error is not small even when the scale is right: restricted to images with $s^{*}$ within [0.8, 1.25], the mean scale-aligned Rel of DDAD is 0.107 to 0.116 against 0.028 to 0.078 on the other 6 datasets, and the corresponding mean metric $\delta_1$ is 0.864 to 0.911, below the 0.925 to 0.980 elsewhere.

\textbf{Points off the step in Figure~\hyperref[fig:2]{2}.} Section~\hyperref[sec:4.3]{4.3} refers to the points in the bottom row of Figure~\hyperref[fig:2]{2} that lie off the step: $s^{*}$ within [0.8, 1.25] but metric $\delta_1$ below 0.6, or $s^{*}$ outside the range but metric $\delta_1$ above 0.3. The three models of Figure~\hyperref[fig:2]{2} (DA3METRIC, MoGe-2, UniK3D) have 127, 125 and 89 such points. We define ``near the boundary'' as $\lvert\ln(s^{*}/0.8)\rvert$ or $\lvert\ln(s^{*}/1.25)\rvert$ below 0.1, i.e. $s^{*}$ between 0.72 and 0.88 or between 1.13 and 1.38. Near the boundary there are 109, 85 and 80 points, most with intact structure (91, 55 and 69): when the scale lies right at the boundary, any slight non-uniformity in depth puts some pixels inside the factor of 1.25 and others outside, so $\delta_1$ falls between 0 and 1. Away from the boundary there are 18, 40 and 9 points, almost all with flawed structure (17, 38 and 9). The other 3 have intact structure; their $s^{*}$ lies 0.10 to 0.18 from the boundary in log, just outside the boundary window defined above, so they are the same boundary effect.

\textbf{Shared failures counted by family.} The $k$ of Section~\hyperref[sec:5.1]{5.1} treats the 11 architectures as independent judgements, but some share code or authors. We merge them into 6 families: the MoGe family (MoGe-2, MoGe-3, MA-PointMap, FoundationGeo, which share MoGe's post-hoc solve; MA-PointMap is fine-tuned from MoGe-2), the Uni family (UniDepthV2, UniK3D), the Depth Pro family (Depth Pro, MA-DepthMap; the latter reuses the network and focal-length formula of the former, \citealp{ma2026metricanything}), and one family each for Metric3D-v2, DA3METRIC and DepthVLM. A family fails an image when more than half of its members fail it, and an image is a shared failure when at least 4 families fail it. This gives 320 shared-failure images (instead of 387): 221 on HAMMER, 206 of them among the original 218; 95 on DIODE, all among the original 157; and 4 on the other 5 datasets. With datasets weighted equally, they make up 5.9\% of the images (instead of 7.3\%) and account for 32\% to 40\% of the error of the top 6 models by metric $\delta_1$ (instead of 38\% to 46\%). Splitting Depth Pro and MA-DepthMap into two families (7 in total) and still requiring at least 4 failing families gives 461 shared-failure images (247 on HAMMER, 204 on DIODE), 8.7\% of the images and 38\% to 48\% of the error of the top 6 models. The HAMMER shared failures are thus stable under every count; the number on DIODE depends on how architectures are counted and drops by about four in ten with 6 families. Rebuilding the shared failures from the 7 architectures evaluated without intrinsics (at least 4 failing) gives 198 HAMMER images, 179 of them among the 218 with K, and 231 DIODE images, of which only 94 are among the 157 with K. The HAMMER shared failures are thus largely the same images under both intrinsics conditions, whereas those on DIODE change markedly with the condition. Under the three counts, DIODE has 157 shared failures (11 architectures, k $\geq$ 6), 108 (without Metric3D-v2) and 95 (6 families), and HAMMER 218, 190 and 206; the main results throughout use the first of these counts.

\textbf{Held-out selection.} The shared-failure images are selected by the 11 architectures' own failures, so their failure rates on these images are biased by the selection. We select images with only 5 architectures from different families (MoGe-2, UniDepthV2, Metric3D-v2, Depth Pro, DepthVLM), requiring at least 4 to fail, and look at the 6 architectures that took no part. HAMMER yields 212 images: DA3METRIC fails on 91\% of the 212 selected images and on 9\% of the other 563; MoGe-3 on 84\% and 4\%; MA-PointMap on 85\% and 5\%; FoundationGeo on 79\% and 12\%; UniK3D on 92\% and 24\% (each pair gives the failure rates on two groups of images, not two parts of one whole). Only MA-DepthMap is reversed (6\% and 34\%): it is the only architecture that underestimates depth on HAMMER (bias $-$0.14), so it happens to be right on the images where the others overestimate (Section~\hyperref[sec:5.2]{5.2}). DIODE yields 165 images: DA3METRIC 57\% and 11\%, MoGe-3 42\% and 8\%, MA-PointMap 52\% and 9\%, FoundationGeo 53\% and 14\%, UniK3D 40\% and 12\%, MA-DepthMap 29\% and 22\%. Of the 6 architectures that took no part, only DA3METRIC belongs to a family outside the selectors; it fails on 91\% of the selected HAMMER images against 9\% of the others, and 57\% against 11\% on DIODE. The family-held-out version, in which each family is excluded from selecting its own test images, is in Table~\hyperref[tab:I4]{I.4}. Hard images selected by one set of models are thus much harder than the remaining images for the models that took no part. The same held-out test applies to the error share: for each of the top 6 models, we consider only the architectures outside its family (7 for the MoGe family, 9 or 10 for the others), rank the images of each dataset by how many of them fail, breaking ties by their mean metric $\delta_1$ (lowest first), and take as many images as there are original shared failures. Models from different families thus use different sets of images (the same set for the 4 models of the MoGe family), which overlap the original shared failures by 81\% to 93\%. On each dataset this amounts to ``at least 3 to 7 of the other-family architectures fail'', a threshold that varies with dataset and family; we fix the number rather than the threshold so that the error share can be compared directly with the original shared failures and with random images. An alternative does not fix the number but requires the share of failing other-family architectures to be at least 6/11 (the same share as $k \geq 6$ among 11, e.g. at least 4 of 7 other-family architectures); the images so selected account for 36\% to 45\% of the error, but the number selected per dataset differs from the original shared failures, so we report the fixed-number result. The selected images account for 34\% to 40\% of the model's $1-\delta_1$ error (MoGe-2 38\%, MoGe-3 35\%, MA-PointMap 38\%, FoundationGeo 34\%, UniK3D 39\%, DA3METRIC 40\%), against 38\% to 46\% when all 11 architectures select; the held-out selections overlap the original shared failures by 81\% to 93\%, and the same number of random images accounts for only 14\% to 18\%. Another control, which is not held out: taking, for each model and dataset, the same number of images with the model's own largest error, these images account for 47\% to 54\% of its error, and 70\% to 82\% of them are shared-failure images; that is, each model's worst images are largely the same images on which most models fail.

\textbf{Random baseline.} Section~\hyperref[sec:5.1]{5.1} tests whether, within a scene, the architectures tend to fail on the same images. For each architecture, we randomly permute its failure flags (whether it fails each image) within each recorded scene, keeping its failure rate in every scene, and repeat 200 times. The images of DIODE, HAMMER and ETH3D come from a limited number of captures, and one capture (e.g. scene\_00022 of DIODE or scene10 of HAMMER) is called a recorded scene (Section~\hyperref[sec:3.2]{3.2}). Table~\hyperref[tab:G5]{G.5} compares three statistics: the number of images with $k \geq 6$, the within-scene variance of $k$, and the pairwise correlation ($\varphi$ coefficient) between the failure flags of the architectures within a scene.

\begin{table}[htbp]
\centering
\renewcommand{\thetable}{G.5}
\caption{Within-scene permutation control (11 architectures with intrinsics, 200 permutations). Each cell gives observed / permuted, where ``permuted'' is the mean after permutation; the variance of $k$ is within scenes, $\varphi$ is the pairwise correlation between the failure flags of the architectures within a scene, and the brackets give the 95\% interval of the permuted $\varphi$.}
\label{tab:G5}
\small
\setlength{\tabcolsep}{4pt}
\begin{tabular}{lrrl}
\toprule
Stratum & $k \geq 6$ & \makecell{Variance of $k$} & \makecell{Failure correlation $\varphi$ [95\%]} \\
\midrule
HAMMER, 7 scenes & 218 / 308 & 4.42 / 1.21 & 0.228 / 0.002 [$-$0.048, 0.052] \\
HAMMER, scene $\times$ empty/objects (13 strata) & 218 / 246 & 2.38 / 0.82 & 0.197 / $-$0.001 [$-$0.034, 0.034] \\
DIODE, 6 scenes & 157 / 83 & 5.08 / 1.61 & 0.240 / $-$0.001 [$-$0.011, 0.010] \\
\bottomrule
\end{tabular}
\end{table}

On both datasets the variance and $\varphi$ are far above their permuted values, so within a scene the architectures tend to fail on the same images. On HAMMER the observed number of images with $k \geq 6$ is below the permuted one because the hard scenes have high failure rates (0.58 to 0.66 in scene10, scene3 and scene5): the count of $k \geq 6$ does not change monotonically with the correlation. For example, if all 11 architectures have a failure rate of 0.7, an image has $k \geq 6$ with probability about 0.92 when they are independent but only 0.70 when they are perfectly synchronised; with failure rates of 0.3, the probabilities are about 0.08 and 0.30. The main text therefore uses $\varphi$ rather than the number of images with $k \geq 6$. As a supplement, permuting only within each dataset, which keeps each architecture's failure rate per dataset, gives 220 images with $k \geq 6$ on average (95\% interval 203 to 240) against 387 observed.

\textbf{Without DDAD.} The training data of DepthVLM, Metric3D-v2, DA3METRIC and the two MetricAnything models include DDAD (Section~\hyperref[sec:3.1]{3.1}). On the other 6 benchmarks, 98.3\% of failures still no longer fail after scale correction, and the best of the 18 evaluations (MoGe-3 with K) fails on 8.7\%.

\textbf{The four kinds of failure.} The categories used in Sections~\hyperref[sec:4]{4} and \hyperref[sec:5]{5} are summarised in Table~\hyperref[tab:2]{2} (Section~\hyperref[sec:3.3]{3.3}).

\section{Details and Checks of the Diagnostic Data}\label{app:H}

\begin{figure}[htbp]
\centering
\renewcommand{\thefigure}{H.1}

\includegraphics[width=1.00\linewidth]{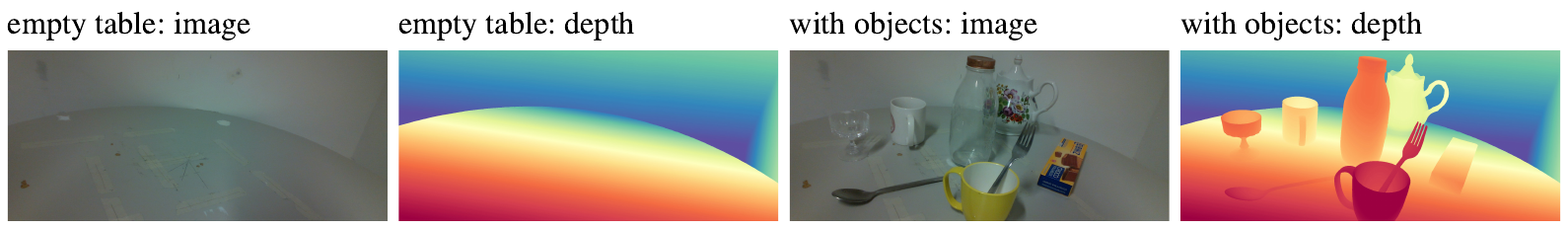}
\caption{An example HAMMER pair (scene5\_traj2, frame 160): adding objects reduces the scale error $\lvert\ln s^{*}\rvert$ of MoGe-2 (without intrinsics) from 1.12 (a severe failure) to 0.08 (nearly correct). The raised object regions are clearly visible in the depth maps, while the background (table edge, wall) has the same shape and gradient in both frames.}
\label{fig:H1}
\end{figure}

\textbf{Background check of the HAMMER pairs.} The pairs of Section~\hyperref[sec:6.1]{6.1} assume that the two passes differ only in the objects on the table. We verified this on the data: taking as background the pixels whose depths differ by less than 1 cm between the two frames and that are more than 15 pixels from any object (68\% to 96\% of all pixels, 83\% on average), we compare their RGB values in the two frames. Over the 371 pairs, the median RGB difference of background pixels is 4.3 (on a 0–255 scale; 90th percentile 7.0), and no pair exceeds 20; it is 33.3 in object regions and 8.0 in the shadow band within 15 pixels of objects. The backgrounds of the two passes agree on all pairs, and the differences appear only on the objects and in a narrow band around them. An example is shown in Figure~\hyperref[fig:H1]{H.1}.

\textbf{Solving the scale on background pixels only.} To check whether the change in Section~\hyperref[sec:6.3]{6.3} comes only from the object pixels themselves, we restrict the solve for $s^{*}$ to the background pixels shared by both frames. The background is defined as in the previous paragraph (ground-truth depths differing by less than 1 cm between the frames, and more than 15 pixels from object regions); the mask depends only on the ground truth, is shared by both frames and is built on the pixel grid of the evaluation pipeline; the solver is the same as in the main text. With the margin, the background covers 73.0\% of the pixels on average. Table~\hyperref[tab:H1]{H.1} compares the paired change in $\lvert\ln s^{*}\rvert$ (with objects minus empty table, 11 architectures with K, 371 pairs; 95\% intervals from resampling the 6 scenes 2,000 times) under the full-image and the background $s^{*}$. Excluding MA-DepthMap, the median ratio of the background change to the full-image change is 1.01; the change in bias on the background is negative for all 11 models.

\textbf{Per-model tests.} Section~\hyperref[sec:6.3]{6.3} reports only the scene-level sign-flip test of the change averaged over 11 models (negative in 5 of the 6 scenes, one-sided p = 0.031, two-sided p = 0.06). The test is one-sided because the direction was fixed in advance: it tests whether adding objects reduces the scale error; with 6 scenes the smallest possible one-sided p is 1/64 = 0.016. Per model, Metric3D-v2 and DA3METRIC are negative in all 6 scenes (p = 0.016), Depth Pro, MoGe-3 and FoundationGeo have p = 0.047, the others lie between 0.06 and 0.11, and MA-DepthMap goes the other way (0.953). The scene-resampled bootstrap intervals (Table~\hyperref[tab:H1]{H.1}) exclude zero on the full image for all 10 models whose scale error shrinks, but with only 6 scenes these intervals are optimistic, so the main text relies on the sign-flip test.

\begin{table}[htbp]
\centering
\renewcommand{\thetable}{H.1}
\caption{Change in $\lvert\ln s^{*}\rvert$ when objects are added to the HAMMER pairs, solving the scale on the full image or on background pixels only. Intervals resample the 6 scenes 2,000 times and are themselves unstable with only 6 scenes. The last column is the one-sided p-value of a scene-level sign-flip test on the full-image change: under the hypothesis that adding objects does not change the scale error, the mean change of each of the 6 scenes is equally likely to be positive or negative, giving $2^6$ = 64 sign combinations; the p-value is the share of these 64 combinations whose mean change is not larger than the observed one, so the smallest possible p is 1/64 = 0.016.}
\label{tab:H1}
\small
\setlength{\tabcolsep}{4pt}
\begin{tabular}{lllr}
\toprule
\makecell{Model\\(with K)} & \makecell{Full image [95\% interval]} & \makecell{Background [95\% interval]} & Sign-flip p \\
\midrule
Metric3D-v2 & $-$0.188 [$-$0.242, $-$0.120] & $-$0.181 [$-$0.237, $-$0.110] & 0.016 \\
DA3METRIC & $-$0.125 [$-$0.271, $-$0.026] & $-$0.127 [$-$0.277, $-$0.027] & 0.016 \\
Depth Pro & $-$0.186 [$-$0.326, $-$0.051] & $-$0.190 [$-$0.328, $-$0.058] & 0.047 \\
MA-DepthMap & +0.043 [+0.007, +0.081] & +0.050 [+0.010, +0.087] & 0.953 \\
DepthVLM & $-$0.146 [$-$0.340, $-$0.007] & $-$0.139 [$-$0.328, $-$0.010] & 0.109 \\
MoGe-2 & $-$0.209 [$-$0.494, $-$0.021] & $-$0.210 [$-$0.491, $-$0.027] & 0.094 \\
MoGe-3 & $-$0.197 [$-$0.408, $-$0.039] & $-$0.199 [$-$0.408, $-$0.044] & 0.047 \\
MA-PointMap & $-$0.207 [$-$0.448, $-$0.031] & $-$0.202 [$-$0.434, $-$0.032] & 0.062 \\
FoundationGeo & $-$0.184 [$-$0.331, $-$0.052] & $-$0.185 [$-$0.332, $-$0.052] & 0.047 \\
UniDepthV2 & $-$0.132 [$-$0.272, $-$0.014] & $-$0.136 [$-$0.281, $-$0.013] & 0.109 \\
UniK3D & $-$0.173 [$-$0.318, $-$0.045] & $-$0.175 [$-$0.323, $-$0.045] & 0.062 \\
\bottomrule
\end{tabular}
\end{table}

\begin{figure}[htbp]
\centering
\renewcommand{\thefigure}{H.2}

\includegraphics[width=1.00\linewidth]{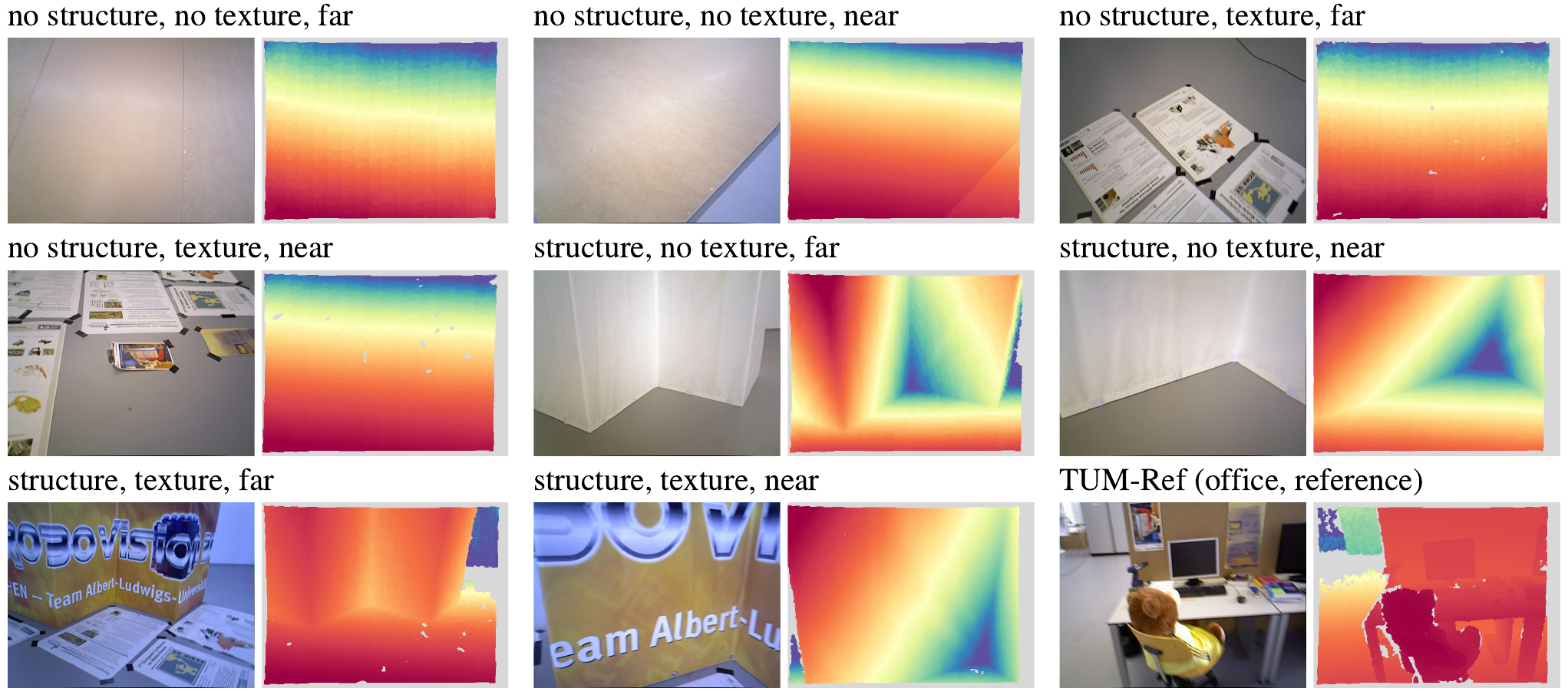}
\caption{One representative frame (the middle frame) of each of the 8 TUM-Family sequences, plus TUM-Ref, 9 images in total; each cell shows the input image on the left and the ground-truth depth on the right (logarithmic colour scale). It only illustrates the three variables (structure, texture, distance) and enters no statistic.}
\label{fig:H2}
\end{figure}

\textbf{Frame sampling and ground-truth check of TUM-Family.} RGB and depth frames are paired when their timestamps differ by less than 20 ms; a frame is kept when the camera has moved at least 0.10 m or rotated at least 10$^\circ$ since the last kept frame (the first frame is always kept), and when it has at least 500 valid depth pixels, giving 514 images (30 to 131 per sequence). These rules depend only on poses, sensor metadata and depth coverage, and were fixed before looking at any model results. The motion thresholds are loose because the far sequences are short: with 0.15 m or 15$^\circ$, the number of frames drops to about 345. Valid depths range from 0.74 to 3.86 m (1st and 99th percentiles). The two camera distances differ only moderately: by the within-sequence median of the per-frame median ground-truth depth, the near sequences are at 0.93 to 1.24 m and the far ones at 1.76 to 2.41 m, a factor of about 1.6 to 2.1. We use the official intrinsics of the Freiburg 3 colour camera (fx = 535.4, fy = 539.2, cx = 320.1, cy = 247.6 pixels) and the official depth scale factor of 5000. Attributing the scale error of a model to image content requires metrically accurate ground-truth depth. We confirmed this with a model-independent test: for the 4 sequences with structure, we reproject each frame into the next using the ground-truth poses and compare the reprojected depth with the measured depth of the next frame. The error is smallest at a depth scale of 1.00 and clearly higher at 0.5 or 2, so depth maps and poses are metrically consistent. The median log error is then 0.004 to 0.010, while the scale errors of the models (0.3 to 0.9) are at least 30 times larger, so the ground truth is accurate enough to resolve differences in scale error between models and between sequences. In the 4 flat sequences the image shows one flat surface without depth variation to compare against, so the test is not sensitive enough there and is done on the sequences with structure; the flat sequences share the sensor and depth scale factor. Figure~\hyperref[fig:H2]{H.2} shows a representative frame of each sequence, together with one frame of TUM-Ref.

\textbf{Changes between TUM-Family sequences and factor-by-factor comparisons.} Not only the direction on each sequence agrees, but also the change in bias between sequences. The 8 sequences form 28 pairs; moving from one sequence to the other, the biases of all 11 models move in the same direction on 19 pairs, and those of 10 models on another 2. Whether the scale error grows or shrinks depends on which side the model was on. For example, moving from far to near shifts the biases of the models towards overestimation: on S-NT all 11 models do so (median +0.28), and since they mostly overestimated depth on S-NT-far already, the scale error (by absolute bias) grows for 10 of the 11; on NS-T, 10 models do so (median +0.25), and since they underestimated depth on NS-T-far, the scale error shrinks for the same 10. The sequence thus changes a bias shared by all models, and the size of the error is determined by the combination of this shared bias with each model's own bias. The shared failures of Section~\hyperref[sec:5]{5} are the same phenomenon: on the same scenes, the scale of most models errs in the same direction. Within the 8 sequences, fixing two factors and changing only the third isolates each factor. The effect of distance is described above. Structure: in the two groups without texture, switching to the zigzag board shifts the biases of all 11 models towards overestimation (median +0.43 at far and +0.31 at near distance); in the two groups with texture, the direction is inconsistent (6 and 7 of the 11 shift towards overestimation). The effect of texture is discussed separately in Section~\hyperref[sec:6.4]{6.4}. Both ends of these comparisons are separate recordings with different viewpoints and trajectories. The HAMMER pairs fix the viewpoint (Section~\hyperref[sec:6.3]{6.3}).

\section{Model Size}\label{app:I}

This appendix tests separately what increasing model size does for scale recovery and for structure recovery. Families with several public sizes are evaluated with the protocol of Section~\hyperref[sec:3]{3}: Metric3D-v2 (ViT-S, ViT-L, ViT-g), MoGe-2 (ViT-S, ViT-B, ViT-L), MoGe-3 (ViT-L, ViT-G), UniDepthV2 (ViT-S, ViT-B, ViT-L) and UniK3D (ViT-S, ViT-B, ViT-L). Metric3D-v2 is evaluated only with intrinsics, and the other four families both without and with intrinsics, for 25 model evaluations in total. Metric $\delta_1$ measures overall accuracy and scale-aligned $\delta_1$ measures structure (approximately the accuracy reached with a perfectly correct scale); their difference is the part of scale estimation not yet realised, i.e. roughly how much metric $\delta_1$ could still gain if scale estimation were perfect. Different sizes of one family may use different training data and recipes, so sizes are compared only within a family, not across families. For MoGe-2, all three sizes use the official \texttt{-normal} weights; the ViT-L here is therefore \texttt{moge-2-vitl-normal}, a different set of weights from the \texttt{moge-2-vitl} of the main table, with the same 7-benchmark metric $\delta_1$ without K (0.769) and scale-aligned $\delta_1$ of 0.972 and 0.971.

\begin{figure}[htbp]
\centering
\renewcommand{\thefigure}{I.1}

\includegraphics[width=0.95\linewidth]{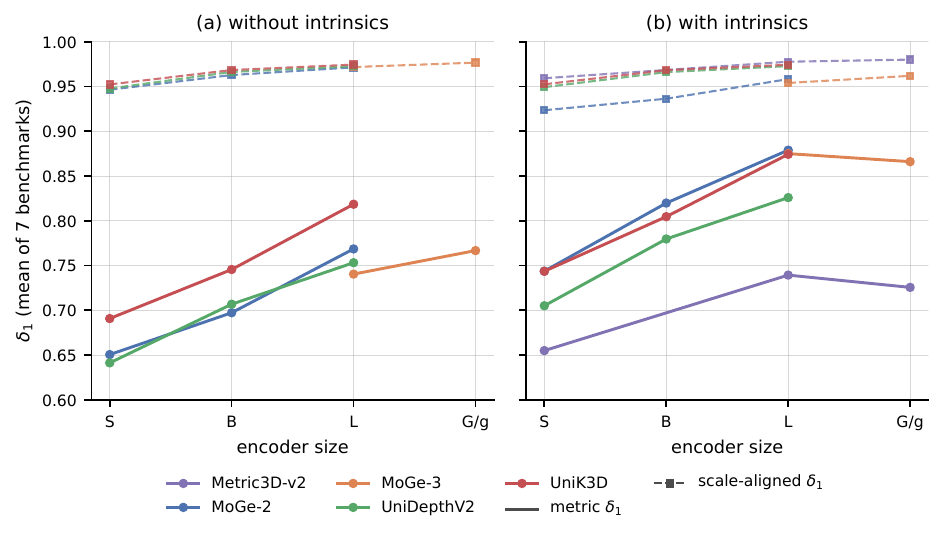}
\caption{Metric $\delta_1$ (solid) and scale-aligned $\delta_1$ (dashed) of each family as size increases, averaged over the 7 benchmarks; left: without intrinsics, right: with intrinsics.}
\label{fig:I1}
\end{figure}

\begin{table}[htbp]
\centering
\renewcommand{\thetable}{I.1}
\caption{Results of different sizes of each family on the 7 benchmarks, with datasets weighted equally (Section~\hyperref[sec:3.3]{3.3}). ``Gap'' is the scale-aligned $\delta_1$ minus the metric $\delta_1$, i.e. roughly how much metric $\delta_1$ could gain if scale estimation were perfect: the larger the gap, the larger the part of the metric error that comes from scale; the smaller the gap, the more of the remaining metric error comes from structure. (a) Without intrinsics, 4 families; (b) with intrinsics, 5 families (Metric3D-v2 is evaluated only with intrinsics). All three MoGe-2 sizes use the \texttt{-normal} checkpoints, so ViT-L with K has a metric $\delta_1$ of 0.879 and a failure rate of 8.0\%, slightly different from the 0.877 and 8.1\% of the main-table checkpoint in Table~\hyperref[tab:3]{3}.}
\label{tab:I1}
\small
\begin{minipage}[b]{0.49\linewidth}
\centering
\textit{(a) Without intrinsics}\\[2pt]
\setlength{\tabcolsep}{3pt}
\begin{tabular}[b]{llrrrr}
\toprule
Family & Size & Metric $\delta_1$ & SA $\delta_1$ & Gap & Fail. \\
\midrule
MoGe-2 & ViT-S & 0.651 & 0.946 & 0.296 & 31.5\% \\
MoGe-2 & ViT-B & 0.697 & 0.963 & 0.266 & 27.7\% \\
MoGe-2 & ViT-L & 0.769 & 0.971 & 0.203 & 20.7\% \\
\addlinespace[2pt]
MoGe-3 & ViT-L & 0.740 & 0.972 & 0.231 & 23.8\% \\
MoGe-3 & ViT-G & 0.767 & 0.977 & 0.210 & 21.5\% \\
\addlinespace[2pt]
UniDepthV2 & ViT-S & 0.641 & 0.948 & 0.307 & 33.1\% \\
UniDepthV2 & ViT-B & 0.707 & 0.966 & 0.259 & 26.0\% \\
UniDepthV2 & ViT-L & 0.753 & 0.973 & 0.219 & 21.6\% \\
\addlinespace[2pt]
UniK3D & ViT-S & 0.691 & 0.952 & 0.262 & 26.7\% \\
UniK3D & ViT-B & 0.746 & 0.968 & 0.223 & 22.0\% \\
UniK3D & ViT-L & 0.818 & 0.974 & 0.156 & 15.0\% \\
\bottomrule
\end{tabular}
\end{minipage}\hfill
\begin{minipage}[b]{0.49\linewidth}
\centering
\textit{(b) With intrinsics}\\[2pt]
\setlength{\tabcolsep}{3pt}
\begin{tabular}[b]{llrrrr}
\toprule
Family & Size & Metric $\delta_1$ & SA $\delta_1$ & Gap & Fail. \\
\midrule
Metric3D-v2 & ViT-S & 0.655 & 0.959 & 0.304 & 31.5\% \\
Metric3D-v2 & ViT-L & 0.739 & 0.978 & 0.238 & 23.1\% \\
Metric3D-v2 & ViT-g & 0.726 & 0.980 & 0.254 & 24.5\% \\
\addlinespace[2pt]
MoGe-2 & ViT-S & 0.744 & 0.923 & 0.180 & 18.9\% \\
MoGe-2 & ViT-B & 0.820 & 0.936 & 0.117 & 11.5\% \\
MoGe-2 & ViT-L & 0.879 & 0.958 & 0.079 & 8.0\% \\
\addlinespace[2pt]
MoGe-3 & ViT-L & 0.875 & 0.954 & 0.079 & 7.7\% \\
MoGe-3 & ViT-G & 0.866 & 0.962 & 0.096 & 9.3\% \\
\addlinespace[2pt]
UniDepthV2 & ViT-S & 0.705 & 0.949 & 0.244 & 25.7\% \\
UniDepthV2 & ViT-B & 0.780 & 0.966 & 0.186 & 18.2\% \\
UniDepthV2 & ViT-L & 0.826 & 0.973 & 0.147 & 14.4\% \\
\addlinespace[2pt]
UniK3D & ViT-S & 0.744 & 0.952 & 0.209 & 21.7\% \\
UniK3D & ViT-B & 0.805 & 0.968 & 0.164 & 15.9\% \\
UniK3D & ViT-L & 0.874 & 0.974 & 0.100 & 9.8\% \\
\bottomrule
\end{tabular}
\end{minipage}
\end{table}
\textbf{In $\delta_1$ points, the gain from size is mainly in scale.} A ``step'' is a comparison between two adjacent sizes of the same family under the same intrinsics condition (for example, MoGe-2 without intrinsics from ViT-S to ViT-B); there are 16 steps (Figure~\hyperref[fig:I1]{I.1}). The point estimate of each step is the difference between the two corresponding rows of Table~\hyperref[tab:I1]{I.1} (7-benchmark averages).

To judge whether this difference is reliable or noise that might vanish with another set of images of the same kind, we estimate 95\% intervals with a paired bootstrap: for each dataset, we draw with replacement as many images as the dataset contains, use the same drawn images for the larger and the smaller size, and compute the difference on that dataset; the differences of the 7 datasets are averaged within one draw to give a ``combined difference''; repeating this 1,000 times gives 1,000 combined differences, whose 2.5th and 97.5th percentiles form the interval. Using the same drawn images for both sizes makes the interval reflect only how the difference in relative performance between the two sizes changes with another set of images, without the extra noise of each size drawing different images (Section~\hyperref[sec:3.3]{3.3}).

The scale-aligned $\delta_1$ rises at every step, and none of the 95\% intervals contains zero, but the gains are small, at most 0.022 for one step. In 14 of the steps the metric $\delta_1$ rises together with the scale-aligned $\delta_1$, by 2.7 to 12.0 times the scale-aligned gain of the same step (median 4.9 times). Measured in $\delta_1$, the gain in metric accuracy from larger models in these families therefore comes mostly from better scale estimation. For MoGe-2 without intrinsics, from ViT-S to ViT-L, the scale-aligned $\delta_1$ rises from 0.946 to 0.971, the metric $\delta_1$ from 0.651 to 0.769, and the part removable by a global scale shrinks from 0.296 to 0.203.

\textbf{Size does not improve scale monotonically.} In 2 steps the metric $\delta_1$ actually drops: for MoGe-3 with intrinsics from ViT-L to ViT-G by 0.009 (95\% interval of the change [-0.017, -0.001]), and for Metric3D-v2 from ViT-L to ViT-g by 0.014 (interval [-0.021, -0.008]), while the scale-aligned $\delta_1$ of both steps still rises (by +0.008 and +0.002). In these two families the largest size has better structure but not better scale. Even at the largest sizes, the error removable by a global scale remains considerable (0.096 for MoGe-3 ViT-G with intrinsics, 0.254 for Metric3D-v2 ViT-g), so making models larger does not remove the scale failures described in Sections~\hyperref[sec:4]{4} and \hyperref[sec:5]{5}. In $\delta_1$ the effect of size on structure looks small because structure is already above 0.92 at ViT-S; in relative terms, structure errors shrink about as much as scale errors (Table~\hyperref[tab:I5]{I.5}). Each family has only two or three sizes, and we evaluated only the one set of weights released by the authors for each size, so size itself cannot be separated from training settings that change with size.

\textbf{Failure rate of the shared-failure images across sizes and conditions.} Table~\hyperref[tab:I2]{I.2}, cited in Section~\hyperref[sec:7]{7}, gives the failure rate of all 36 model evaluations (the 18 of the main table and the 18 of the size study) on the two sets of shared failures of Section~\hyperref[sec:5]{5}. Both sets are fixed as the images with $k \geq 6$ under the 11 architectures with K at the ViT-L tier (218 on HAMMER, 157 on DIODE); the 11 evaluations marked $^\dagger$ selected the images, so their failure rates are biased by the selection. The other 25 did not select the images, but they are other sizes or the intrinsics-free runs of the same families, so they are not fully independent of the selection either; Table~\hyperref[tab:I4]{I.4} removes this dependence.

\begin{table}[htbp]
\centering
\renewcommand{\thetable}{I.2}
\caption{Failure rate (metric $\delta_1$ $<$ 0.5) on the shared-failure images of Section~\hyperref[sec:5]{5} (218 on HAMMER, 157 on DIODE), all 36 model evaluations. Each cell gives the failure rate on the shared-failure images / on the other images of the same dataset; $^\dagger$ marks the 11 evaluations used to select the images. ``—'' means that the architecture is evaluated only with intrinsics. MoGe-2 ViT-L (normal) is the \texttt{-normal} checkpoint of the size study.}
\label{tab:I2}
\small
\setlength{\tabcolsep}{4pt}
\begin{tabular}{llrrll}
\toprule
Model & Size & \makecell{HAMMER, w/o K} & DIODE, w/o K & HAMMER, w/ K & DIODE, w/ K \\
\midrule
Metric3D-v2 & ViT-S & — & — & 93\% / 77\% & 94\% / 92\% \\
Metric3D-v2 & ViT-L & — & — & 95\% / 36\%$^\dagger$ & 100\% / 100\%$^\dagger$ \\
Metric3D-v2 & ViT-g & — & — & 94\% / 39\% & 98\% / 99\% \\
DA3METRIC & ViT-L & — & — & 96\% / 6\%$^\dagger$ & 68\% / 9\%$^\dagger$ \\
Depth Pro & ViT-L & 84\% / 16\% & 80\% / 58\% & 78\% / 5\%$^\dagger$ & 87\% / 50\%$^\dagger$ \\
MA-DepthMap & ViT-H+ & — & — & 6\% / 35\%$^\dagger$ & 40\% / 20\%$^\dagger$ \\
DepthVLM & Qwen3-VL & — & — & 100\% / 74\%$^\dagger$ & 82\% / 34\%$^\dagger$ \\
MoGe-2 & ViT-S & 84\% / 13\% & 69\% / 42\% & 98\% / 38\% & 65\% / 27\% \\
MoGe-2 & ViT-B & 51\% / 39\% & 62\% / 26\% & 78\% / 1\% & 59\% / 23\% \\
MoGe-2 & ViT-L & 63\% / 22\% & 55\% / 26\% & 82\% / 1\%$^\dagger$ & 75\% / 5\%$^\dagger$ \\
MoGe-2 & ViT-L (normal) & 56\% / 24\% & 52\% / 25\% & 86\% / 3\% & 63\% / 4\% \\
MoGe-3 & ViT-L & 57\% / 32\% & 43\% / 27\% & 83\% / 4\%$^\dagger$ & 58\% / 4\%$^\dagger$ \\
MoGe-3 & ViT-G & 71\% / 20\% & 38\% / 23\% & 92\% / 11\% & 39\% / 14\% \\
MA-PointMap & ViT-L & 81\% / 2\% & 54\% / 17\% & 89\% / 3\%$^\dagger$ & 67\% / 6\%$^\dagger$ \\
FoundationGeo & ViT-L & 58\% / 15\% & 62\% / 31\% & 82\% / 11\%$^\dagger$ & 68\% / 10\%$^\dagger$ \\
UniDepthV2 & ViT-S & 100\% / 62\% & 68\% / 46\% & 100\% / 72\% & 77\% / 37\% \\
UniDepthV2 & ViT-B & 100\% / 46\% & 61\% / 42\% & 100\% / 45\% & 69\% / 27\% \\
UniDepthV2 & ViT-L & 100\% / 36\% & 55\% / 44\% & 100\% / 35\%$^\dagger$ & 62\% / 23\%$^\dagger$ \\
UniK3D & ViT-S & 85\% / 40\% & 63\% / 45\% & 90\% / 54\% & 73\% / 32\% \\
UniK3D & ViT-B & 93\% / 36\% & 52\% / 36\% & 100\% / 42\% & 60\% / 21\% \\
UniK3D & ViT-L & 96\% / 20\% & 36\% / 21\% & 95\% / 22\%$^\dagger$ & 50\% / 9\%$^\dagger$ \\
\bottomrule
\end{tabular}
\end{table}

Outside the 11 selecting evaluations, the failure rate is 51\% to 100\% on HAMMER and 36\% to 98\% on DIODE; a low value here does not mean that a size or condition fixed these images, since part of them was never failed by that family. Among the 11 selecting evaluations, only MA-DepthMap stays below one half on HAMMER (6\%), for the reason given in Section~\hyperref[sec:5.2]{5.2}. Apart from MA-DepthMap on HAMMER, which is reversed, and Metric3D-v2 on DIODE, which fails almost everywhere, every evaluation fails more often on the shared-failure images than on the other images. Read from Table~\hyperref[tab:I4]{I.4}, where no size of the family under test takes part in the selection, larger models lower the DIODE failure rate for MoGe-2 with intrinsics (55\% $\to$ 45\% $\to$ 37\%), UniDepthV2 and UniK3D, and leave it unchanged for MoGe-3 from ViT-L to ViT-G (27\% and 28\% in both sizes).

\textbf{Largest size with intrinsics.} Table~\hyperref[tab:I3]{I.3}, cited in Section~\hyperref[sec:7]{7}, gives the failure rate of each architecture at its largest public size with ground-truth intrinsics.

\begin{table}[htbp]
\centering
\renewcommand{\thetable}{I.3}
\caption{Failure rate (metric $\delta_1$ $<$ 0.5) of each architecture with ground-truth intrinsics at its largest public size, over the 7 benchmarks (datasets weighted equally) and on DIODE and HAMMER. MoGe-2 ViT-L is listed with both of its public checkpoints (the main-table weights and the \texttt{-normal} weights).}
\label{tab:I3}
\small
\setlength{\tabcolsep}{4pt}
\begin{tabular}{llrrr}
\toprule
Model & Size & 7 benchmarks & DIODE & HAMMER \\
\midrule
Metric3D-v2 & ViT-g & 24.5\% & 98.6\% & 54.7\% \\
DA3METRIC & ViT-L & 9.0\% & 20.9\% & 31.1\% \\
Depth Pro & ViT-L & 31.4\% & 57.7\% & 25.4\% \\
MA-DepthMap & ViT-H+ & 10.8\% & 23.7\% & 26.7\% \\
DepthVLM & Qwen3-VL & 20.4\% & 44.0\% & 81.5\% \\
MoGe-2 & ViT-L & 8.1\% & 19.1\% & 23.9\% \\
MoGe-2 & ViT-L (normal) & 8.0\% & 16.3\% & 25.9\% \\
MoGe-3 & ViT-G & 9.3\% & 19.1\% & 33.9\% \\
MA-PointMap & ViT-L & 9.0\% & 18.0\% & 26.8\% \\
FoundationGeo & ViT-L & 9.7\% & 21.9\% & 30.7\% \\
UniDepthV2 & ViT-L & 14.4\% & 31.3\% & 53.4\% \\
UniK3D & ViT-L & 9.8\% & 17.6\% & 42.8\% \\
\bottomrule
\end{tabular}
\end{table}

\textbf{Selection without the family under test.} Table~\hyperref[tab:I4]{I.4}, cited in Section~\hyperref[sec:7]{7}, rebuilds the shared set without the family being tested, so that no size of that family took part in the selection.

\begin{table}[htbp]
\centering
\renewcommand{\thetable}{I.4}
\caption{Failure rate on shared failures selected without the family under test (selected images / other images of the same dataset). For each family, the shared set is rebuilt from the other architectures with intrinsics: for the MoGe family (MoGe-2, MoGe-3), 7 selectors with at least 4 failing (346 images on HAMMER, 269 on DIODE); for UniDepthV2 and UniK3D, 9 selectors with at least 5 (194 and 173); and for Metric3D-v2, 10 selectors with at least 6 failing (190 images on HAMMER and 108 on DIODE).}
\label{tab:I4}
\small
\setlength{\tabcolsep}{4pt}
\begin{tabular}{lllrr}
\toprule
Family & K & Size & HAMMER & DIODE \\
\midrule
Metric3D-v2 & K & ViT-S & 92\% / 78\% & 95\% / 92\% \\
Metric3D-v2 & K & ViT-L & 95\% / 39\% & 100\% / 100\% \\
Metric3D-v2 & K & ViT-g & 94\% / 42\% & 98\% / 99\% \\
MoGe-2 & w/o K & ViT-S & 58\% / 13\% & 55\% / 44\% \\
MoGe-2 & w/o K & ViT-B & 43\% / 42\% & 43\% / 28\% \\
MoGe-2 & w/o K & ViT-L & 43\% / 25\% & 38\% / 29\% \\
MoGe-2 & K & ViT-S & 86\% / 30\% & 55\% / 24\% \\
MoGe-2 & K & ViT-B & 51\% / 1\% & 45\% / 23\% \\
MoGe-2 & K & ViT-L & 51\% / 2\% & 37\% / 9\% \\
MoGe-3 & w/o K & ViT-L & 51\% / 29\% & 27\% / 32\% \\
MoGe-3 & w/o K & ViT-G & 48\% / 24\% & 27\% / 26\% \\
MoGe-3 & K & ViT-L & 53\% / 5\% & 28\% / 8\% \\
MoGe-3 & K & ViT-G & 74\% / 2\% & 28\% / 14\% \\
UniDepthV2 & w/o K & ViT-S & 100\% / 63\% & 61\% / 47\% \\
UniDepthV2 & w/o K & ViT-B & 100\% / 48\% & 51\% / 44\% \\
UniDepthV2 & w/o K & ViT-L & 96\% / 40\% & 47\% / 46\% \\
UniDepthV2 & K & ViT-S & 100\% / 73\% & 70\% / 38\% \\
UniDepthV2 & K & ViT-B & 100\% / 47\% & 61\% / 28\% \\
UniDepthV2 & K & ViT-L & 96\% / 39\% & 50\% / 26\% \\
UniK3D & w/o K & ViT-S & 86\% / 42\% & 57\% / 46\% \\
UniK3D & w/o K & ViT-B & 94\% / 38\% & 46\% / 38\% \\
UniK3D & w/o K & ViT-L & 92\% / 24\% & 28\% / 23\% \\
UniK3D & K & ViT-S & 92\% / 55\% & 67\% / 33\% \\
UniK3D & K & ViT-B & 100\% / 44\% & 54\% / 22\% \\
UniK3D & K & ViT-L & 91\% / 27\% & 38\% / 12\% \\
\bottomrule
\end{tabular}
\end{table}

\begin{table}[htbp]
\centering
\renewcommand{\thetable}{I.5}
\caption{Unbounded structure and scale errors from the smallest to the largest size (7 benchmarks weighted equally). Structure is the scale-aligned Rel, scale is $\operatorname{mean}(\lvert\ln s^{*}\rvert)$; relative changes in parentheses.}
\label{tab:I5}
\small
\setlength{\tabcolsep}{4pt}
\begin{tabular}{lll}
\toprule
\makecell{Family and condition} & \makecell{Structure: SA Rel} & \makecell{Scale: $\operatorname{mean}\lvert\ln s^{*}\rvert$} \\
\midrule
MoGe-2 w/o K (S$\to$L) & 0.064$\to$0.046 ($-$29\%) & 0.183$\to$0.143 ($-$22\%) \\
MoGe-2 with K (S$\to$L) & 0.078$\to$0.056 ($-$28\%) & 0.137$\to$0.090 ($-$35\%) \\
MoGe-3 w/o K (L$\to$G) & 0.046$\to$0.041 ($-$11\%) & 0.150$\to$0.147 ($-$2\%) \\
MoGe-3 with K (L$\to$G) & 0.059$\to$0.052 ($-$11\%) & 0.090$\to$0.103 (+14\%) \\
UniDepthV2 w/o K (S$\to$L) & 0.066$\to$0.049 ($-$26\%) & 0.252$\to$0.163 ($-$35\%) \\
UniDepthV2 with K (S$\to$L) & 0.065$\to$0.048 ($-$26\%) & 0.216$\to$0.131 ($-$39\%) \\
UniK3D w/o K (S$\to$L) & 0.066$\to$0.048 ($-$28\%) & 0.212$\to$0.127 ($-$40\%) \\
UniK3D with K (S$\to$L) & 0.065$\to$0.047 ($-$27\%) & 0.183$\to$0.097 ($-$47\%) \\
Metric3D-v2 with K (S$\to$g) & 0.059$\to$0.043 ($-$27\%) & 0.179$\to$0.147 ($-$18\%) \\
\bottomrule
\end{tabular}
\end{table}

\end{document}